\documentclass[10pt,twocolumn,letterpaper]{article}

\usepackage[pagenumbers]{iccv} 

\usepackage{graphicx}
\usepackage{amsmath}
\usepackage{amssymb}
\usepackage{booktabs}
\usepackage{blindtext}

\definecolor{iccvblue}{rgb}{0.21,0.49,0.74}
\usepackage[pagebackref,breaklinks,colorlinks,allcolors=iccvblue]{hyperref}

\usepackage{float} 
\usepackage{colortbl}
\usepackage{tikz}
\usepackage{wasysym}

\usepackage[export]{adjustbox}
\usepackage{listings}

\definecolor{codegreen}{rgb}{0,0.6,0}
\definecolor{codegray}{rgb}{0.5,0.5,0.5}
\definecolor{backcolour}{RGB}{245,248,250}
\definecolor{emph}{RGB}{166,88,53}
\definecolor{nightblue}{RGB}{9,49,105}
\definecolor{keywords}{RGB}{207,33,46}
\definecolor{lightpurple}{RGB}{130,81,223}

\lstdefinestyle{mystyle}{
    backgroundcolor=\color{backcolour},   
    commentstyle=\color{codegreen},
    keywordstyle=\color{keywords},
    stringstyle=\color{nightblue},
    basicstyle=\ttfamily\footnotesize,
    breakatwhitespace=false,         
    breaklines=true,                 
    captionpos=b,                    
    keepspaces=true,                 
    showspaces=false,                
    showstringspaces=false,
    showtabs=false,                  
    tabsize=2,
    emph={AutoTokenizer,AutoModelForSequenceClassification,Explainer},
    emphstyle={\color{emph}},
    emph={[2]from_pretrained,compute_table},
    emphstyle={[2]\color{lightpurple}}
}

\newcommand{\Ximg}{{\includegraphics[scale=0.07]{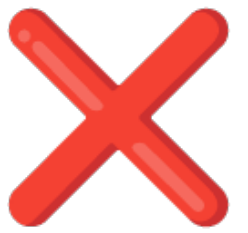}}\xspace}

\newcommand{\na}{\textcolor{gray}{n/a}\xspace}

\usepackage{soul,xcolor}
\definecolor{attention_red}{rgb}{0.80, 0.55, 0.57}

\usepackage[T1]{fontenc}
\usepackage[cal = pxtx, scr = dutchcal]{mathalpha}

\usepackage{pifont}
\usepackage{multirow}

\newcommand{\loss}{\mathcal{L}}
\newcommand{\simx}{\textit{s}}%

\newcommand{\attention}[1]{\sethlcolor{attention_red}\hl{#1}}

\newcommand{\gt}[1]{\textcolor{ForestGreen}{#1}}

\newcommand{\gtbox}{\textcolor{green}{$\Box$}}

\newcommand{\subsec}[1]{\noindent\textbf{#1}~~}

\newcommand{\clevr}{CLEVR\mbox{-}Change\xspace}
\newcommand{\openIn}{OpenImages-I\xspace}
\newcommand{\coco}{COCO-Inpainted\xspace}
\newcommand{\stdfull}{Spot-the-Diff\xspace}
\newcommand{\std}{STD\xspace}
\newcommand{\virat}{VIRAT Video\xspace}

\newcommand{\PG}{pointing game accuracy\xspace}
\newcommand{\MA}{pointing game$^+$ accuracy\xspace}
\newcommand{\MAshort}{PG$^+$\xspace}
\newcommand{\PGshort}{PG\xspace}
\newcommand{\bertscore}{BERTScore\xspace}

\newcommand{\captioning}{\textcolor{Plum}{captioning}\xspace}
\newcommand{\Captioning}{\textcolor{Plum}{Captioning}\xspace}
\newcommand{\editing}{\textcolor{Periwinkle}{editing}\xspace}
\newcommand{\Editing}{\textcolor{Periwinkle}{Editing}\xspace}
\newcommand{\localization}{\textcolor{Magenta}{localization}\xspace}
\newcommand{\Localization}{\textcolor{Magenta}{Localization}\xspace}

\newcommand{\tab}{TAB\xspace}
\newcommand{\tabforidc}{TAB4IDC\xspace}

\newcommand{\llava}{LLaVA-1.5-13B\xspace}
\newcommand{\minigemini}{Mini-Gemini-7B\xspace}
\newcommand{\internvl}{InternVL2-8B\xspace}

\newcommand{\editlogo}{\raisebox{-0.14cm}{\includegraphics[scale=0.025]{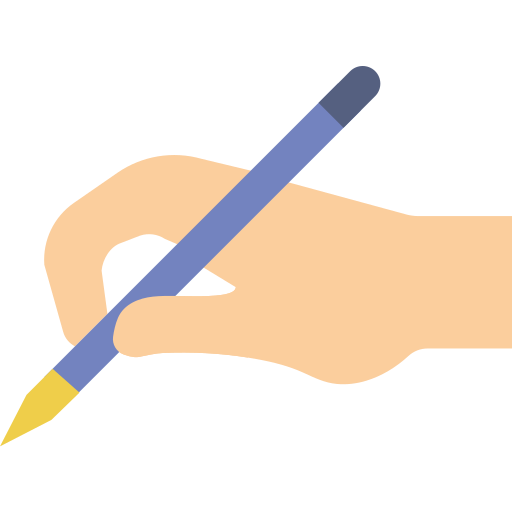}}}
\newcommand{\clevrlogo}{\raisebox{-0.1cm}{\includegraphics[scale=0.025]{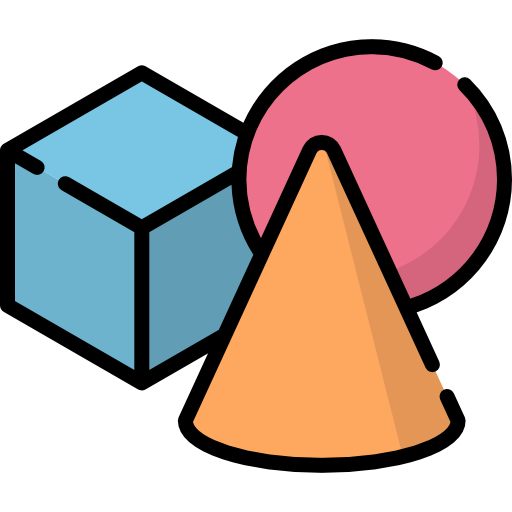}}}
\newcommand{\stdlogo}{\raisebox{-0.09cm}{\includegraphics[scale=0.03]{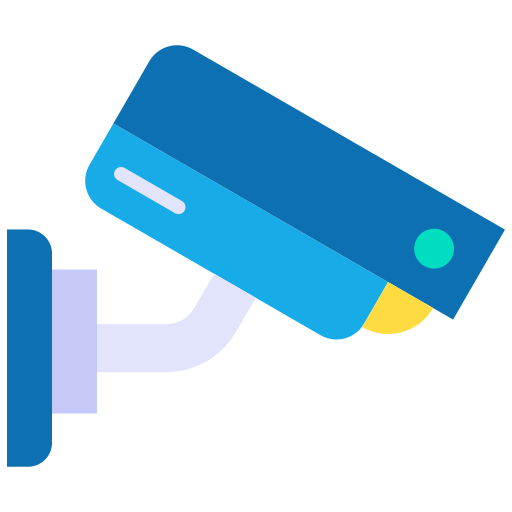}}}
\newcommand{\openlogo}{\raisebox{-0.075cm}{\includegraphics[scale=0.025]{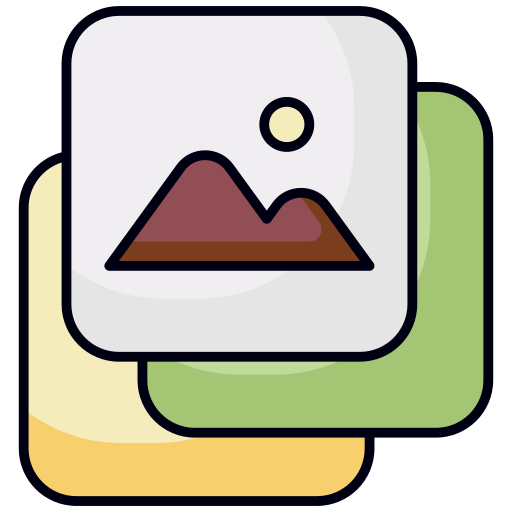}}}
\newcommand{\cocologo}{\raisebox{-0.075cm}{\includegraphics[scale=0.08]{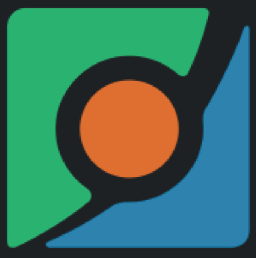}}}

\newcommand{\zeroatt}{\textsc{ZeroAttention}\xspace}
\newcommand{\crtatt}{\textsc{CorrectAttention}\xspace}

\newcommand{\VITlow}{B/16\xspace}
\newcommand{\VIThigh}{B/32\xspace}
\newcommand{\clipforidc}{CLIP4IDC\xspace}

\newcommand{\increase}[1]{(\textcolor{ForestGreen}{+#1})}
\newcommand{\increasenoparent}[1]{\textcolor{ForestGreen}{+#1}}
\newcommand{\decrease}[1]{(\textcolor{red}{-#1})}
\newcommand{\decreasenoparent}[1]{\textcolor{red}{-#1}}

\definecolor{whitesmoke}{rgb}{0.96, 0.96, 0.96}
\newcommand{\class}[1]{\sethlcolor{whitesmoke}\hl{#1}}

\newcommand{\cmark}{\ding{51}}%
\newcommand{\xmark}{\ding{55}}%
\newcommand{\greentick}{\textcolor{ForestGreen}{\cmark}\xspace}
\newcommand{\redxmark}{\textcolor{red}{\xmark}\xspace}

\definecolor{Gray}{gray}{0.9}

\newcolumntype{a}{>{\columncolor{Gray}}c}
\newcolumntype{x}{>{\columncolor{Gray}}l}
\def\institute#1{\gdef\@institute{#1}}

\def\paperID{****} 
\def\confName{ICCV}
\def\confYear{2025}

\newcommand{\papertitle}{\tab: Transformer Attention Bottlenecks enable User Intervention and Debugging in Vision-Language Models}
\title{\papertitle}

\author{Pooyan Rahmanzadehgervi$^\dagger$\\
{\tt\small pooyan.rmz@gmail.com}
\and
Hung Huy Nguyen$^\dagger$\\
{\tt\small hhn0008@auburn.edu}
\and
Rosanne Liu$^\clubsuit$\\
{\tt\small mimosavvy@gmail.com}
\and
Long Mai$^\diamondsuit$\\
{\tt\small mai.t.long88@gmail.com}
\and
Anh Totti Nguyen$^\dagger$\\
{\tt\small anh.ng8@gmail.com}\\
\hspace{-6cm}{\small $\dagger$ Auburn University\hspace{1cm}$\clubsuit$ Google DeepMind, ML Collective\hspace{1cm}$\diamondsuit$ Adobe Research}
}

\begin{document}

\twocolumn[{
\renewcommand\twocolumn[1][]{#1}
\maketitle
\begin{center}
\begin{flushleft}
\includegraphics[width=\linewidth]{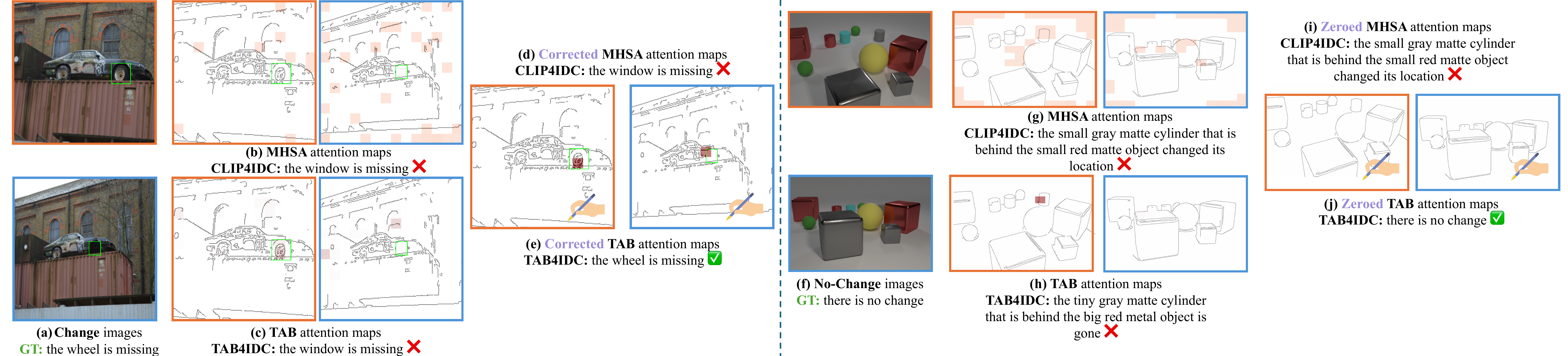}
\end{flushleft}
\captionof{figure}{
In a \class{change} case in \openIn, both CLIP4IDC and TAB4IDC generate a wrong caption (b--c).
However, guiding VLMs to look by highlighting {\scalebox{0.7}\editlogo} attention patches inside \gtbox~in MHSA attention maps does not cause CLIP4IDC output to change (d).
In contrast, intervention on the TAB bottleneck causes TAB4IDC output to change to an expected caption (e).
Similarly, zeroing out the TAB attention map causes TAB4IDC to generate ``there is no change'' as expected (j).
However, zeroing out MHSA attention values in CLIP4IDC results in no changes to the caption (i), showing that the highlighted input patches (g) might not causally contribute to CLIP4IDC outputs.
}
\label{fig:teaser}
\end{center}
}]

\begin{abstract}
Multi-head self-attention (MHSA) is a key component of Transformers \cite{vaswani2017attention}, a widely popular architecture in both language and vision.
Multiple heads intuitively enable different parallel processes over the same input. 
Yet, they also obscure the attribution of each input patch to the output of a model \cite{Chefer_2021_ICCV,bansal2020sam}.
We propose a novel 1-head Transformer Attention Bottleneck (\tab) layer, inserted after the traditional MHSA architecture, to serve as an attention bottleneck for interpretability and intervention.
Unlike standard self-attention \cite{vaswani2017attention}, \tab constrains the total attention over all patches to $\in [0, 1]$.
That is, when the total attention is 0, no visual information is propagated further into the network, and the vision-language model (VLM) would default to a generic, image-independent response (\cref{fig:teaser}j).
To demonstrate the advantages of \tab, we train VLMs with \tab to perform image-difference captioning.
Over three datasets, our models perform similarly to baseline VLMs in captioning but the bottleneck is \textbf{superior in localizing} changes and in identifying when no changes occur.
\tab is the \textbf{first architecture to enable users to debug} by editing attention, which often produces expected outputs by VLMs.
\end{abstract}

\section{Introduction}
\label{sec:intro}

Failures of image-comparison models, \eg, in identifying a face or changes in parking lot across two CCTV views, can lead to life-critical consequences such as denial of unemployment benefits \cite{Sato2023Sep} or false arrests by police \cite{Hill2021Jan,Burke2021Jul,Ryan-Mosley2021Apr,Harwell2021Apr}.
Even state-of-the-art (SotA) VLMs (GPT-4, LLaVA, Gemini) also perform poorly at describing image changes \cite{tong2024eyes}, leaving their decoded speech incomprehensible to users. 
A faithful visualization of what VLMs look at while identifying differences across two images is key to understanding their decoded outputs.


Vision Transformers (ViTs) \cite{dosovitskiy2021an} are the backbone architecture of modern image comparison and VLMs; however, there is no reliable method for accurately identifying what exact input patches they are attending to due to the following possible obstacles: 
(a) lack of registers \cite{darcet2024vision,RudyGilm_register}; 
(b) many self-attention heads (MHSA) in many layers show diverse or diffused patterns \cite{chen2022gscorecam}.
Feature attribution methods could show plausible regions in an attribution map.
But we find that modifying the attention map in ViTs may not cause the outputs to change at all, \ie, no causal relationship between attention (or attribution map) with VLM outputs is established in the literature (\cref{fig:teaser}).

To address the above problems, we propose \tab, a novel, 1-head co-attention \cite{lu2019vilbert} layer, inserted after the standard MHSA blocks \cite{vaswani2017attention}, to serve as a bottleneck that (1) shows exact per-patch attention values, requiring no post-hoc processing; (2) can be supervised with groundtruth bounding-box annotations during training; and (3) can be edited by users (\cref{fig:teaser}e and j) at test time for debugging or human-machine collaboration.

In this work, we test \tab in VLMs trained to caption changes given two input images.
Change captioning is an ideal application for \tab because (1) groundtruth attention maps are available; (2) the attention bottleneck directly filters the visual information that flows into the language model, and therefore editing visual attention has a meaningful, direct impact on a VLM's responses (\cref{fig:framework}).

Change captioning (CC) has a wide range of applications including remote sensing \cite{ferrod2024towards,hoxha2022change}, camera surveillance \cite{jhamtani2018learning}, medical imaging \cite{li2023dynamic,lu2024spot,mimc-dffvqa}, urban planning \cite{Sun_2024_CVPR}, and describing birds \cite{yao2022image,yanl2c} or photo edits \cite{tan2019expressing,black2024vixen}.
CC methods \cite{park2019robust,qiu2021describing,yao2022image} often rely on post-hoc techniques to visualize their ViT attention and do not offer any mechanism for editing attention at test time.
We insert \tab into a SotA VLM \cite{guo2022clip4idc} and train it for change captioning.
Our experiments on three different benchmarks show that
\footnote{Code and data are available on \href{https://github.com/anguyen8/TAB}{GitHub}.}:

\begin{itemize}

    \item With \tab, VLMs outperform their direct no-\tab baselines \cite{guo2022clip4idc} and other ViT-based models consistently on three \captioning benchmarks (\cref{sec:caption}).

    \item In addition to a standard captioning loss, we supervise the attention in \tab using groundtruth bounding-box annotations, enabling \tab to be superior in localizing changes to the common MHSA attention of ViTs (\cref{sec:localizatioin}).

    \item Our test-time \textcolor{Periwinkle}{intervention} experiments show strong causality between the attention values in \tab and VLM responses (\cref{sec:editability}).
    That is, editing \tab to focus on objects corresponding to human-annotated changes causes VLMs to improve captions (\cref{fig:teaser}c).
    In contrast, zeroing out the attention in \tab will always result in a generic ``there is no change'' response by VLMs (\cref{fig:teaser}f). 
    
    \item Interestingly, the \tab of our VLM trained on \openIn \cite{nguyenWACVChange} localizes changes on an \emph{unseen} multi\mbox{-}change dataset (Spot-the-Diff \cite{jhamtani2018learning}) remarkably well---better than baseline VLMs without \tab \cite{guo2022clip4idc} and only slightly worse than a SotA detector \cite{sachdeva2023change} (\cref{sec:zeroshot}).

\end{itemize}

\section{Related Work}
\label{sec:related}

\subsec{ViT attention visualization}
Many feature attribution methods for ViTs aggregate attention heads across all layers into a single heatmap using linear weights computed via gradients \cite{Chefer_2021_ICCV,subramanian2022reclip} or heuristics \cite{chen2022gscorecam,phan2024fast,abnar2020quantifying}.
Other approaches visualize the similarity scores between text and image tokens \cite{kim2021vilt,samek2021explaining}.
However, these methods can be unreliable \cite{nourelahi2022explainable,bansal2020sam} and may not provide meaningful utility to users \cite{nguyen2021effectiveness}.
In contrast, \tab eliminates the need for post-processing and is the first approach to enable users to intervene on an attention map at test time.

\begin{figure*}[ht]
\centering
\includegraphics[width=\textwidth]{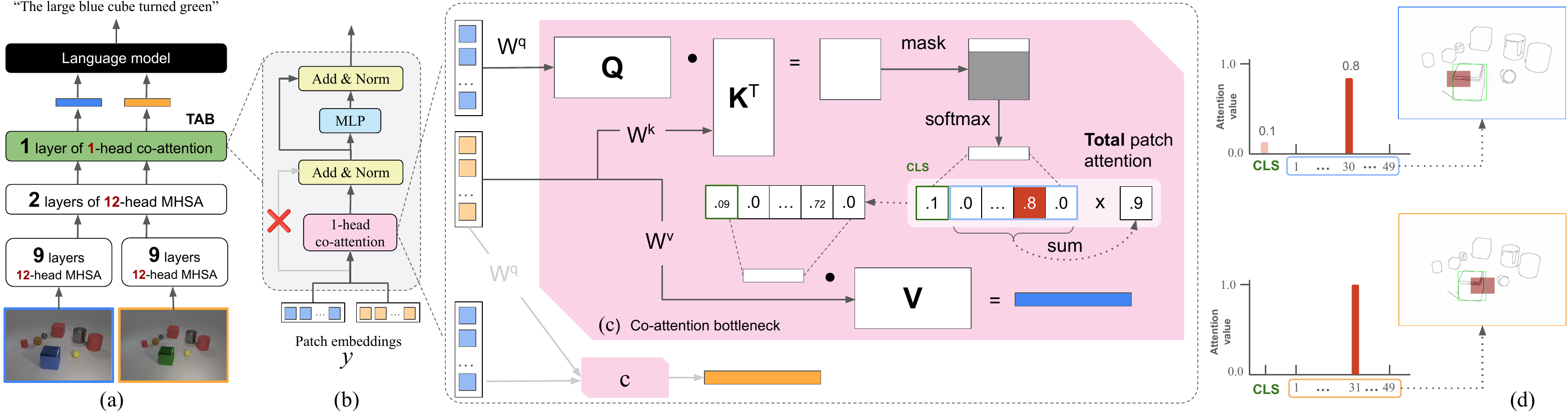}
\caption{
We insert an attention bottleneck (\tab) into a VLM architecture and train it to perform change \captioning (a).
\tab is a 1-head co-attention \cite{lu2019vilbert} Transformer layer (b) that has a skip connection removed \Ximg so that the 1-head attention (c) serves as an information bottleneck, akin to the forget gate in LSTM, directly controls the visual information that flows into the language model.
The \attention{attention} maps in \tab 
show precisely how much each patch contributes to the VLM response (d).
}
\label{fig:framework}
\end{figure*}

\subsec{Class embeddings} On top of image patches, ViTs include an extra [CLS] token \cite{dosovitskiy2021an}. Consequently, the largest ViT, \ie $14\times14$px patches on $224\times224$ images, generates 256 image patches plus the [CLS] as the final representation of the input image. In each attention head, the \emph{softmax} function sums to 1 over all 257 patches. Prior works \cite{zhai2022scaling,beyer2022better} showed that using global average pooling instead of adding the [CLS], can sufficiently aggregate the embeddings from image patches to represent the images, which also improves the scalability and the memory usage of ViTs \cite{zhai2022scaling}. In our CC setup, we visualize the \attention{attention} values (\cref{fig:framework}) for both \class{change} and \class{no\mbox{-}change} pairs, and desire an empty, all-zero attention map for \class{no\mbox{-}change} pairs, \ie, all image patches should have 0 value in their \attention{attention}. 
Simply removing the [CLS] embedding from the ViT causes the \emph{softmax} to sum to 1 over all image patches, which may yield a misleading interpretation that there \textit{is} always a change even for \class{no\mbox{-}change} pairs, as it generates nonzero attention values.

Since the [CLS] pays more attention to class-specific tokens \cite{caron2021emerging}, another approach to improve the scalability of ViTs is to identify the most important tokens based on their attention value to the [CLS] token \cite{haurum2023tokens,osti_10431879}. The tokens with small attention values can be removed to reduce the computations further. Yet, in this work, we drop the image patches and directly use the [CLS] to bottleneck the final representation for both \localization and \captioning tasks. 


\subsec{Concept bottleneck models} (CBMs) \cite{pmlr-v119-koh20a,chen2020concept,kazhdan2020now,losch2019interpretability,wang2023learning,pham2024peeb} are a class of self-interpretable classifiers that map the input image to a set of human-predefined features (a.k.a. ``concepts''), which are then used to derive model predictions.
That is, CBMs offer a bottleneck that linearly combines the pre-defined features.
In contrast, \tab offers a bottleneck for attention over patch embeddings.
These two approaches are orthogonal and can be used together.

\subsec{Model \editing} enables users to change a certain parameter of a machine learning model at test time for intervention or reprogramming \cite{meng2022locating}.
These efforts ranged from editing facts in language models \cite{meng2022locating,mitchell2022memory}, to textual descriptors in image classifiers \cite{pham2024peeb} or neurons in classifiers and generative models \cite{bau2020rewriting,bau2020understanding}.
Our work is the first to explore editing the \emph{attention} of a VLM's vision encoder to allow sanity checks, debugging, and causal evaluation of attention maps, and differs from modifying facts in the model \editing literature \cite{meng2022locating}.

\subsec{Change \captioning}
Most CC models often rely on MHSA to extract image features and to compare two images \cite{yao2022image,guo2022clip4idc,park2019robust}.
However, their ViTs often contain many layers and attention heads, posing a challenge to feature attribution and intervention.
In contrast, we take the VLM architecture by \cite{guo2022clip4idc} designed for CC and \textbf{simplify} it by converting the last cross-attention layer (inside the vision encoder) into a \tab (\cref{fig:framework}), which results in (1) improved \captioning accuracy; (2) superior change \localization via attention; and (3) \textcolor{Periwinkle}{editability} for users. Furthermore, most CC research trained a VLM to caption changes directly without explicitly teaching it how to localize changes \cite{yao2022image,guo2022clip4idc,park2019robust,tan2019expressing}.
In contrast, we supervise VLMs (via \tab) with localization annotations, yielding improved captions and change \localization capability.

\section{Method}
\label{sec:method}

\subsection{Architecture}
To demonstrate \tab for VLMs, we take \clipforidc architecture \cite{guo2022clip4idc}, a SotA CC model, and simplify its last cross-attention (MHSA \cite{vaswani2017attention}) layer into a \tab (\cref{fig:framework}a) and retrain this modified architecture (\tabforidc) for captioning.
Specifically, below are the architectural changes we make:
\begin{enumerate}
    \item Replacing cross-attention by co-attention \cite{lu2019vilbert} (\cref{fig:framework}c). 
    \item Reducing the number of heads from 12 to 1.
    \item Removing the initial skip connection (\cref{fig:framework}b).
    \item Adding a dynamic attention gating mechanism that zeroes out the [CLS] attention when the total attention over image patches is zero (\cref{fig:framework}c).
\end{enumerate}

\tabforidc first extracts visual features from the two images and then localizes changes (if any) using \tab (\cref{fig:framework}a).
Then, it passes the [CLS] embedding of each image to a language model (LM), which outputs a caption that either states that no changes are detected or what the changes are.

\subsec{Vision encoder}
Following \clipforidc, we use the initial convolution layer in CLIP's ViT \cite{radford2021learning} (for both \VITlow and \VIThigh)
to patchify each image $I_{1}$ and $I_{2}$ into $n$ patches $\in \mathbb{R}^{d}$:

\begin{small}
\begin{align}
I_{1}&=\{p_{\text{cls}}^1,p_1^1,...,p_{n}^1\}+pos
\\
I_{2}&=\{p_{\text{cls}}^2,p_1^2,...,p_{n}^2\}+pos
\end{align}
\end{small}
where $p_{\text{cls}}$ and $p_{i}$ represent the [CLS] and patch tokens of each image, respectively, and 
$pos \in \mathbb{R}^{(n+1)\times d}$ is the positional embedding. Each token is expressed by an embedding vector of size $d=768$ after the convolution layer.
Next, we use a 9-layer, 12-head MHSA Transformer encoder ($F$) to extract features from each single image (\cref{fig:framework}a). 

Then, we concatenate the patch embeddings from two images and pass them to a 2-layer, 12-head ViT ($F'$), akin to CrossViT \cite{chen2021crossvit}, to enable \textbf{image comparison}:
\begin{small} 
\begin{align}\label{eq:vit}
y=F'\big(\{&F(I_{1})+e_1,F(I_{2})+e_2\}+pos'\big)
\end{align}
\end{small}
We use a learnable joint positional embedding, $pos'\in\mathbb{R}^{2(n+1)\times d}$, which lets the model represent the position of each patch in the input images. 
Following \clipforidc \cite{guo2022clip4idc} and CrossViT \cite{chen2021crossvit}, we use a pair of extra learnable positional embeddings, $e_1$ and $e_2\in\mathbb{R}^{d}$, to represent the order of each image in the pair. From \cref{eq:vit}, $y$ can be seen as $y=[f^1 \mid\mid f^2]$ (\cref{fig:framework}b), where $f^1,f^2 \in\mathbb{R}^{(n+1)\times d}$ are the contextualized embeddings of the first and second image after $F'$, and $[{} \mid\mid {}]$ is the concatenation operation.

\subsec{Bottleneck}
We take the vanilla MSHA layer \cite{vaswani2017attention} and simplify it by restricting its information flow through the co-attention \cite{lu2019vilbert} mechanism for our attention bottleneck.

The co-attention \cite{lu2019vilbert}, given the intermediate contextualized embeddings $f^1$ and $f^2$ from each image, obtains $Q$, $K$, and $V$ as in the vanilla Transformer layer \cite{vaswani2017attention}. Yet, $K$ and $V$ are swapped between the attention branches of $f^1$ and $f^2$ (\cref{fig:framework}c). 
That is, the co-attention produces two embeddings, $H^1(f^1, f^2)$ and $H^2(f^2, f^1)$ via:

\begin{small}
\begin{equation}\label{eq:translayer}
Q = W^qf^q,~~~~K = W^kf^k,~~~~V = W^kf^k
\end{equation}
\begin{equation}\label{eq:weights}
A = softmax(\frac{QK^T}{\sqrt{d_k}}) 
\end{equation}
\begin{equation}\label{eq:gate}
A' = A_\text{cls}\times \sum_{i=1}^n{A_{\text{cls},i}} 
\end{equation}
\begin{equation}\label{eq:coatt}
H^q(f^q, f^k) = W^oA'V
\end{equation}
\end{small}

where $W^q$, $W^k$, and $W^o$ are learnable weights.
$A_{\text{cls},i}$ is the attention value between the [CLS] row in the $Q$ matrix and the $i$-th patch row in the $K$ matrix.

For intervention and interpretability, we reduce the number of heads from 12 to 1, yielding only one attention map (\cref{eq:weights}) per image (\cref{fig:framework}d). We then remove the first skip connection in the Transformer layer to block any information leak in the flow (\cref{fig:framework}b). Finally, we mask all the tokens in the attention map, except the [CLS]. This yields $H^1_\text{cls}, H^2_\text{cls}\in\mathbb{R}^{d}$, from \cref{eq:coatt}, as the final image representations obtained in \tab (\cref{fig:framework}c). We then apply a linear projection to map them to $P^1, P^2\in\mathbb{R}^{512}$.

\subsec{Caption generation}
 Like \clipforidc \cite{guo2022clip4idc}, we use a 6-layer encoder-decoder LM that translates the outputs of the vision encoder into a caption.
 Specifically, we pass $P^1$ and $P^2$ to a 3-layer, 8-head language encoder to generate intermediate representations, which are then decoded into a caption via a 3-layer, 8-head decoder. (see \href{https://github.com/anguyen8/TAB/blob/main/modules/module_decoder.py}{code} and \href{https://github.com/anguyen8/TAB/blob/main/modules/decoder-base/decoder_config.json}{configuration}).

\subsection{Training}
Like CLIP4IDC \cite{guo2022clip4idc}, our training has two stages: (1) adaptation and (2) captioning. 
Stage 1 aims to align the visual representations of two images to their change caption in the text space via CLIP's contrastive loss \cite{radford2021learning}.
This pre-trained vision encoder from Stage 1 then serves as an initialization for Stage 2, where both the vision encoder and the LM are trained with a next-token prediction and an attention loss.

\subsec{Stage 1: Adaptation}
We use a retrieval loss function \cite{guo2022clip4idc} to adapt the visual and textual representations in Stage 1:

\begin{equation}
\loss_{i2t}=\frac{-1}{B}\sum\limits_{i}^B \log\frac{\exp(\simx(v_i,g_i)/\tau)}{\sum_{j=1}^B \exp(\simx(v_i,g_j)/\tau)}
\end{equation}
\begin{equation}
\loss_{t2i}=\frac{-1}{B}\sum\limits_{i}^B \log\frac{\exp( \simx(v_i,g_i)/\tau))}{\sum_{j=1}^B \exp(\simx(v_j,g_i)/\tau)}
\end{equation}
\begin{equation}
\loss_{\text{Stage1}}=\loss_{i2t}+\loss_{t2i} \label{eq:retrieval_loss}
\end{equation}
where $\loss_{i2t}$ and $\loss_{t2i}$ are the loss functions of image pair-to-text and text-to-image pair retrieval with a learnable temperature ($\tau$), respectively. $v_i$ is the mean pooling of $P^1$ and $P^2$; and $g_i$ is the textual embeddings extracted from the caption using CLIP \cite{radford2021learning} text encoder for $i$-th image-text pair. 

\subsec{Stage 2: Captioning}
Stage 2 training has two losses: $
\loss_{\text{Stage2}}=\loss_{CE}+\loss_{att}
$. Like \cite{guo2022clip4idc}, we supervise VLMs to generate a caption given two images using a Cross Entropy (CE) loss over the next word (see \cref{appsec:training}).

In addition to a captioning CE loss, we also use an \textbf{attention supervision} loss ($\loss_{att}$) to encourage TAB to focus on relevant image patches. 
We derive a groundtruth attention map $G \in [0,1]^{n}$ from a human-annotated bounding box $b$ (available in \clevr and \openIn) by assigning 1 to the patch that has the largest intersection with $b$ and 0 to all other $n-1$ patches.
That is, when there is no change, the attention value should be 1 for the [CLS] embedding and 0 for all $n$ patch embeddings.

Our attention loss minimizes the cosine distance between the groundtruth attention map $G$ and the attention map $A_\text{cls} \in [0,1]^{n}$ in TAB:

\begin{equation}
\loss_{att} = 1 - \frac{<A_\text{cls}. G>}{||A_\text{cls}||.||G||} \label{eq:det_loss}
\end{equation}

\section{Evaluation}
\subsection{Datasets}
\label{sec:data}

We train and evaluate \tabforidc's \captioning and \tab's \localization capabilities on three distinct CC datasets that vary in the number of changes, types of changes, difficulty levels, and types of images.

\subsec{\clevr \clevrlogo}
\cite{park2019robust} contains $\sim$80K image pairs where 40K contain exactly one change (\class{change}) and the other 40K has no changes (\class{no\mbox{-}change}).
Images contain primitive 3D shapes and are rendered using Blender \cite{blendercite}.

\subsec{\openIn \openlogo}
\cite{nguyenWACVChange} consists of $\sim$2.5M pairs of images ($\sim$1.28M for each \class{change} and \class{no\mbox{-}change}) where one object is removed.
In each \class{change} pair, one image is taken from OpenImages \cite{openimages} and the other is a clone but has one object removed by the LaMa inpainter \cite{suvorov2021resolution} based on its human-annotated bounding box. 
Following \citet{nguyenWACVChange}, we apply random rotation and zoom to a random image in each pair to create viewpoint differences. We use the object-name annotations of OpenImages and follow Park et al. \cite{park2019robust} to create change captions using predefined templates (see \cref{appsec:caption_generation}).

\subsec{\stdfull (\std) \stdlogo}
\cite{jhamtani2018learning} has $\sim$13K image pairs captured from CCTV videos in \virat dataset \cite{viratdataset}. 
Each image pair contains more than one change and is captioned by humans. 
For \localization evaluation, we add a \class{no\mbox{-}change} class (1K pairs of identical images) to the test set, which initially contains 1K pairs of changed images.

\begin{table*}[ht]
\centering
\caption{We conduct experiments on three distinct change \captioning datasets of varying image sources, numbers of changes, and difficulties.
}
\label{tab:data}
\resizebox{0.8\textwidth}{!}{%
     \begin{tabular}{lrclccl}
     \toprule
     Dataset & \# Pairs & Image type & Change type & \# Changes & \class{No\mbox{-}change}& Annotations \\
     \midrule
     \multirow{2}{*}{\clevrlogo ~\cite{park2019robust}} & \multirow{2}{*}{79,606} & \multirow{2}{*}{3D-rendered} & Color, Material&\multirow{2}{*}{1}&\multirow{2}{*}{\greentick}& caption \\ 
     &&&Add, Remove, Move&&&bounding box\\
     \midrule
     \multirow{3}{*}{\openlogo ~\cite{nguyenWACVChange}} & \multirow{3}{*}{2,559,918} & \multirow{3}{*}{real-world} & \multirow{3}{*}{Add, Remove} & \multirow{3}{*}{1} &\multirow{3}{*}{\greentick}& caption \\
     &&&&&&bounding box\\
     &&&&&& object name\\
     \midrule
     \stdlogo ~\cite{jhamtani2018learning} & 13,192 & real-world & Add, Remove, Move & $\in[1...7]$ &\redxmark& caption\\
     \midrule
     \cocologo ~\cite{sachdeva2023change} & 60,000 & real-world & Add, Remove & $\in[1...24]$ &\redxmark& bounding box\\
     \bottomrule
     \end{tabular}}
\end{table*}

\subsection{Metrics}
\label{sec:eval}

\subsec{Change \captioning}
Following \cite{park2019robust,guo2022clip4idc}, we use five metrics for evaluating the n-grams of generated captions:
BLEU-4 \cite{BLEU2002},
METEOR \cite{banerjee-lavie-2005-meteor}, ROUGE-L \cite{lin-2004-rouge}, and CIDEr \cite{vedantam2015cider}. 
For a semantics assessment, we also use \bertscore \cite{zhang2019bertscore,ge2024openagi} to evaluate the meanings of captions.

\subsec{Attention map visualization}
The attention bottleneck in \tabforidc attends to the change locations in the attention map (\cref{eq:weights}) and, therefore, acts as a change detector. To evaluate \tab's \localization performance, we derive heatmaps from $A_\text{cls}$, \ie, the [CLS] row in the attention map (\cref{eq:weights}) in each branch of the co-attention (see \cref{fig:framework}c).
This row contains exactly $n+1$ \attention{attention values} per image ($n=49$ and $196$ in \VIThigh and \VITlow, respectively). 
We drop the [CLS] attention value ($A_{\text{cls},\text{cls}}$) and keep the $n$ attention values of the image patches (see \cref{fig:framework}d), \ie, the $A_{\text{cls},i}$ values  where $i\in[1...n]$.

We reshape the vector $A_{\text{cls}, i}$ of size $n\in\{49, 196\}$ to a square matrix of size $7\times7$ and $14\times14$ for \VIThigh and \VITlow, respectively. Then, we upscale the reshaped $A_{\text{cls},i}$ to the input image size via the nearest-neighbor interpolation (see \cref{fig:vis_code_ours}) for \tab and the CC baselines \cite{guo2022clip4idc,park2019robust,qiu2021describing} to obtain the final heatmaps.
Compared to attention visualization methods in CC baselines \cite{guo2022clip4idc,park2019robust} that use cubic interpolation (see \cref{fig:vis_code_org}), our nearest-neighbor approach better shows the actual values of the attention map per patch (\cref{fig:our-vis}; 1b vs. 1c), and yields fewer nonzero pixels on \class{no\mbox{-}change} (see \cref{fig:our-vis}; 2b vs. 2c).


\subsec{Change \localization}
In this task, we aim to approximate the accuracy of the attention map in highlighting the image patches that contribute to the predicted captions in \class{change} and \class{no\mbox{-}change} pairs. A common approach to attention map evaluation is using the \emph{object localization} techniques \cite{zhou2016learning, bansal2020sam,agarwal2020explaining,zhang2018top}. However, these metrics, such as \PG (\PGshort) \cite{zhang2018top}, evaluate the attention maps by localizing a target object present in an image \cite{zhou2016learning, bansal2020sam,agarwal2020explaining}. That is, for change \localization, they evaluate the attention maps only for \class{change} pairs that contain a groundtruth changed object. The attention map should not highlight any objects in \class{no\mbox{-}change} pairs, as there are no changes across the images.

Here, we introduce \textbf{\MA (\MAshort)} to measure the localization accuracy for both \class{change} and \class{no\mbox{-}change} pairs. Similar to \cite{agarwal2020explaining,fong2017interpretable,cao2015topdown}, we derive multiple bounding boxes from the attention map, which is reshaped to the input image size, by thresholding it at different $t\in[0.0:0.05:0.95]$ values. At each $t$ value, for \class{change} pairs, accuracy is the frequency of when a derived bounding box intersects with the groundtruth box, and for \class{no\mbox{-}change} pairs, accuracy is when the attention map is all zero after the thresholding (see \cref{fig:our-vis}-2d). We choose the best $t$ value that yields the best mean accuracy over \class{change} and \class{no\mbox{-}change} pairs on the validation sets.

\subsec{Effects of visualization on localization accuracy} Since the visualization method in CC baselines \cite{guo2022clip4idc,park2019robust,qiu2021describing}, always yields an attention map with at least one peak even for \class{no\mbox{-}change} pairs (see \cref{fig:our-vis}-2b), their original localization is poor (see \cref{tab:vis_effect_onLoc}). However, our proposed thresholding and interpolation method (\cref{fig:vis_code_ours}) improves the baseline methods \cite{guo2022clip4idc,park2019robust,qiu2021describing} \localization accuracy (see \cref{tab:vis_effect_onLoc}), specifically on \class{no\mbox{-}change} pairs.

\subsec{Attention \editing}
A caption in change captioning must include two pieces of information:
(1) Is there a change across two images? (2) If there is a change, which object?
Therefore, we measure how \textcolor{Periwinkle}{editing} TAB's attention values affect these two pieces of information in the generated caption via:
\begin{itemize}
    \item \textbf{\class{Change} and \class{no\mbox{-}change} accuracy:} 
    We evaluate the correctness of the captions by comparing their template to the groundtruth caption templates (see \cref{appsec:caption_generation}). The evaluation is based on exact word matching, and we report the model’s accuracy in expressing whether a change has occurred or not.
    \item \textbf{Object accuracy:} \class{Change} pairs in \openlogo~are created by removing one object \cite{nguyenWACVChange}, for which we can access the groundtruth object name. Since a major error in \tabforidc captions is predicting a wrong object name, we measure the object accuracy before and after editing the attention values by (1) extracting the object name from the predicted caption using templates (\cref{appsec:caption_generation}), and (2) applying exact-match against groundtruth object name.
\end{itemize}

\begin{table}
\centering
\caption{The common interpolation method used for upscaling the attention map leads to low \MAshort accuracy (0.0\%) on \openlogo, here for \clipforidc \cite{guo2022clip4idc}. We use the nearest neighbor approach (\cref{fig:vis_code_ours}) and, then, threshold the attention map at different values to find the most important patches based on their values. This improves the baseline method's overall \MAshort (90.15\%).}
\label{tab:vis_effect_onLoc}
\resizebox{\columnwidth}{!}{%
\begin{tabular}{lcccccc}
\toprule
Method & ViT & Interpolation & Thresh. & {\class{Change}} & {\class{No\mbox{-}change}} & \textbf{Mean}  \\
\midrule
Baseline & \VIThigh & cubic  & \xmark  & 79.98 & 0.0 & 39.99  \\
Ours & \VIThigh & nearest neighbor  & \cmark  & 84.10 & 96.21 & \textbf{90.15}          \\
\bottomrule
\end{tabular}}
\end{table}

\section{Results}

\subsection{\tabforidc outperforms other Transformer-based baselines in change \captioning}
\label{sec:caption}

By design, \tab reduces the capacity of the vision encoder, which may limit the VLM's captioning performance.
Therefore, after building a \tab bottleneck in the baseline VLM (\clipforidc) and training it, we test how our model (\tabforidc) performs in \captioning compared to CLIP4IDC and other VLMs in the supervised learning setting.

\subsec{Experiment}
Following \cite{yao2022image,park2019robust}, we train our model on CLEVR-Change (\clevrlogo) and compare it to \clipforidc \cite{guo2022clip4idc}, DUDA \cite{park2019robust}, MCCFormer \cite{qiu2021describing} and IDC-PCL \cite{yao2022image}. 
We also train and test our models on the real-world images in OpenImages-I (\openlogo) and the camera surveillance dataset of STD (\stdlogo). 
We reproduce the results of DUDA, MCCFormers, and CLIP4IDC using their public code bases \cite{clip4idc,duda,iccv}.

\subsec{Results}
At the higher-resolution attention maps, \ie, smaller patch sizes of \VITlow, \tabforidc consistently outperforms the counterpart model of the \clipforidc \cite{guo2022clip4idc}. This suggests that \textbf{the bottleneck does not limit the information flowing into the LM}. Furthermore, our proposed training regime with attention supervision improves the \captioning accuracy (\increasenoparent{1.5} vs. \increasenoparent{1.1} on \openlogo; \cref{tab:cap-results-open}), compared to the baseline. With low-res attention (ViT-\VIThigh), \tabforidc performs on par with its baseline (\decreasenoparent{0.6} on \clevrlogo, \increasenoparent{1.4} on \openlogo; \cref{tab:cap-results-clevr,tab:cap-results-open}).
Our model is the SotA on \clevrlogo~and \openlogo~perhaps because the training set is large and the groundtruth boxes are available to supervise the attention maps.

\begin{table}
\centering
\caption{\textbf{\Captioning \clevrlogo:} \tabforidc outperforms CC methods that use RNN attention, DUDA \cite{park2019robust}, and the models that combine CNN and Transformers \cite{qiu2021describing}. 
Compared to pure Transformer-based VLMs, \tabforidc is far better than IDC-PCL and on par with its predecessor, \clipforidc.}
\label{tab:cap-results-clevr}
\resizebox{\columnwidth}{!}{%
\begin{tabular}{lccccax}
\toprule
Method & ViT & \textbf{B-4} & \textbf{M} & \textbf{R} & \textbf{C}& \textbf{\bertscore}\\
\midrule
DUDA \cite{park2019robust} & \xmark &   47.3      &    33.9   & \na &  112.3     &    67.4      \\
MCCFormer-D \cite{qiu2021describing}  & \xmark &    52.4     &   38.3      &  \na    & 121.6        & \na \\
MCCFormer-S \cite{qiu2021describing}  & \xmark &    57.4    &   41.2  &     \na      &   125.5      & \na  \\
IDC-PCL \cite{yao2022image} & \xmark &   51.2   &    36.2      &    71.7     &     128.9   &  \na \\
\clipforidc \cite{guo2022clip4idc} & \VIThigh &    56.9    &   38.4  &   76.4           &   150.7      &  74.3 \\
\clipforidc \cite{guo2022clip4idc} & \VITlow & 52.9 & 38.7 & 76.4 & 144.7 & 74.2 \\
\tabforidc   &\VIThigh &   55.5   &   38.0      &   76.3  & 149.6 & 73.7 \decrease{-0.6}\\
\tabforidc   &\VITlow   &   56.4   &   38.6     &   77.7  & \textbf{153.0} & \textbf{75.8} \increase{1.6}\\
\bottomrule
\end{tabular}}
\end{table}

\begin{table}
\centering
\caption{\textbf{\Captioning \openlogo:} Our proposed \tabforidc outperforms \clipforidc on captioning changes across natural images. The attention supervision in \tab improves the captioning accuracy in \tabforidc (\increasenoparent{0.4} in \bertscore).}
\label{tab:cap-results-open}
\resizebox{\columnwidth}{!}{%
\begin{tabular}{lcccccax}
\toprule
Method & ViT & Attn. Sup. &\textbf{B-4} & \textbf{M} & \textbf{R} & \textbf{C}& \textbf{\bertscore}\\
\midrule
\clipforidc \cite{guo2022clip4idc} &\VIThigh & \xmark &  82.8 & 56.0 & 93.6 &  296.5 &  92.4\\
\clipforidc \cite{guo2022clip4idc} & \VITlow & \xmark & 87.5 & 60.3 & 95.6 & 328.7 & 95.1\\
\tabforidc   & \VIThigh & \cmark &  91.4   & 62.1    &  94.7   & 302.0 & 93.8 (\increasenoparent{1.4})\\
\tabforidc   & \VITlow   & \cmark &  \textbf{92.4}  &  \textbf{64.5}  &  \textbf{96.9}  & \textbf{335.6} & \textbf{96.6} (\increasenoparent{1.5})\\
\tabforidc   & \VITlow   & \xmark & 91.9  &  63.8  &  96.6  & 324.4 & 96.2 (\increasenoparent{1.1})\\
\bottomrule
\end{tabular}}
\end{table}

The largest difference is on \stdlogo---TAB4IDC outperforms CLIP4IDC at high-res attention (\increasenoparent{5.3} in \VITlow; \cref{tab:cap-results-std}) but underperforms at low resolution (\decreasenoparent{6.8} in \VIThigh; \cref{tab:cap-results-std}).
An explanation is that there are no groundtruth bounding boxes in \stdlogo, so no attention supervision is applied.
Furthermore, the image pairs in \stdlogo~often contain many small changes, which benefit from high-res attention. Moreover, SotA open-source VLMs such as \llava \cite{jiao2024img}, fine-tuned on \stdlogo, perform worse than both \clipforidc and \tabforidc (see \cref{tab:cap-results-std}) in the captioning task. We also find that the SotA generalist change captioning model, \ie, OneDiff \cite{hu2024onediff}, outperforms \tabforidc on \stdlogo by \increasenoparent{5.3} in CIDEr. Perhaps OneDiff's larger vision encoder and language model, L/14 and 7B, respectively, and its more diverse training dataset are the main driving reasons.

\begin{table}
\centering
\caption{\textbf{\Captioning \stdlogo:} \tabforidc with no attention supervision, requires more patches (the higher-resolution ViT-\VITlow) to be on par with \clipforidc and outperforms SotA open-source VLMs, \eg, \minigemini and \llava.}
\label{tab:cap-results-std}
\resizebox{\columnwidth}{!}{%
\begin{tabular}{lccccax}
\toprule
Method & ViT &\textbf{B-4} & \textbf{M} & \textbf{R} & \textbf{C}& \textbf{\bertscore} \\
\midrule
\llava \cite{jiao2024img}& L/14 & 9.9 & 13.1 & 31.4 & 45.8 & 25.9\\
\minigemini \cite{jiao2024img}& - & 10.8 & 13.1 & 33.0 & 53.3 & 26.4\\
\internvl \cite{jiao2024img}& InternViT & 8.4 & 12.8 & 28.5 & 32.2 & 25.3\\
OneDiff \cite{hu2024onediff}& L/14 & 12.8 & 14.6 & 35.8 & \textbf{56.6} & -\\
\midrule
\clipforidc \cite{guo2022clip4idc} & \VIThigh &    11.6  &   14.2  &   35.0 &   47.4  &  \textbf{29.4} \\
\clipforidc \cite{guo2022clip4idc} & \VITlow & 10.0   &  12.7  & 32.6 & 47.6  &  23.0 \\
\tabforidc   & \VIThigh &  11.0    &   13.5   &  33.0   & 37.7 & 22.6 (\decreasenoparent{6.8}) \\
\tabforidc   & \VITlow   & 11.2   &   13.5   &  33.0 & 51.3 & 28.3 (\increasenoparent{5.3}) \\
\bottomrule
\end{tabular}}
\end{table}

\subsection{\tab outperforms MHSA layers and RNN-based attention in change \localization}
\label{sec:localizatioin}

We aim to test how well \tab attention communicates the information (1) that no change is detected (\class{no\mbox{-}change} cases); and (2) of the location of the changed object in \class{change} cases. 


\subsec{Experiment}
We take the VLM trained in \cref{sec:caption} and conduct an object \localization experiment on its \tab to evaluate the \MAshort accuracy across three datasets (\clevrlogo, \openlogo, and \stdlogo). 
We compare its accuracy to CC methods with \localization ability. We hypothesize that given the attention supervision during training, \tab should yield a more precise attention map for each predicted caption.

\begin{figure}
    \centering
    \begin{subfigure}[b]{0.15\textwidth}
    \centering
         \includegraphics[width=\linewidth]{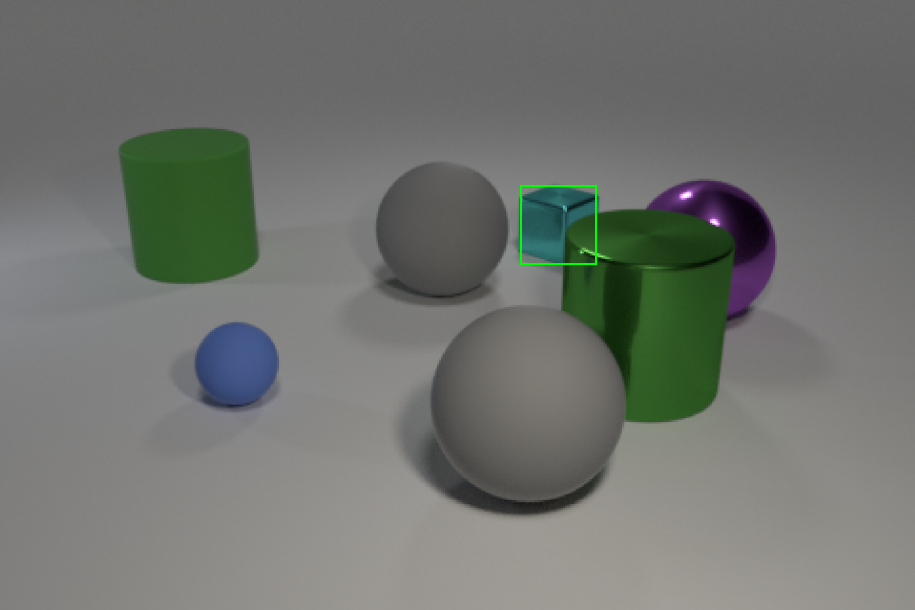}
         \\
         \vspace{-0.1em}
         \includegraphics[width=\linewidth]{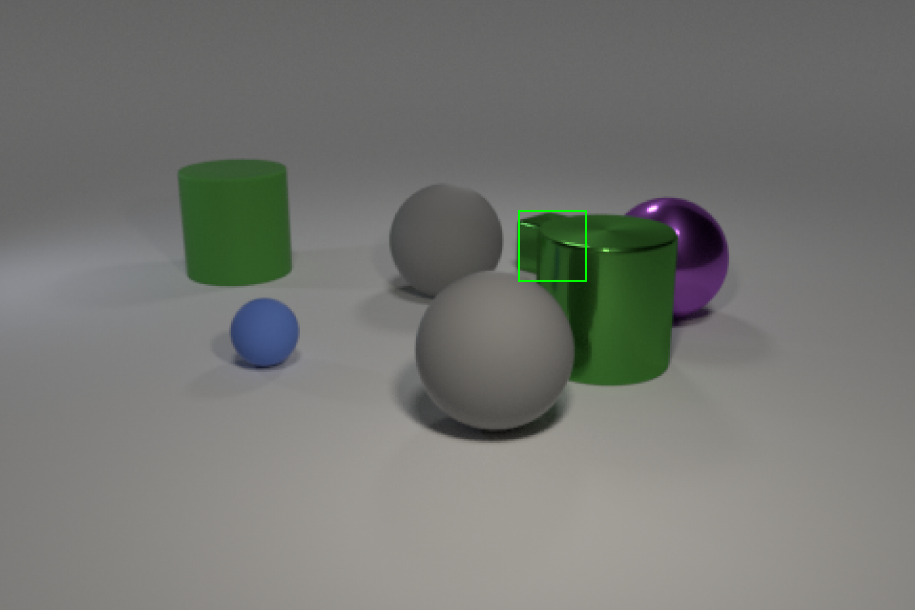}
         \caption{\centering \textbf{\gt{GT:}} the tiny metal thing changed to green \newline \newline \newline}
    \end{subfigure}
    \begin{subfigure}[b]{0.15\textwidth}
    \centering
         \includegraphics[width=\linewidth]{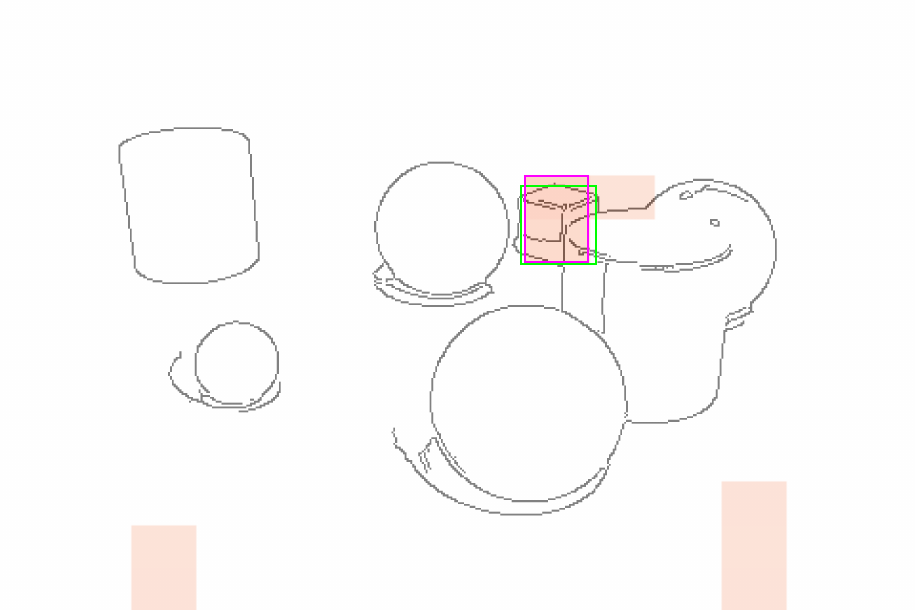}
         \\
         \vspace{-0.1em}
         \includegraphics[width=\linewidth]{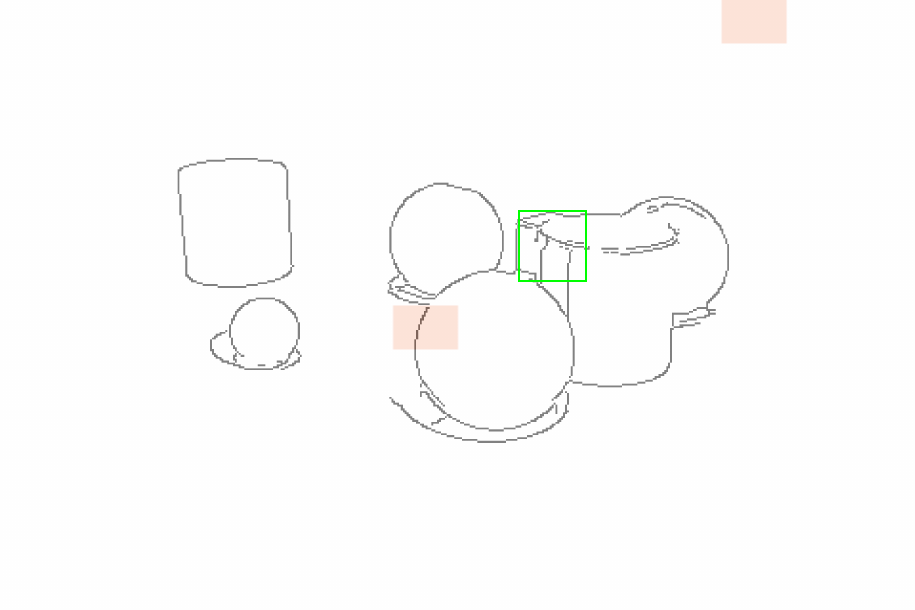}
         \caption{\centering \textbf{\clipforidc:} there is no change \redxmark \newline \newline \newline}
    \end{subfigure}
    \begin{subfigure}[b]{0.15\textwidth}
    \centering
         \includegraphics[width=\linewidth]{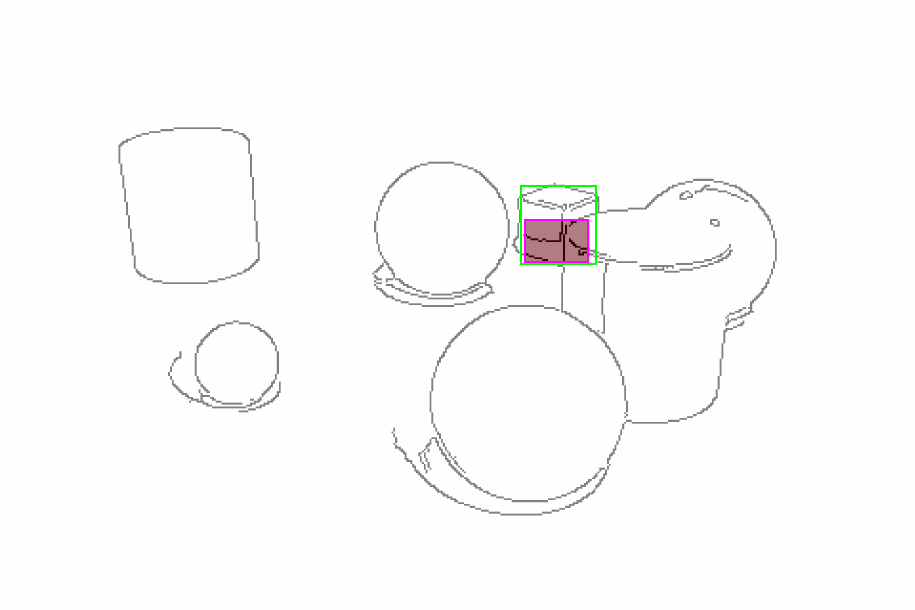}
         \\
         \vspace{-0.1em}
         \includegraphics[width=\linewidth]{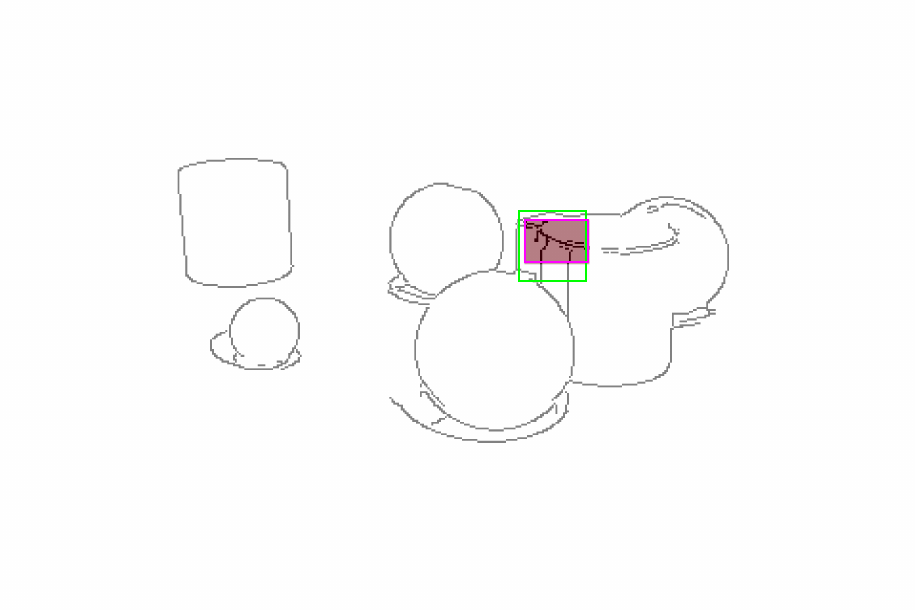}
         \caption{\centering \textbf{\tabforidc:} the small cyan metal block that is behind the big green cylinder changed to green \greentick}
    \end{subfigure}
    \caption{Compared to MHSA layer in \clipforidc (b) \tab better localizes the changed object that contributes to the predicted caption (c), for quantitative results we evaluate \MAshort against the groundtruth (\gtbox) in \clevrlogo.}
    \label{fig:clevr_qual}
\end{figure}

\begin{figure}
    \centering
    \begin{subfigure}[b]{0.15\textwidth}
    \centering
         \includegraphics[width=\linewidth]{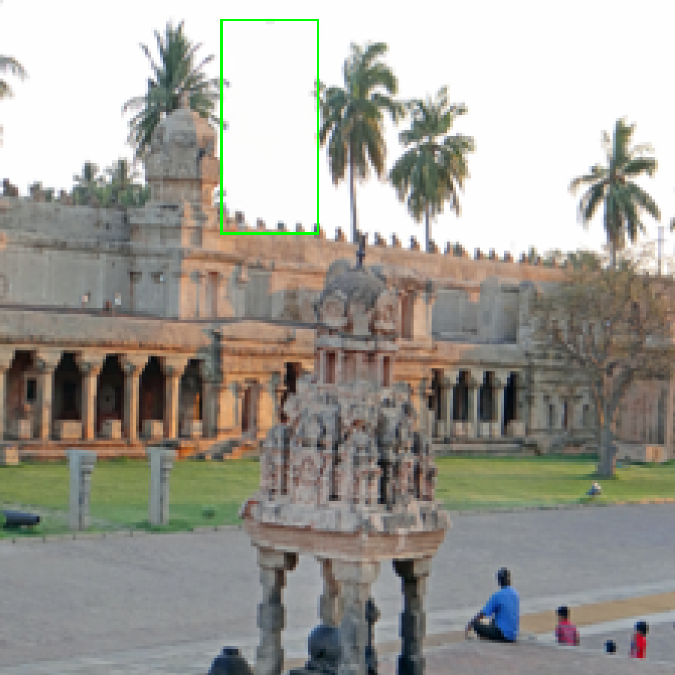}
         \\
         \vspace{-0.1em}
         \includegraphics[width=\linewidth]{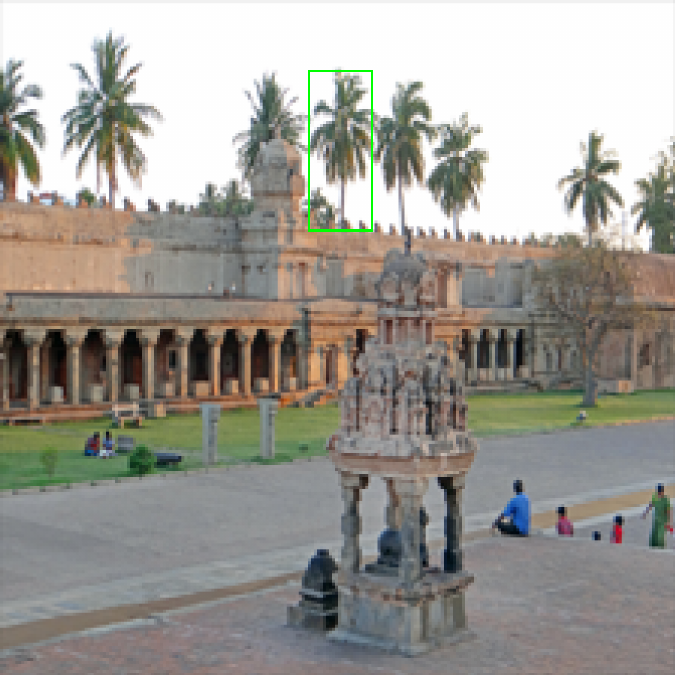}
         \caption{\centering \textbf{\gt{GT:}} the palm tree has been added \newline}
    \end{subfigure}
    \begin{subfigure}[b]{0.15\textwidth}
    \centering
         \includegraphics[width=\linewidth]{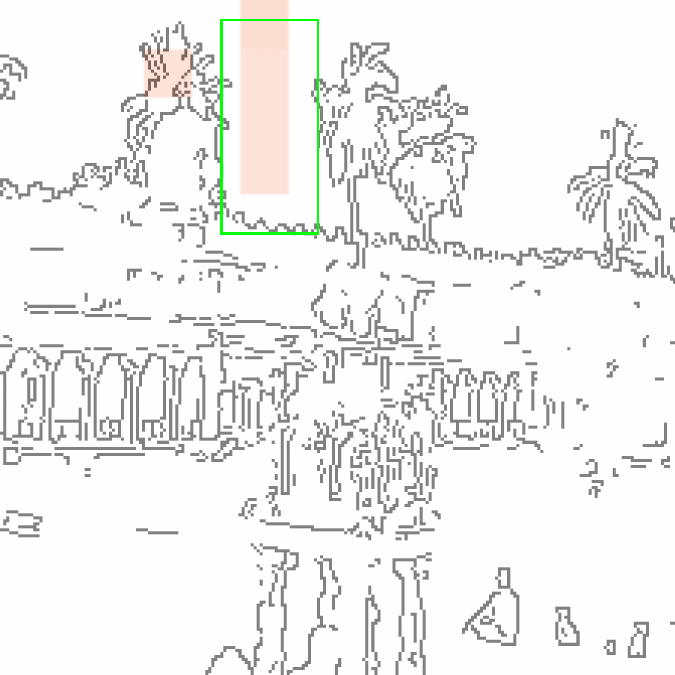}
         \\
         \vspace{-0.1em}
         \includegraphics[width=\linewidth]{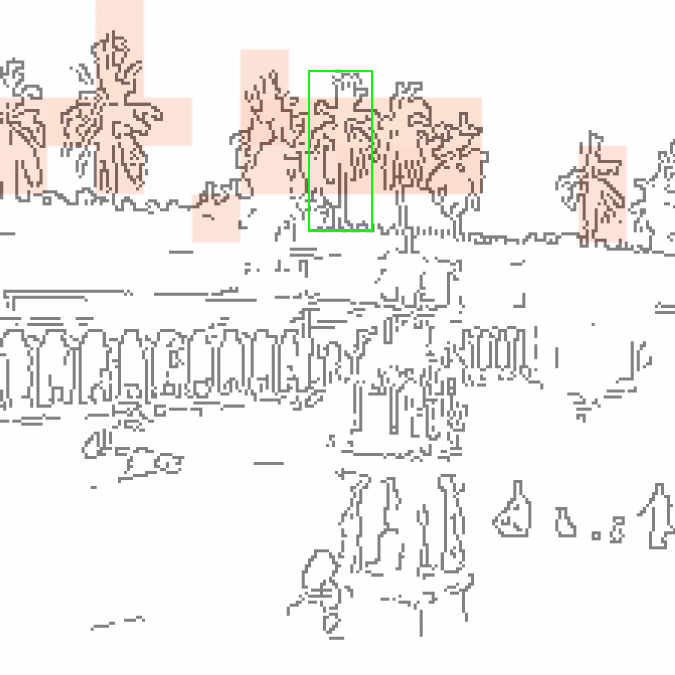}
         \caption{\centering \textbf{\clipforidc:} the palm tree has been added \greentick}
    \end{subfigure}
    \begin{subfigure}[b]{0.15\textwidth}
    \centering
         \includegraphics[width=\linewidth]{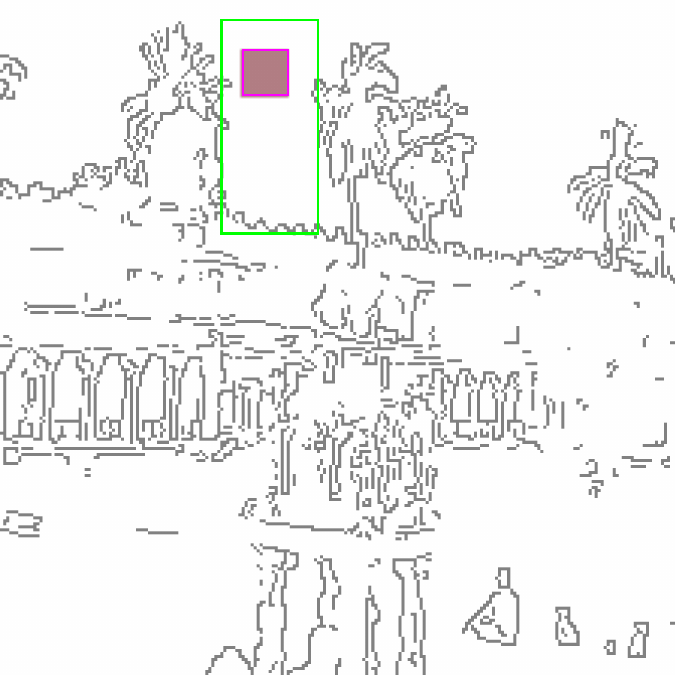}
         \\
         \vspace{-0.1em}
         \includegraphics[width=\linewidth]{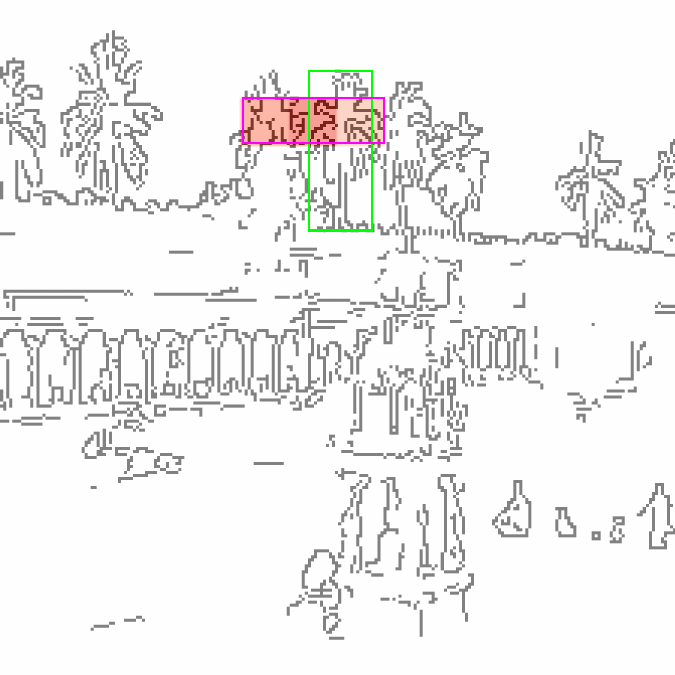}
         \caption{\centering \textbf{\tabforidc:} the palm tree has been added \greentick}
    \end{subfigure}
    \caption{On \openlogo, the MHSA attention of \clipforidc \cite{guo2022clip4idc} often highlights many patches corresponding to the class name (here, \class{palm tree}) in the input image (b). 
    In contrast, \tab points at exactly the only palm tree that is changed.}
    \label{fig:open_qual}
\end{figure}

\begin{figure}
    \centering
    \begin{subfigure}[b]{0.15\textwidth}
    \centering
         \includegraphics[width=\linewidth]{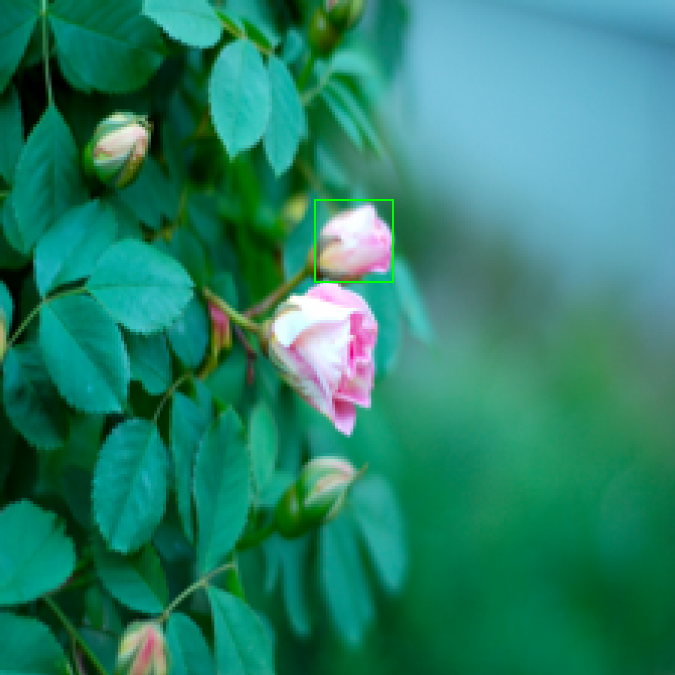}
         \\
         \vspace{-0.1em}
         \includegraphics[width=\linewidth]{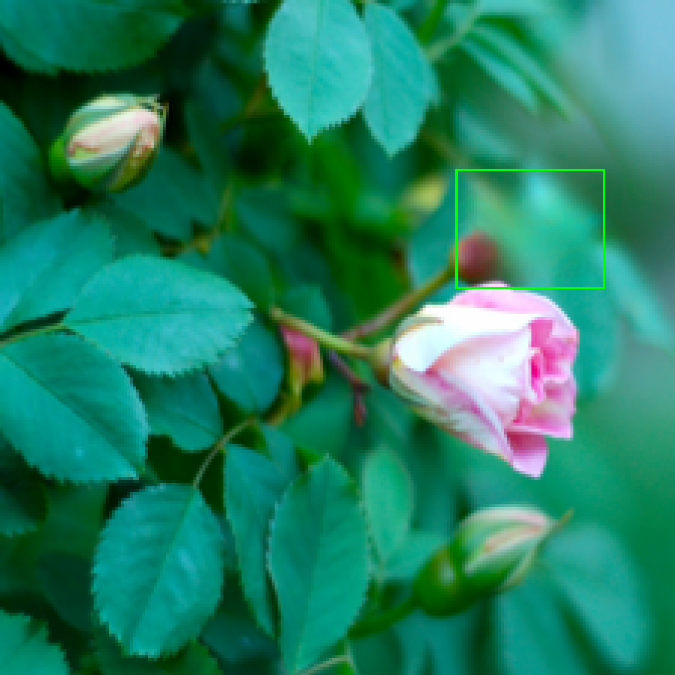}
         \caption{}
    \end{subfigure}
    \begin{subfigure}[b]{0.15\textwidth}
    \centering
         \includegraphics[width=\linewidth]{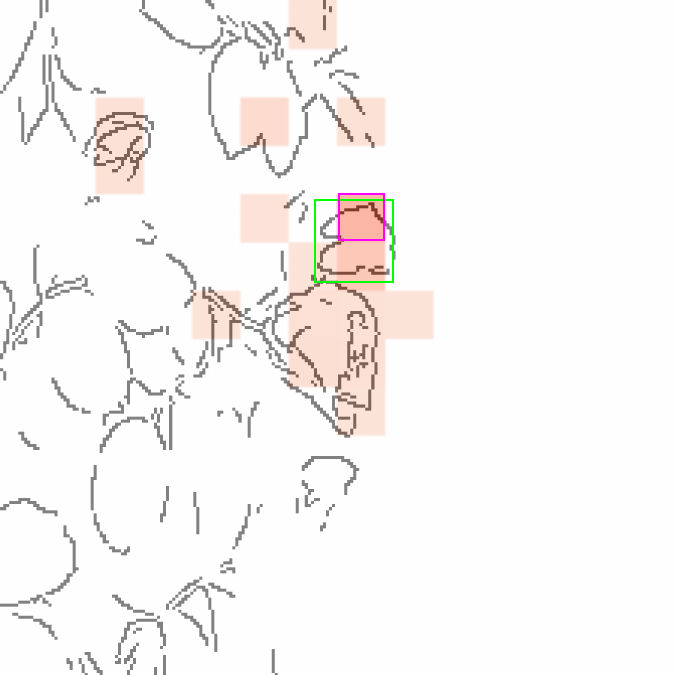}
         \\
         \vspace{-0.1em}
         \includegraphics[width=\linewidth]{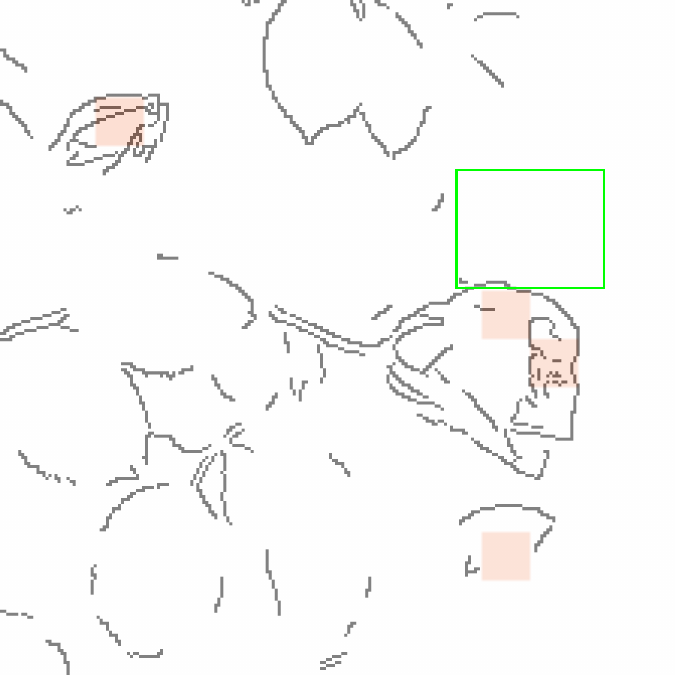}
         \caption{}
    \end{subfigure}
    \begin{subfigure}[b]{0.15\textwidth}
    \centering
         \includegraphics[width=\linewidth]{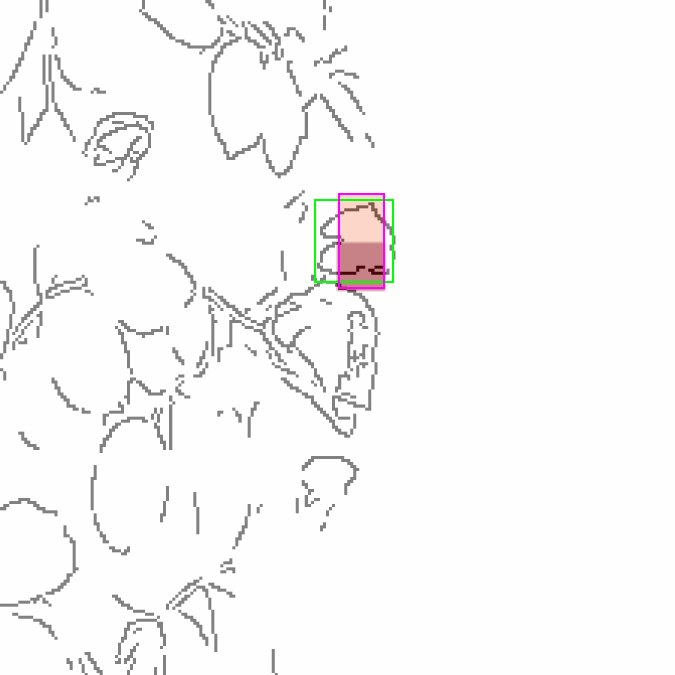}
         \\
         \vspace{-0.1em}
         \includegraphics[width=\linewidth]{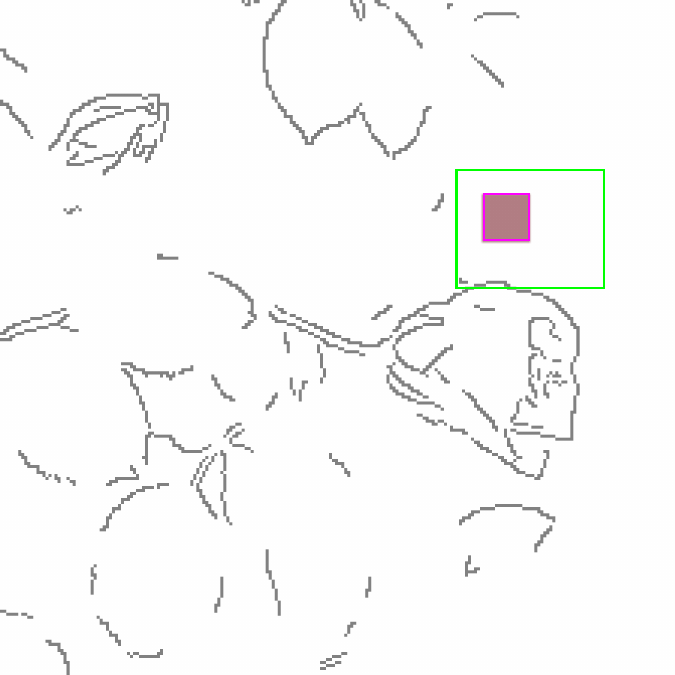}
         \caption{}
    \end{subfigure}
    \caption{\textbf{Ablation:} \tab, without the Attention Supervision often only attends to the location that the object is present (b) in \openlogo. The attention supervision mitigates this problem (c).}
    \label{fig:open_sup}
\end{figure}

\subsec{Results}
Across different attention map resolutions (ViT-\VITlow and \VIThigh) and three datasets (\clevrlogo, \openlogo~and \stdlogo), \tab consistently localizes \emph{object-level} differences better than the MHSA layer in \clipforidc. That is, \textbf{our proposed 1-head co-attention Transformer bottleneck yields attention maps that outperform baselines in communicating the predicted caption for both \class{change} and \class{no\mbox{-}change} pairs} (see \cref{fig:clevr_qual,fig:open_qual}). Perhaps, the reason is our proposed attention supervision during VLM training, where the attention map in \tab is forced to match the groundtruth attention (\increasenoparent{46.21} vs. \increasenoparent{8.75} on \openlogo; \cref{tab:loc-results-open,fig:open_sup}).

\begin{figure}
    \centering
    \begin{subfigure}[b]{0.17\textwidth}
    \centering
         \includegraphics[width=\linewidth]{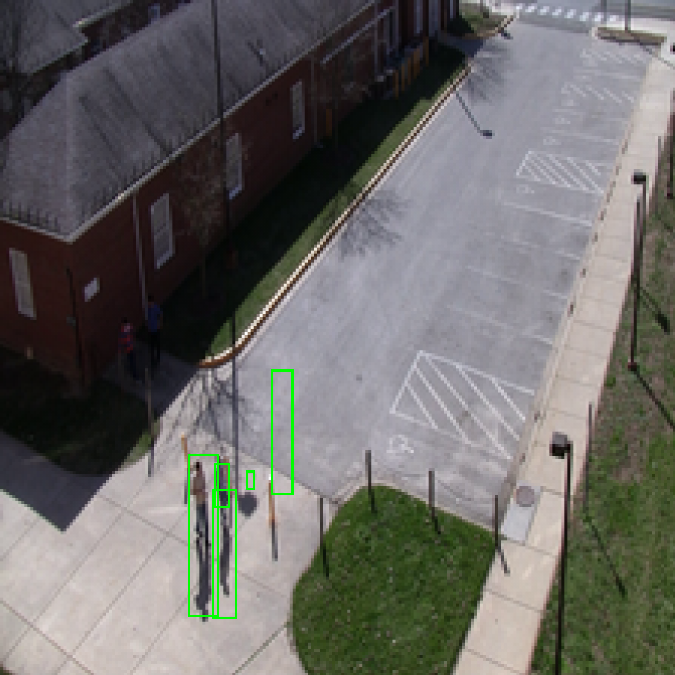}
         \\
         \vspace{-0.1em}
         \includegraphics[width=\linewidth]{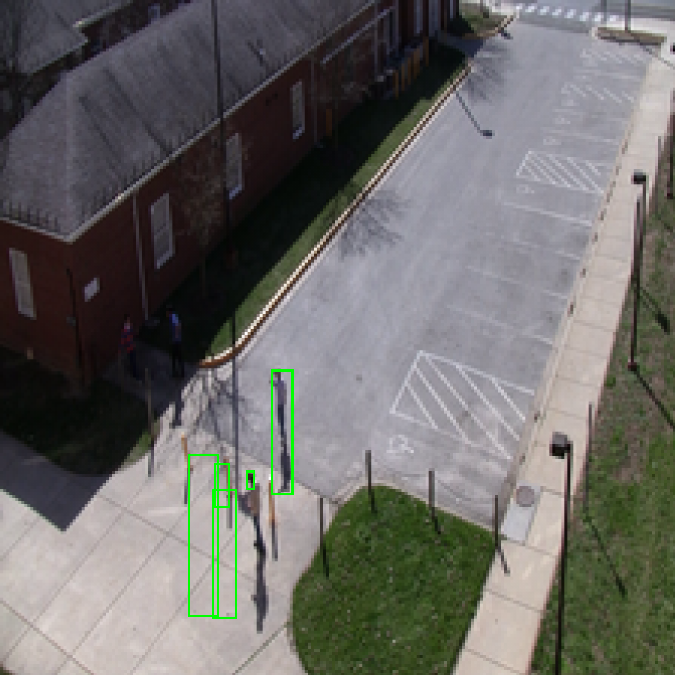}
         \caption{Input images}
    \end{subfigure}
    \begin{subfigure}[b]{0.17\textwidth}
    \centering
         \includegraphics[width=\linewidth]{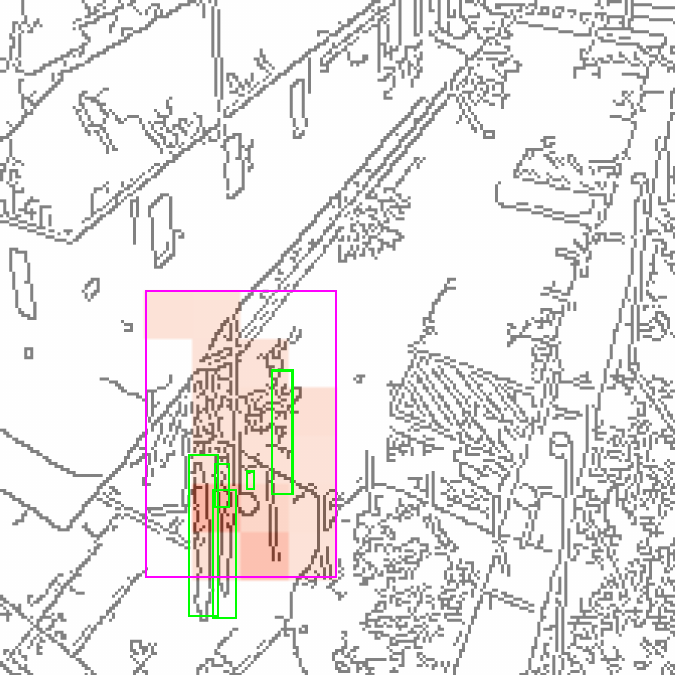}
         \\
         \vspace{-0.1em}
         \includegraphics[width=\linewidth]{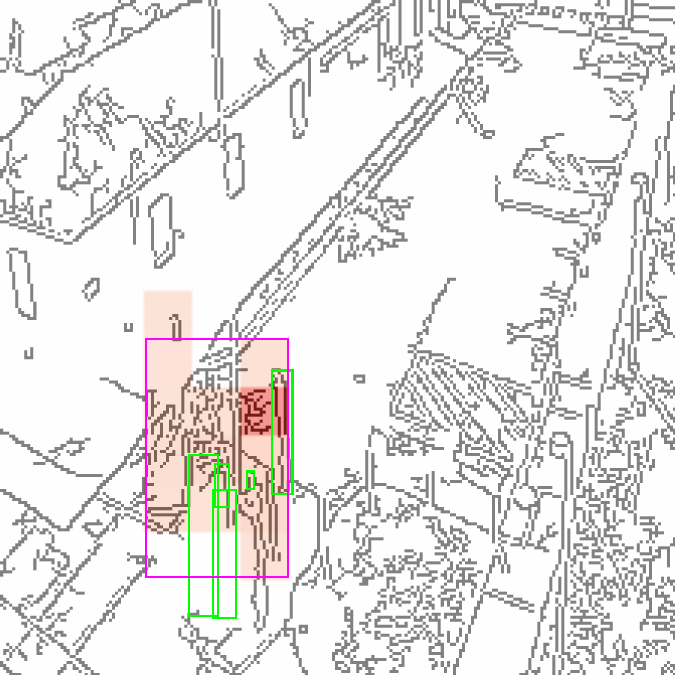}
         \caption{Attention maps in \tab}
    \end{subfigure}
    \caption{The attention maps in \tab (b) can localize multiple changes in \stdlogo. Yet, the values spread over the patches, including the changed objects (\gtbox), which leads to lower attention values. 
    }
    \label{fig:loc_std_main}
\end{figure}

\tab, using the high-res attention (ViT-\VITlow), has the largest delta with its MHSA counterpart in \clipforidc (\increasenoparent{12.28} vs. \increasenoparent{4.98} on \clevrlogo, \increasenoparent{46.21} vs. \increasenoparent{43.97} on \openlogo; \cref{tab:loc-results-clevr,tab:loc-results-open}), compared to the low-res attention (ViT-\VIThigh). The smaller patches in ViT-\VITlow, which help point to small objects in \clevrlogo~and \openlogo, are perhaps the main reason for this delta. On \stdlogo, we use \PGshort for evaluation because it only includes \class{change} pairs, where \tab is SotA change detector, and outperforms \clipforidc by \increasenoparent{3.95} (53.5 vs. 49.55\%). This suggests that the attention map in \tab more frequently attends to image patches that contain the changed object, even without attention supervision (\cref{fig:loc_std_main}).

\begin{table}
\centering
\caption{\textbf{\Localization \clevrlogo:} Under \MAshort, \tab localizes changes far better than the attention maps of CC baselines. 
\tab uses only 1 head compared to the 12-head attention in \clipforidc \cite{guo2022clip4idc} and prior models.
}
\label{tab:loc-results-clevr}
\resizebox{\columnwidth}{!}{%
\begin{tabular}{lcccx}
\toprule
Method & ViT & \textbf{\class{Change}} & \textbf{\class{No\mbox{-}change}} & \textbf{Mean}  \\
\midrule
DUDA \cite{park2019robust}               &     \xmark     & 68.51                   & 94.20             & 81.36        \\
MCCFormer-D \cite{qiu2021describing}     &     \xmark     & 78.85                   & 0.00              & 39.42         \\
MCCFormer-S \cite{qiu2021describing}      &   \xmark      & 72.96                   & 0.61              & 36.78        \\
\clipforidc \cite{guo2022clip4idc} & \VIThigh & 84.10                   & 96.21             & 90.15          \\
\clipforidc \cite{guo2022clip4idc} & \VITlow   & 77.37 & 91.56 & 84.47  \\
\tab  & \VIThigh  & 91.59 & 98.66  & 95.13 \increase{4.98}\\
\tab   & \VITlow   & 94.00 & 99.50  & \textbf{96.75} \increase{12.28}\\
\bottomrule
\end{tabular}}
\end{table}

\begin{table}
\centering
\caption{\textbf{\Localization \openlogo:} \tab substantially outperforms MHSA layers in \clipforidc \cite{guo2022clip4idc} on change localization under \MAshort. 
The high-res backbone (\VITlow) performs \increasenoparent{1.08} better than the low-res (\VIThigh) as most objects in \openlogo~are small. Our proposed attention supervision also improves the \localization performance, especially across \class{change} pairs (32.24 vs. 98.40).
}
\label{tab:loc-results-open}
\resizebox{\columnwidth}{!}{%
\begin{tabular}{lccccx}
\toprule
Method & ViT & Attn. Sup. & \textbf{\class{Change}} & \textbf{\class{No\mbox{-}change}} & \textbf{Mean} \\
\midrule
\clipforidc \cite{guo2022clip4idc} & \VIThigh    &\xmark& 24.55 & 83.74 & 54.14 \\
\clipforidc \cite{guo2022clip4idc} & \VITlow     &\xmark& 6.12 & 99.85 & 52.98 \\
\tab   & \VIThigh &\cmark&  99.56 & 96.66   & 98.11 \increase{43.97}\\
\tab   & \VITlow  &\cmark&  98.40 & 99.98   & \textbf{99.19} \increase{46.21} \\
\tab   & \VITlow  &\xmark&  32.24 & 91.23   & 61.73 \increase{8.75}\\
\bottomrule
\end{tabular}}
\end{table}

\subsection{Intervening \editlogo~by assigning more accurate attention in \tab yields improved VLM captions}
\label{sec:editability}
Our attention bottleneck in \tabforidc successfully performs localization in both \class{change} and \class{no\mbox{-}change} cases.
Furthermore, the overall strong \localization ability of the attention is observed together with the VLM's \captioning performance. 
Here, we aim to test if the attention values in \tab have a cause-and-effect relationship with the captions by \tabforidc. We conduct two \editing experiments: (1) \crtatt, where the user replaces the attention map with groundtruth attention for wrong predictions. We expect that the correct attention maps improve \captioning performance. (2) \zeroatt, where the attention map is replaced with an all-zero map, and we desire that model captions any input as a \class{no\mbox{-}change}.

\subsec{Experiment} Explanations are often seen as a measure to evaluate trust, and a common experiment in the literature is to use them for model debugging \cite{adebayo2020debugging}. We use the groundtruth annotations ($G$) of \clevrlogo~and \openlogo~that we used in \cref{eq:det_loss} to conduct the debugging experiment during inference. Specifically, (1) for those image pairs that the predicted caption is wrong, we replace the attention map ($A_\text{cls}$ in \cref{eq:weights}) with $G$, and (2) for all the pairs, we replace $A_\text{cls}$ with an all-zero vector and make $A_{\text{cls},\text{cls}}=1$. 
Furthermore, we train \clipforidc with the same attention supervision (\cref{eq:det_loss}) as in \tab to investigate the role of our architectural modifications to the MHSA layer. Finally, we evaluate the \editing based on the common errors in the original captions for \tabforidc and \clipforidc with attention supervision.

\subsec{Results}
In \crtatt, replacing the attention map with groundtruth helps VLM to predict captions that match the groundtruth caption more frequently in \tabforidc (see \cref{fig:teaser}). Similarly, in \zeroatt, replacing the original attention map with an all-zero map causes the VLM to predict a \class{no\mbox{-}change} caption for the input pair regardless of the groundtruth class (\cref{appsec:clevr_zeroatt}). On image pairs, where the original predicted caption is wrong, \tab provides an interface during inference to debug the model by correcting its attention values. This suggests that our proposed 1-head attention mechanism in \tab acts as a switch, which turns off the image patches for \class{no\mbox{-}change} pairs and turns them on for \class{change} pairs. This means that not only do \textbf{the explanations in \tab} correlate with the output (\cref{sec:localizatioin}), but they also \textbf{causally contribute to the VLM predictions}.

Specifically, \crtatt improves the \class{change} and \class{no\mbox{-}change} accuracy on both \clevrlogo~and \openlogo~(99.93\% to 100\% on \cref{tab:edit}a; 94.30\% to 100.0\% \cref{tab:edit}b). This is because \tab predicts the caption based on the attention values of the image patches. \zeroatt, conversely, only improves the \class{no\mbox{-}change} accuracy since the new attention values are only correct for \class{no\mbox{-}change}. Moreover, when the attention values in \tab correctly point to the changed object, predicted captions more often broadcast the correct object name (88.92\% to 91.49\%; \cref{tab:edit}a). 

\Editing the attention values of the MHSA layer has \textbf{no} effects on \clipforidc outputs (\cref{appsec:abledit}), which shows the regular MHSA allows information to leak through. However, \tab, even \textbf{without attention supervision, enables user interventions in test time} (see \cref{tab:edit} a and c). This suggests that our 1-head attention bottleneck is perhaps the main reason for the \textcolor{Periwinkle}{editability} feature.

\begin{table}
    \centering
    \caption{\textbf{\Editing \editlogo:} On \openlogo ~and \clevrlogo, \zeroatt alters all the captions to describe \class{no\mbox{-}change} pairs (0\% and 100\% accuracy over \class{change} and \class{no\mbox{-}change}, respectively). \crtatt, in contrast, helps the model to predict the correct pair type more often. Correcting the attention values also helps \increase{2.57} and \increase{2.32} in predicting the correct object name in \openlogo. This shows that, attention values in \tab causally contribute to the VLM output.
    }
    \label{tab:edit}
    \resizebox{\columnwidth}{!}{
    \begin{tabular}{c@{}ccccrcrcr}
    \toprule
         &&&&\multicolumn{2}{c}{Acc. \class{Change}} & \multicolumn{2}{c}{Acc. \class{No\mbox{-}change}}& \multicolumn{2}{c}{Acc. object name}\\
         \cmidrule(rr){5-6}
         \cmidrule(rr){7-8}
         \cmidrule(lr){9-10}
         &Dataset & Sup.& Attention \editlogo & base & \editlogo & base &  \editlogo & base & \editlogo\\
         \midrule
         \multirow{2}{*}{a}&\multirow{2}{*}{\openlogo} & \multirow{2}{*}{\cmark} & \cellcolor{lightgray}\textsc{zero} & \cellcolor{lightgray}99.93 &\cellcolor{lightgray} 0.0 \greentick & \cellcolor{lightgray}100.0 & \cellcolor{lightgray}100.0 \greentick& \cellcolor{lightgray}88.92 & \cellcolor{lightgray}0.0 \greentick\\
         &&& \textsc{correct} $\uparrow$ & 99.93& 100.0 \greentick&100.0& 100.0 \greentick &88.92& 91.49 \greentick\\
         \midrule
         \multirow{2}{*}{b}&\multirow{2}{*}{\clevrlogo} & \multirow{2}{*}{\cmark}  & \cellcolor{lightgray}\textsc{zero}& \cellcolor{lightgray}94.30 & \cellcolor{lightgray}0.0 \greentick&\cellcolor{lightgray} 99.42 & \cellcolor{lightgray}100.0 \greentick& \cellcolor{lightgray}- & \cellcolor{lightgray}-\\
         &&& \textsc{correct} $\uparrow$ & 94.30& 100.0 \greentick&99.42& 100.0 \greentick&-& -\\
         \midrule
         \multirow{2}{*}{c}&\multirow{2}{*}{\openlogo} & \multirow{2}{*}{\xmark} &\cellcolor{lightgray} \textsc{zero} & \cellcolor{lightgray}97.62 & \cellcolor{lightgray}0.0 \cellcolor{lightgray}\greentick&\cellcolor{lightgray}100.0 & \cellcolor{lightgray}100.0 \greentick&\cellcolor{lightgray}86.86 &\cellcolor{lightgray} 0.0 \greentick\\
         &&& \textsc{correct} $\uparrow$ &97.62& 99.83 \greentick&100.0& 100.0 \greentick&86.86& 91.24 \greentick\\
         \bottomrule
    \end{tabular}}
\end{table}

\subsection{\tab is a better zero-shot change localizer compared to the MHSA layers in VLM captioners}
\label{sec:zeroshot}
We use the VLMs trained for CC on \openlogo, which contain only one change, and aim to evaluate if the MHSA layer in \clipforidc \cite{guo2022clip4idc} and \tab in \tabforidc can localize multiple changes in an unseen dataset (\stdlogo) without further training.

\subsec{Experiment} 
We include a SotA change detection (CD) method, CYWS \cite{sachdeva2023change}, as an upper bound, and report the \MAshort on \openlogo~as a baseline accuracy for CYWS, \tabforidc and \clipforidc \cite{guo2022clip4idc}. CYWS \cite{sachdeva2023change} is a U-Net-based CD framework trained on \coco \cocologo, using the bounding box supervision. We consider CYWS \cite{sachdeva2023change} an upper bound accuracy because it is trained on real-world image pairs with multiple changes (\cocologo) similar to \stdlogo.

\subsec{Results}
Overall, \tab outperforms the MHSA layer of \clipforidc \cite{guo2022clip4idc} by $\sim$\increasenoparent{44} points in mean \MAshort (\cref{tab:loc-std-zero}). The main reason is \tab's better performance on \class{no\mbox{-}change} pairs. On average, \tab and the MHSA layer are worse than CYWS \cite{sachdeva2023change} because they use the attention values (softmax output) to localize multiple changes in \stdlogo. We hypothesize that the large gap is due to attention values spreading over many changes (see \cref{appfig:std_zero_loc}), which causes them not to pass the heatmap discretization in \MAshort. On \openlogo, \tab consistently outperforms CYWS \cite{sachdeva2023change} (99.19 vs. 89.76\%). Compared to the MHSA layer in \clipforidc \cite{guo2022clip4idc}, \tab improves the baseline accuracy of a CD method on \openlogo.  

\begin{table}
\centering
\caption{\textbf{Zero-shot \localization:} \tab performs better under \MAshort in zero-shot change \localization in \stdlogo~than the MHSA layer in \clipforidc \cite{guo2022clip4idc}, which is trained on \stdlogo. \tab also has smaller delta with CYWS \cite{sachdeva2023change}, the upper bound zero-shot \localization accuracy on \stdlogo, than \clipforidc \cite{guo2022clip4idc}. 
}
\label{tab:loc-std-zero}
\resizebox{\columnwidth}{!}{%
\begin{tabular}{lccccccxa}
\toprule
 &&&\multicolumn{2}{c}{\class{Change}} & \multicolumn{2}{c}{\class{No\mbox{-}change}} & \multicolumn{2}{a}{Mean}\\
 \cmidrule(lr){4-5}
 \cmidrule(lr){6-7}
 \cmidrule(lr){8-9}
Method &Train &Thresh.&\stdlogo&\openlogo&\stdlogo&\openlogo&\stdlogo&\openlogo\\
\midrule
CYWS \cite{sachdeva2023change} &\cocologo& \xmark  &100.0& 99.91 &0.0& 0.0 &50.0&49.95\\
CYWS \cite{sachdeva2023change}  &\cocologo& \cmark  & 99.92 & 81.73 & 100.0& 99.72& \textbf{99.96}&89.76\\
\midrule
\clipforidc \cite{guo2022clip4idc}  &\openlogo& \cmark  & 74.9 & 24.55 & 12.6 & 83.74& 43.75&54.14\\
\tab   &\openlogo& \cmark & 75.9& 98.40& 100.0& 99.98& 87.95 \increase{44.2}& \textbf{99.19}\\
\bottomrule
\end{tabular}}
\end{table}

\section{Discussion and Conclusion}

\subsec{Limitations}
A limitation of \tab is its reliance on attention loss for accurate \localization in \class{change} pairs. 
Moreover, the attention map in \tab requires more nonzero values over image patches for multi\mbox{-}change images, which makes the \emph{softmax} function spread over more patches (\eg, there are $14\times14 = 196$ patches in ViT-\VITlow compared to $7\times7 = 49$ patches in ViT-\VIThigh). This may cause \tab to have lower attention values over multi-change images (\cref{fig:loc_std_main}). Another constraint is the size of the change in different datasets. For the smaller changes, \tab needs smaller patches for a more accurate \localization and better \captioning performance (\cref{tab:cap-results-std}). Also, the vision encoder in SotA VLMs, \eg CLIP-L/14 in LLaVA 1.5, often lacks additional tokens, such as [CLS] and [REG]. That is, the \emph{softmax} in the attention sums to 1 over the image patches, yielding nonzero attention maps (uniform at best) even when there is no change---which is \emph{not} desired.

We propose \tab, a self-explainable and editable bottleneck layer with a 1-head attention for CC. Our interactive interface allows users to intervene in decision-making, by which one can correct and audit VLMs' decisions. An intriguing future direction is to extend \tab to other core vision tasks, such as classification, image similarity comparison \cite{phan2022deepface,phan2024fast}, and image matching \cite{wiles2021co}. Given \tab's causal contribution to VLMs' predictions, another valuable direction is to adapt \tab to more general-purpose VLMs, such as LLaVA \cite{liu2024improved}, to improve their interpretability. Additionally, one can employ a VLM to audit the attention values within \tab.

\clearpage
\subsection*{Acknowledgment}
We thank Peijie Chen, Thang Pham, Giang Nguyen, and Tin Nguyen at Auburn University for their feedback and discussions of the earlier results.
AN was supported by the NSF Grant No. 1850117 \& 2145767, and donations from NaphCare Foundation \& Adobe Research.


{
    \small
    \bibliographystyle{ieeenat_fullname}

}

\clearpage

{
   \newpage
       \onecolumn
        \centering
        \Large
        \textbf{\papertitle}\\
        \vspace{0.5em}Supplementary Material \\
        \vspace{1.0em}
}

\appendix

\section{Implementation details}
\label{appsec:training}

\subsection{Framework}
\begin{figure*}[hb]
\centering
\includegraphics[width=\linewidth]{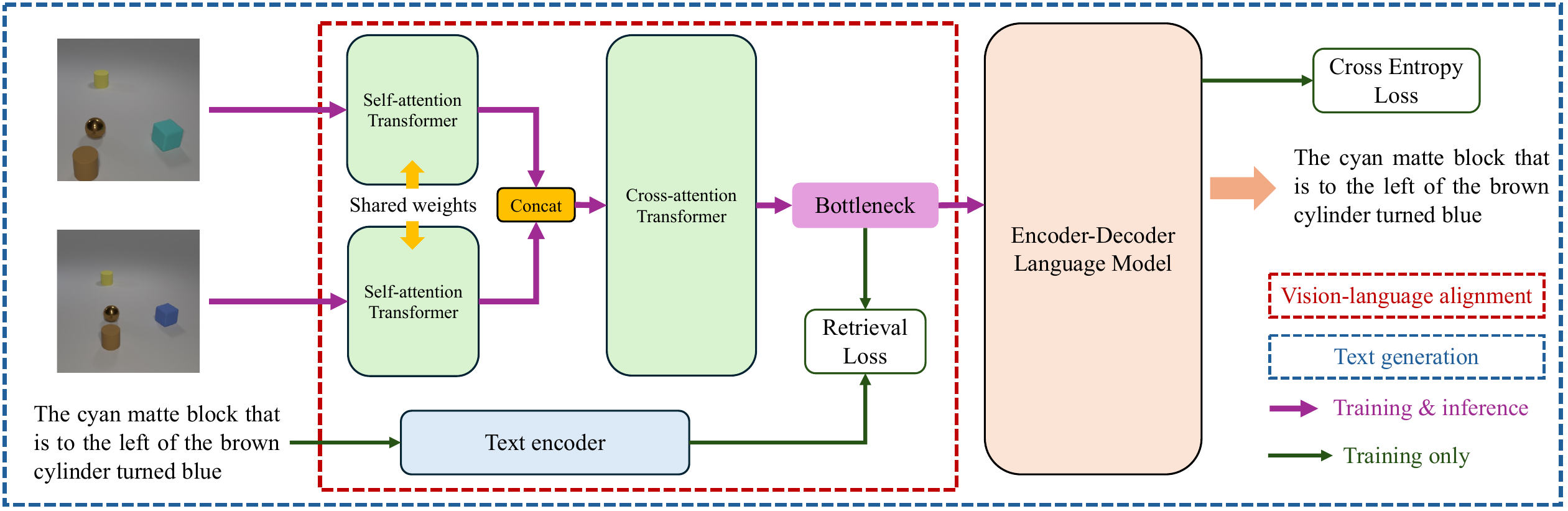}
\caption{We use a two-stage training method for Image Difference Captioning. First, the visual embeddings extracted from the image pair in the bottleneck are aligned with the textual embeddings of the captions. In the second stage, we use a Cross Entropy loss to predict the next token.}
\label{appfig:framework}
\end{figure*}

We adopt our CC framework from \clipforidc \cite{guo2022clip4idc} (see \cref{appfig:framework}). We resize the input images to 224$\times$224. The images are then patchified into 49 patches for \VIThigh and 196 patches for \VITlow using the first convolution layer in CLIP ViT-\VIThigh and ViT-\VITlow, respectively. The intermediate embedding dimension in the vision Transformers is $d=768$ and the final projected embedding is $512$. We use the encoder-decoder language model to perform the next token prediction.

\subsection{Training}
We fix the first convolution layer and use the retrieval loss (\cref{eq:retrieval_loss}) to align vision and language blocks. We use the initial learning rate of $10^{-7}$ and use a cosine scheduler with Adam optimizer. We continue the training in the alignment stage for 12 epochs (see \cref{apptab:alignment}). We drop the retrieval loss and the text encoder from the framework for the text generation stage and connect the pretrained vision Transformer to the language model. We use Cross Entropy loss and leverage the groundtruth box annotations to supervise the activation map in the bottleneck. We train the text generation stage for 50 epochs with Bert's implementation of Adam optimizer settings (see \cref{apptab:captioning}).

\begin{table}[ht]
    \centering
    \caption{Vision-language alignment recipe for \tabforidc.}
    \label{apptab:alignment}
    \begin{tabular}{c|c}
        \toprule
         Config &  Value\\
         \midrule
         Optimizer & Adam\\
         Base LR & 1$e^{-4}$\\
         Scheduler & cosine decay \cite{loshchilov2017sgdr}\\
         Weight decay & 0.2\\
         Momentum & $\beta_1=0.9, \beta_2=0.98$\\
         epsilon & 1$e^{-6}$\\
         Batch size & 128\\
         Warmup proportion & 0.1\\
         Training epochs & 12\\
         \bottomrule
    \end{tabular}
\end{table}

\begin{table}[ht]
    \centering
    \caption{Text generation recipe for \tabforidc.}
    \label{apptab:captioning}
    \begin{tabular}{c|c}
        \toprule
         Config &  Value\\
         \midrule
         Optimizer & Adam\\
         Base LR & 1$e^{-4}$\\
         Scheduler & linear decay\\
         Weight decay & 0.01\\
         Momentum & $\beta_1=0.9, \beta_2=0.999$\\
         epsilon & 1$e^{-6}$\\
         Batch size & 64\\
         Warmup proportion & 0.1\\
         Training epochs & 50\\
         Max words & 32\\
         \bottomrule
    \end{tabular}
\end{table}

\clearpage
\section{Attention visualization details}
\label{appsec:interpolation}

\begin{figure}[H]
    \centering
    \begin{flushleft}
    \hspace{.4cm}\rotatebox{90}{\hspace{-8.4cm}(1) change\hspace{3cm}(2) no change}
    \end{flushleft}
    \resizebox{0.85\columnwidth}{!}{
    \begin{subfigure}[b]{0.24\textwidth}
    \centering
         \includegraphics[width=\linewidth]{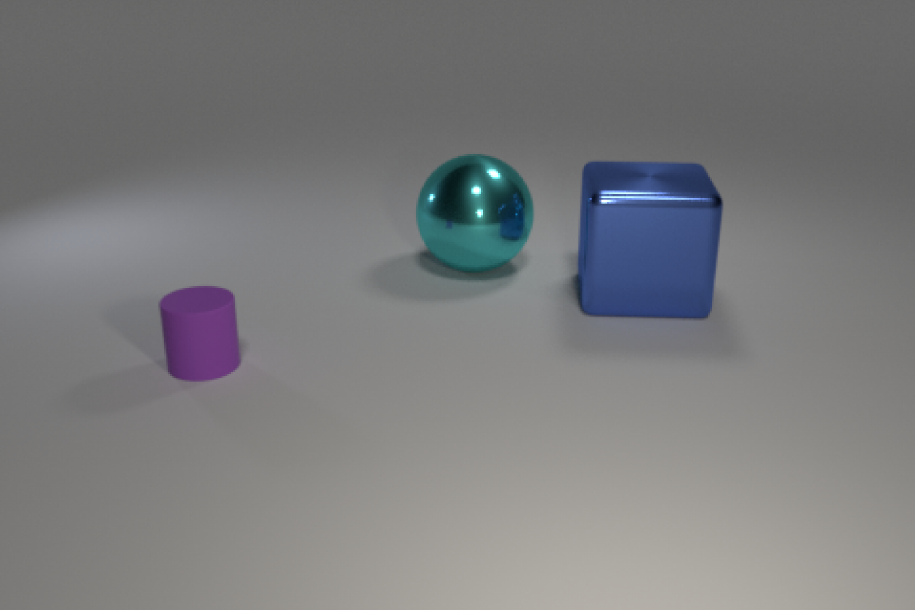}
         \\
         \vspace{-0.2em}
         \includegraphics[width=\linewidth]{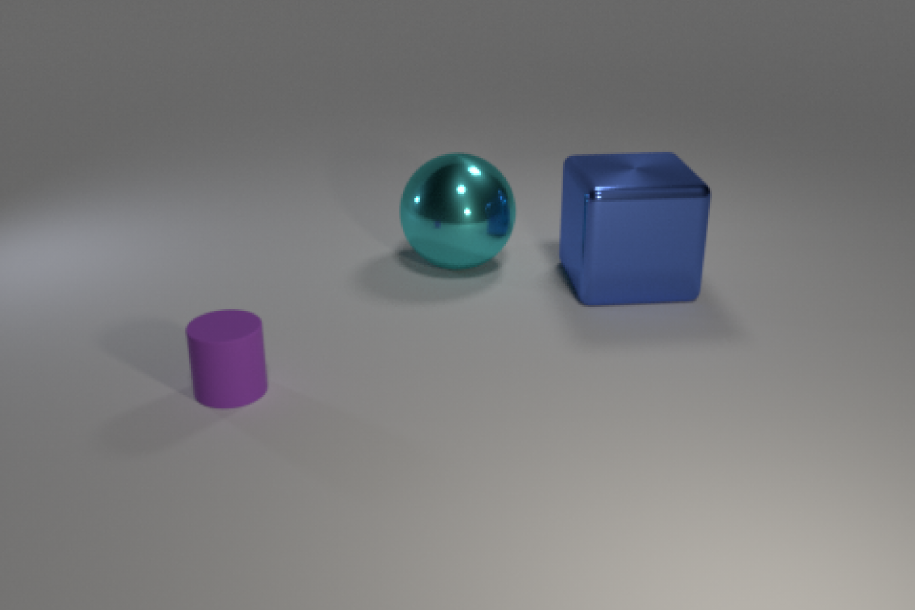}
         \\
         \includegraphics[width=\linewidth]{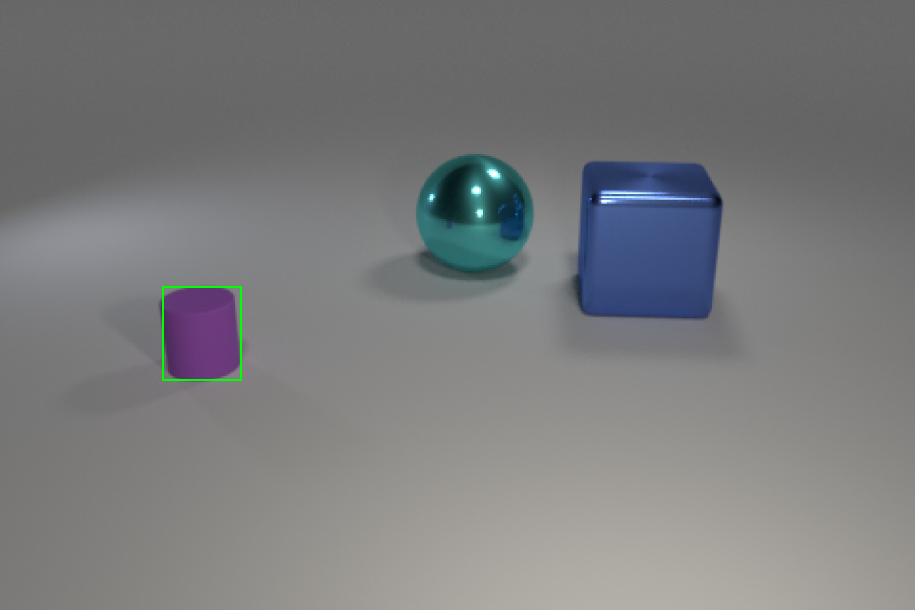}
         \\
         \vspace{-0.2em}
         \includegraphics[width=\linewidth]{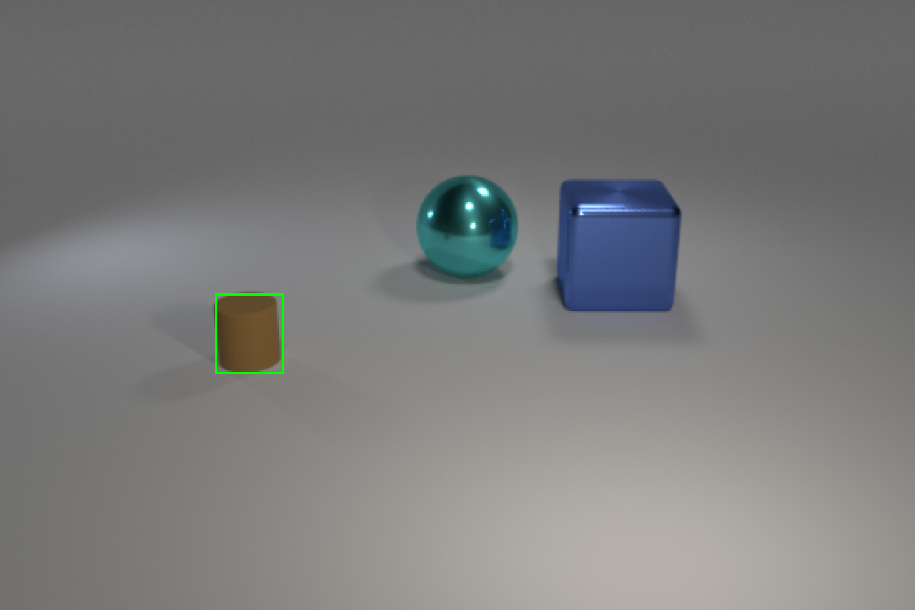}
         \caption{}
    \end{subfigure}
    \begin{subfigure}[b]{0.24\textwidth}
    \centering
         \includegraphics[width=\linewidth]{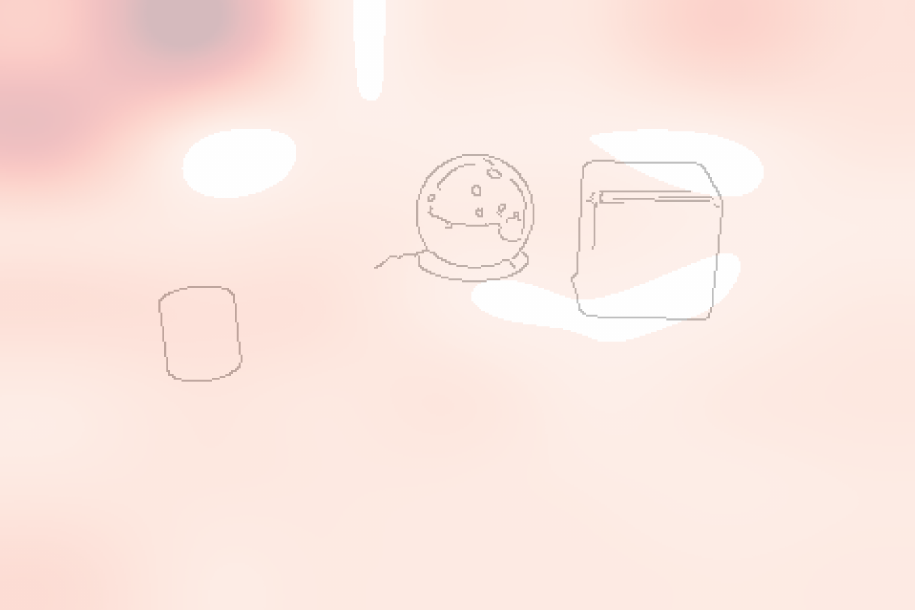}
         \\
         \vspace{-0.2em}
         \includegraphics[width=\linewidth]{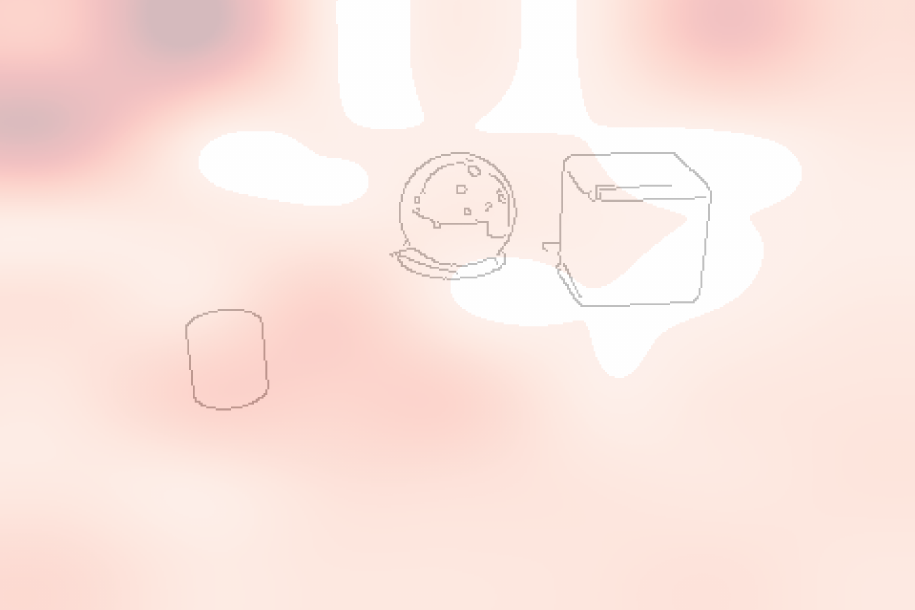}
         \\
         \includegraphics[width=\linewidth]{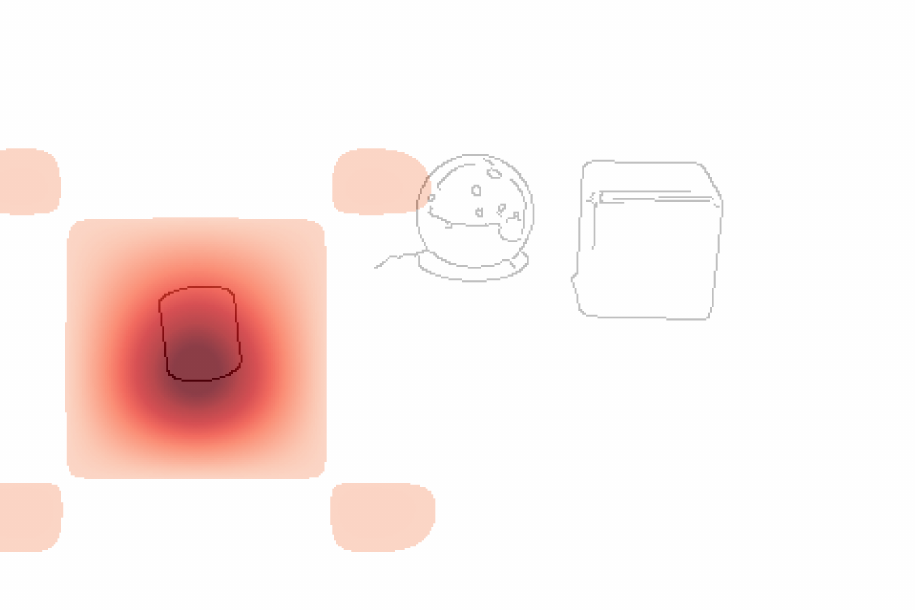}
         \\
         \vspace{-0.2em}
         \includegraphics[width=\linewidth]{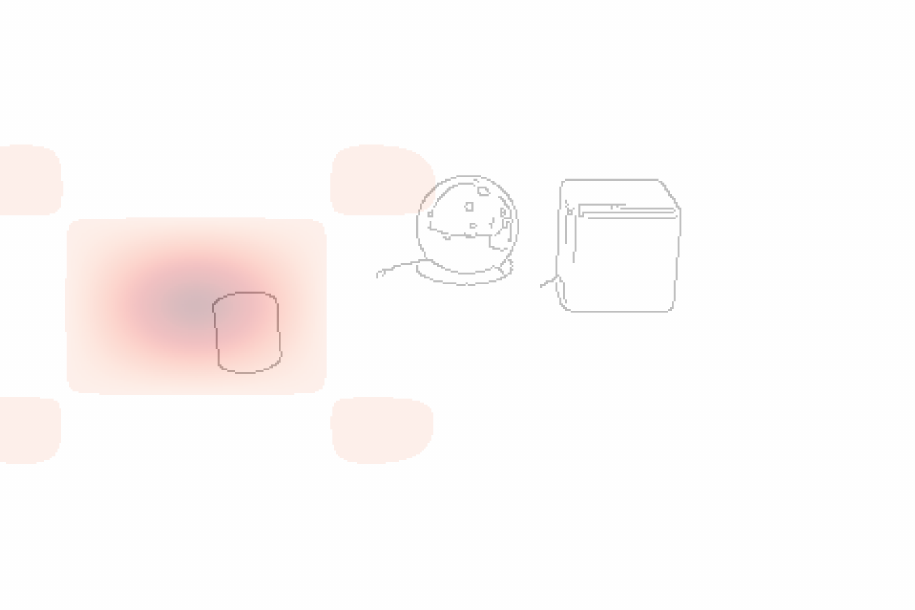}
         \caption{}
    \end{subfigure}
    \begin{subfigure}[b]{0.24\textwidth}
    \centering
         \includegraphics[width=\linewidth]{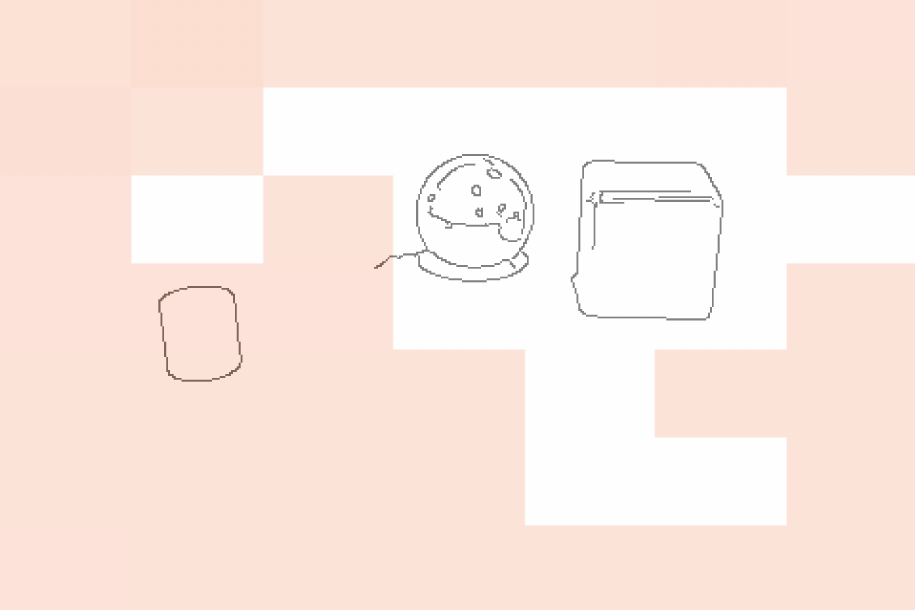}
         \\
         \vspace{-0.2em}
         \includegraphics[width=\linewidth]{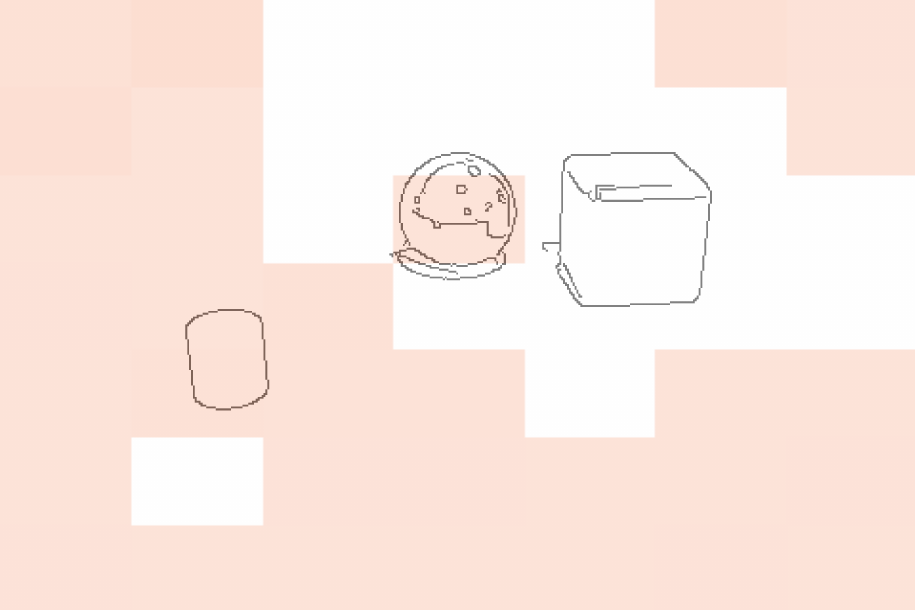}
         \\
         \includegraphics[width=\linewidth]{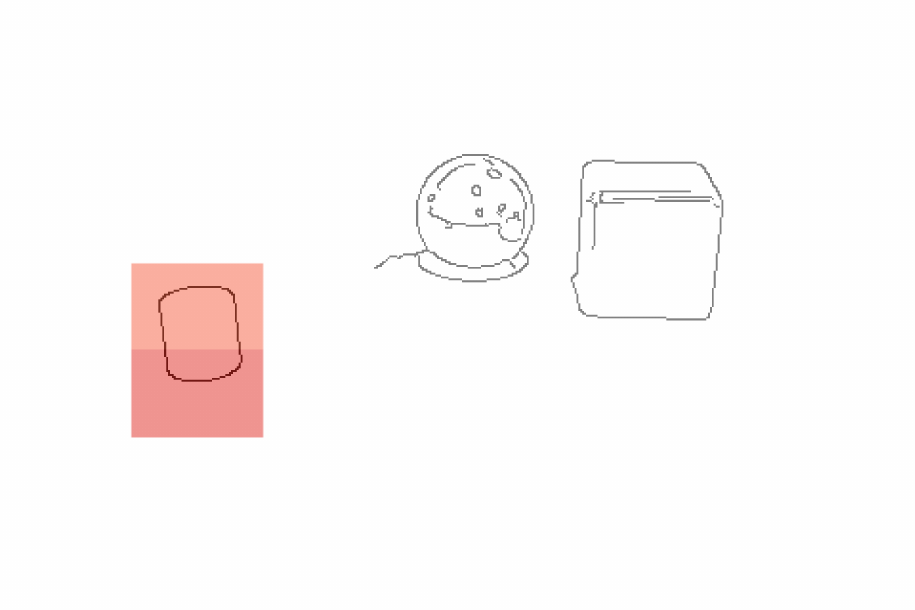}
         \\
         \vspace{-0.2em}
         \includegraphics[width=\linewidth]{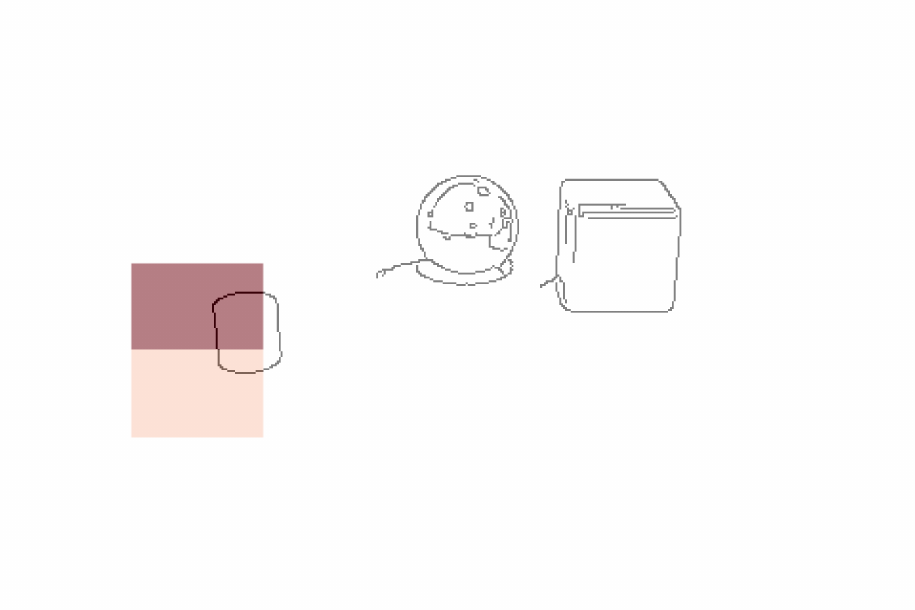}
         \caption{}
    \end{subfigure}
    \begin{subfigure}[b]{0.24\textwidth}
    \centering
         \includegraphics[width=\linewidth]{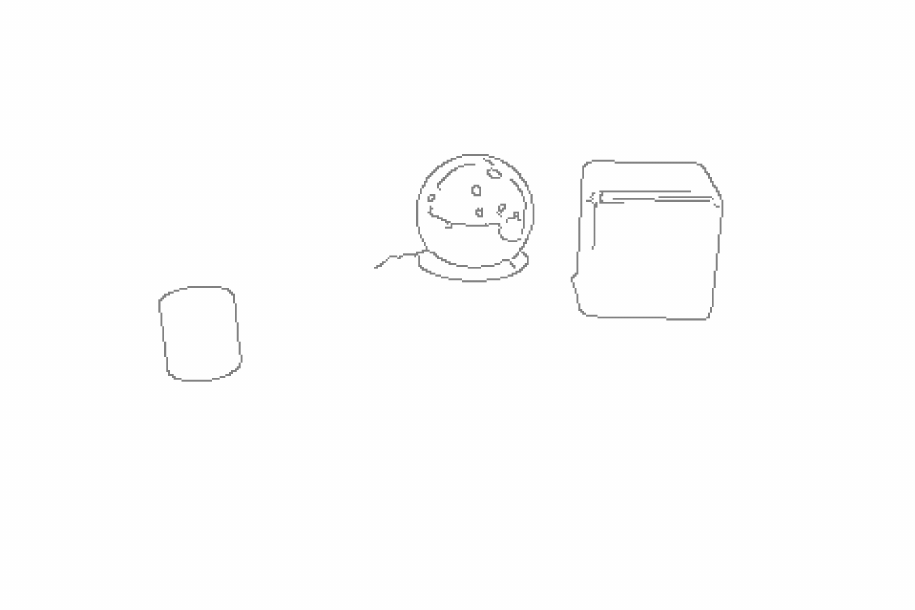}
         \\
         \vspace{-0.2em}
         \includegraphics[width=\linewidth]{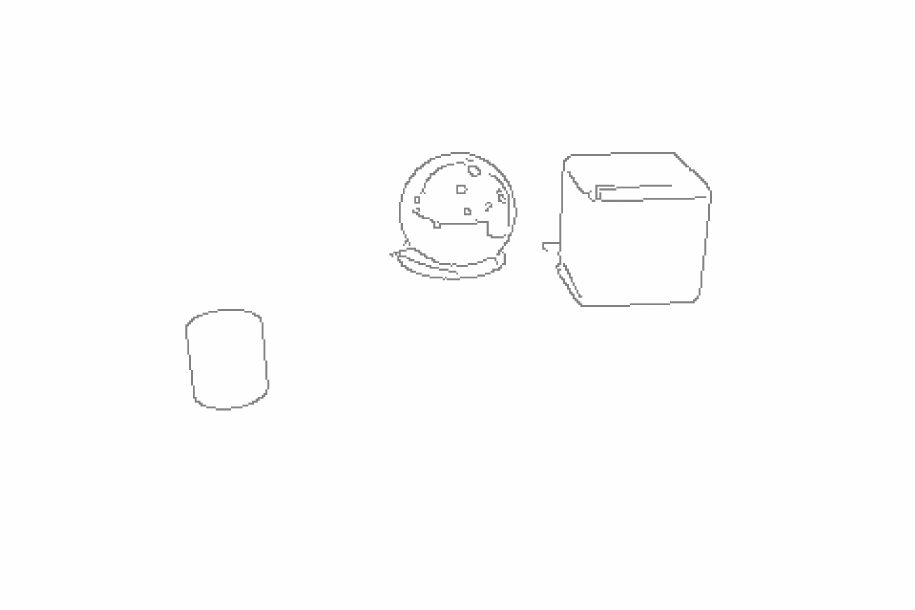}
         \\
         \includegraphics[width=\linewidth]{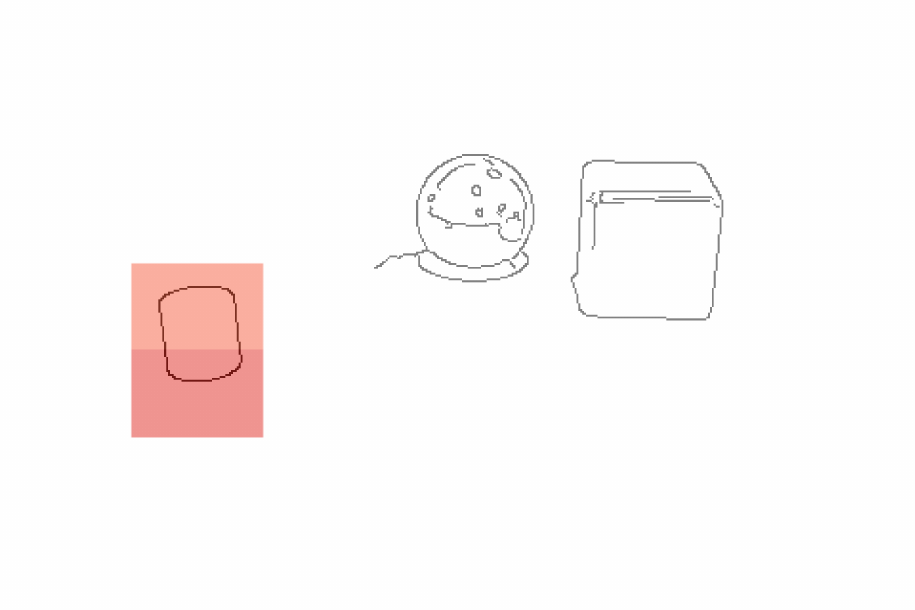}
         \\
         \vspace{-0.2em}
         \includegraphics[width=\linewidth]{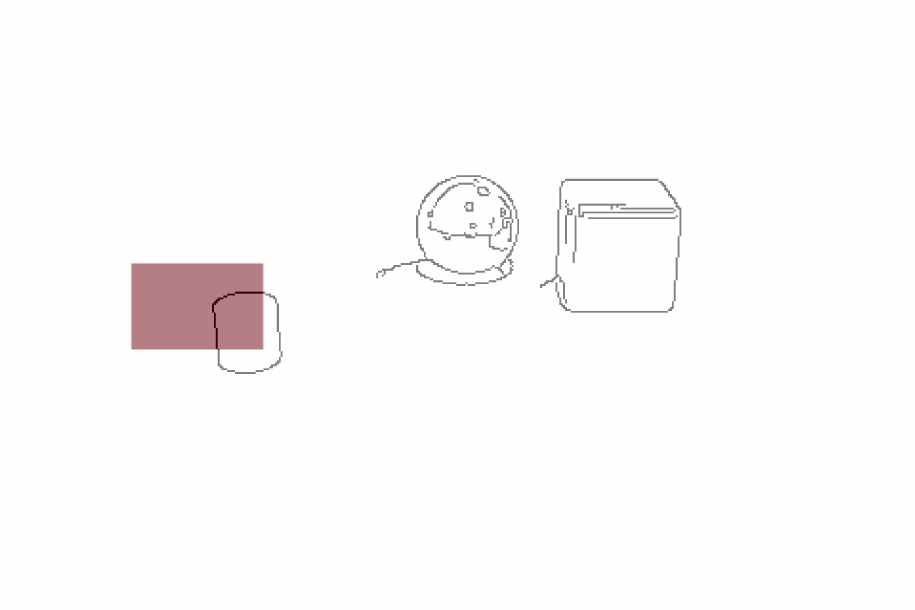}
         \caption{}
    \end{subfigure}}
    \caption{For each input pair in \clevrlogo ~(a), the interpolation method in \clipforidc \cite{guo2022clip4idc} (see \cref{fig:vis_code_org}) yields many nonzero attention values for \class{no\mbox{-}change} pairs (2b). In our visualization, we use the nearest-neighbor method (\cref{fig:vis_code_ours}) that leads to fewer nonzero values on \class{no\mbox{-}change} (2c). For baseline methods, the thresholding substantially improves the attention map's quality for \class{no\mbox{-}change} pairs (2d), where there is no target object for detection.
    }
    \label{fig:our-vis}
\end{figure}

\begin{adjustbox}{width=\textwidth}
    \centering
    \begin{lstlisting}[language=Python]
import cv2
import numpy as np

def resize_map(attention_map=None, input_size=(480, 320)):
    resized_map = cv2.resize(attention_map.astype(np.uint8), input_size, interpolation=cv2.INTER_CUBIC)
    
    return resized_map

\end{lstlisting}
\end{adjustbox}
\captionof{figure}{Prior works \cite{park2019robust,guo2022clip4idc} use cubic interpolation to resize the attention maps that lead to smooth edges in the heatmap. Yet, it also results in peak values over image patches that have near zero attention values.}
\label{fig:vis_code_org}

\medskip

\begin{adjustbox}{width=\textwidth}
    \centering
    \begin{lstlisting}[language=Python]
import cv2
import numpy as np

def resize_map(attention_map=None, input_size=(480, 320)):

    resized_map = cv2.resize(attention_map.astype(np.float32), input_size, interpolation=cv2.INTER_NEAREST)

    return resized_map
    
\end{lstlisting}
\end{adjustbox}
\captionof{figure}{In our visualization paradigm, we replace the interpolation method with the nearest neighbor such that the resized attention has fewer nonzero values.}
\label{fig:vis_code_ours}

\clearpage
\section{Generating groundtruth captions for \openIn}
\label{appsec:caption_generation}

We have access to all the inpainted object names in \openIn \cite{nguyenWACVChange}, and we generate multiple sentences describing a particular change using templates listed in \cref{apptab:temp_captions}.

\begin{table}[hb]
    \centering
    \begin{tabular}{l|l}
    \toprule
         Pair type &  Caption template\\
         \midrule
         \multirow{3}{*}{Add} & the ... has appeared\\
         & the ... has been newly placed\\
         & the ... has been added\\
         \midrule
         \multirow{4}{*}{Drop} & the ... has disappeared\\
         & the ... is missing\\
         & the ... is gone\\
         & the ... is no longer there\\
         \midrule
         \multirow{9}{*}{\class{No\mbox{-}change}} & no change was made \\
         & there is no change \\
         & the two scenes seem identical \\
         & the scene is the same as before \\
         & the scene remains the same \\
         & nothing has changed \\
         & nothing was modified \\
         & no change has occurred \\
         & there is no difference\\
         \bottomrule
    \end{tabular}
    \caption{We follow \cite{park2019robust} and use the caption templates to generate groundtruth change captions.}
    \label{apptab:temp_captions}
\end{table}

\clearpage
\section{\Editing the attention values in MHSA layer does not yield a different caption}
\label{appsec:abledit}
To further investigate the role of our proposed 1-head attention in the \textcolor{Periwinkle}{intervention} task, we also supervise the MHSA attention in \clipforidc during the training and evaluate it similarly to \cref{sec:editability}. Compared to \tab, the MHSA layer does not yield a cause-and-effect relation with the predictions \cref{apptab:editmhsa}, perhaps due to information leakage in the architecture. 

\begin{table}[h]
    \centering
    \caption{Supervising 12 heads in the MHSA layer of \clipforidc does not enable (\redxmark) user intervention as in \tab (\greentick)}
    \label{apptab:editmhsa}
    \resizebox{0.85\columnwidth}{!}{
    \begin{tabular}{c@{}cccrcrcr}
    \toprule
         &&&\multicolumn{2}{c}{Acc. Change} & \multicolumn{2}{c}{Acc. No\mbox{-}change}& \multicolumn{2}{c}{Acc. object name}\\
         \cmidrule(rr){4-5}
         \cmidrule(rr){6-7}
         \cmidrule(lr){8-9}
         &Dataset & Attention \editlogo & base & \editlogo & base &  \editlogo & base & \editlogo\\
         \midrule
         \multirow{2}{*}{MHSA}&\multirow{4}{*}{\openlogo}  & \cellcolor{lightgray}\textsc{zero} & \cellcolor{lightgray}99.97 &\cellcolor{lightgray}99.97 \redxmark & \cellcolor{lightgray}100.0 & \cellcolor{lightgray}100.0& \cellcolor{lightgray}89.78 & \cellcolor{lightgray}89.76 \redxmark\\
         && \textsc{correct} $\uparrow$ & 99.97& 99.97 \redxmark&100.0& 100.0 &89.78& 89.80 \greentick\\
         \cmidrule(rr){3-9}
         \multirow{2}{*}{\tab}&& \cellcolor{lightgray}\textsc{zero} & \cellcolor{lightgray}99.93 &\cellcolor{lightgray} 0.0 \greentick & \cellcolor{lightgray}100.0 & \cellcolor{lightgray}100.0 \greentick& \cellcolor{lightgray}88.92 & \cellcolor{lightgray}0.0 \greentick\\
         && \textsc{correct} $\uparrow$ & 99.93& 100.0 \greentick&100.0& 100.0 \greentick &88.92& 91.49 \greentick\\
         \midrule
         \multirow{2}{*}{MHSA}&\multirow{4}{*}{\clevrlogo} & \cellcolor{lightgray}\textsc{zero}& \cellcolor{lightgray}95.00 & \cellcolor{lightgray}95.0 \redxmark&\cellcolor{lightgray} 99.17 & \cellcolor{lightgray}99.17 \redxmark& \cellcolor{lightgray}- & \cellcolor{lightgray}-\\
         && \textsc{correct} $\uparrow$ & 95.00& 95.0 \redxmark&99.17& 99.17 \redxmark&-& -\\
         \cmidrule(rr){3-9}
         \multirow{2}{*}{\tab}&& \cellcolor{lightgray}\textsc{zero}& \cellcolor{lightgray}94.30 & \cellcolor{lightgray}0.0 \greentick&\cellcolor{lightgray} 99.42 & \cellcolor{lightgray}100.0 \greentick& \cellcolor{lightgray}- & \cellcolor{lightgray}-\\
         && \textsc{correct} $\uparrow$ & 94.30& 100.0 \greentick&99.42& 100.0 \greentick&-& -\\
         \bottomrule
    \end{tabular}}
\end{table}

\clearpage
\section{\std additional results}
\label{appsec:std_add}

\subsection{Localization}
\label{appsec:std_loc}

\begin{figure}[H]
    \centering
    \begin{subfigure}[b]{0.2\textwidth}
    \centering
         \includegraphics[width=\linewidth]{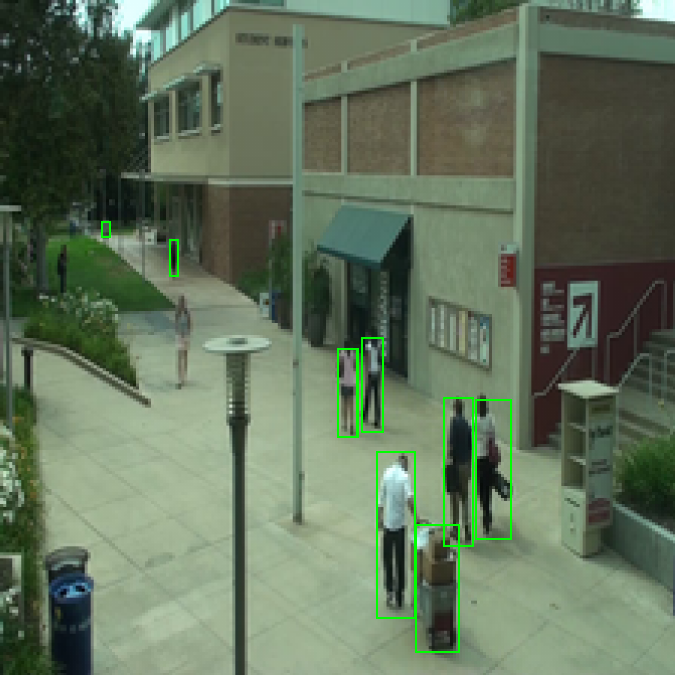}
         \\
         \vspace{-0.1em}
         \includegraphics[width=\linewidth]{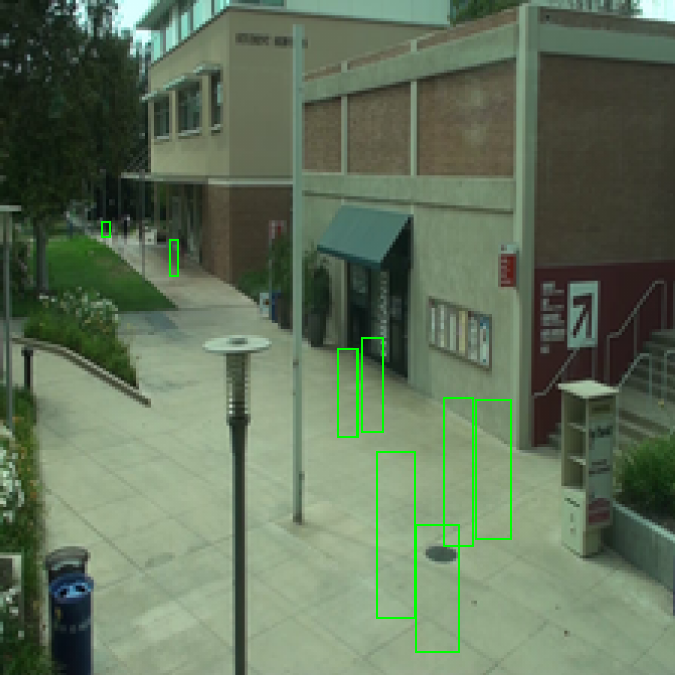}
         \caption{}
    \end{subfigure}
    \begin{subfigure}[b]{0.2\textwidth}
    \centering
         \includegraphics[width=\linewidth]{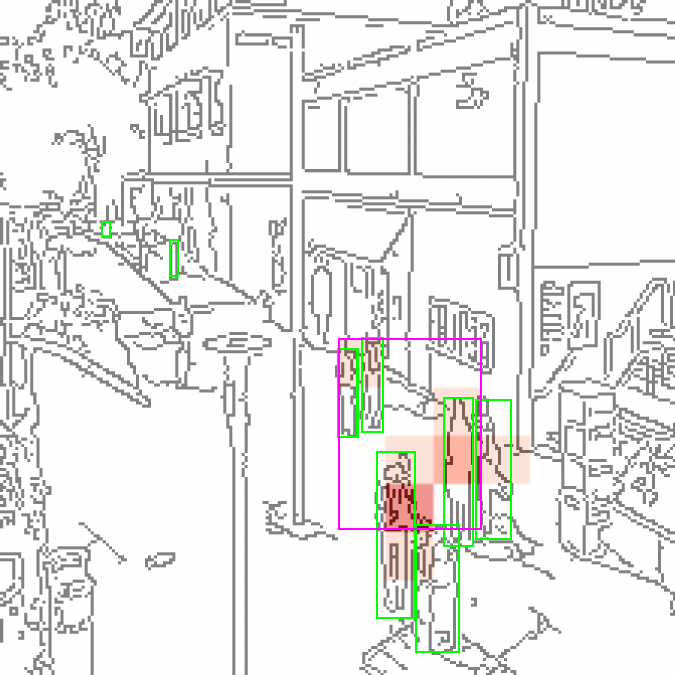}
         \\
         \vspace{-0.1em}
         \includegraphics[width=\linewidth]{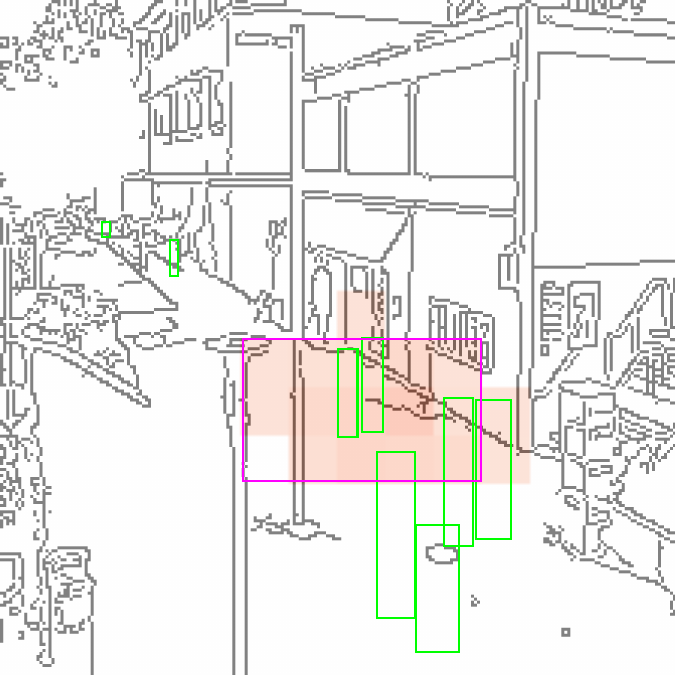}
         \caption{}
    \end{subfigure}
    \hfill
    \begin{subfigure}[b]{0.2\textwidth}
    \centering
         \includegraphics[width=\linewidth]{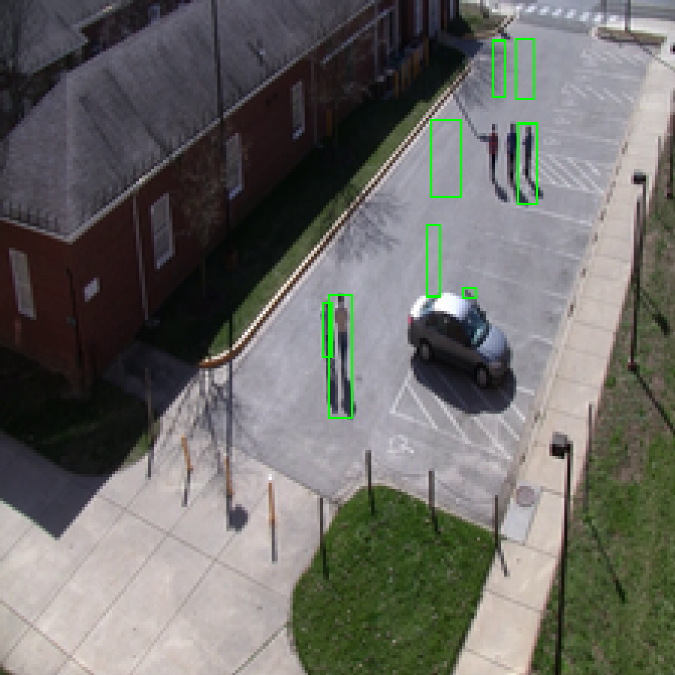}
         \\
         \vspace{-0.1em}
         \includegraphics[width=\linewidth]{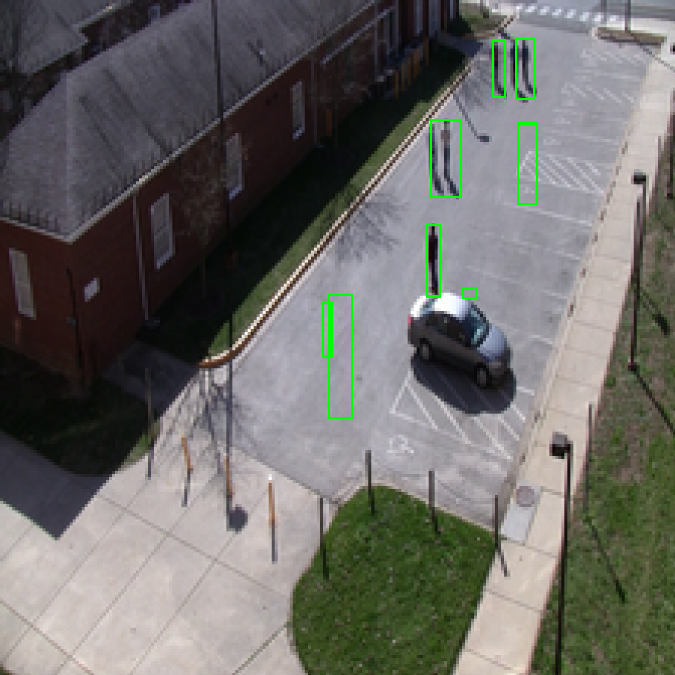}
         \caption{}
    \end{subfigure}
    \begin{subfigure}[b]{0.2\textwidth}
    \centering
         \includegraphics[width=\linewidth]{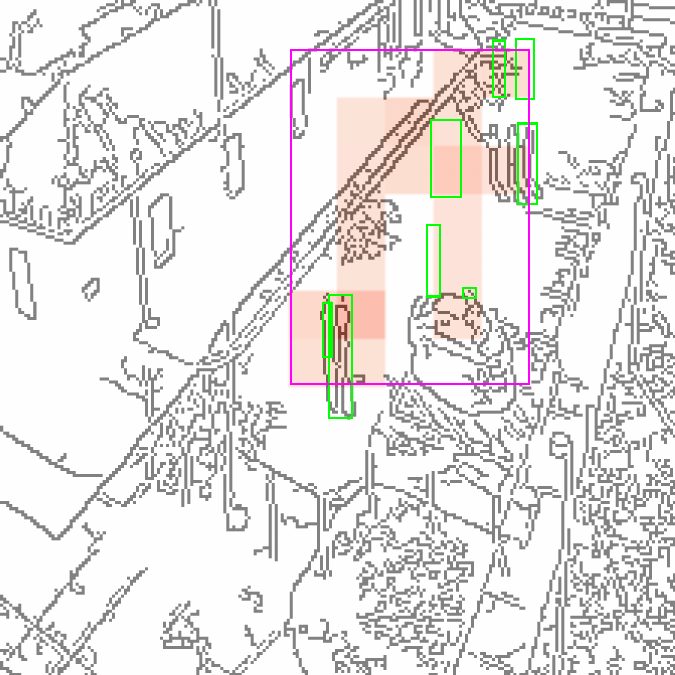}
         \\
         \vspace{-0.1em}
         \includegraphics[width=\linewidth]{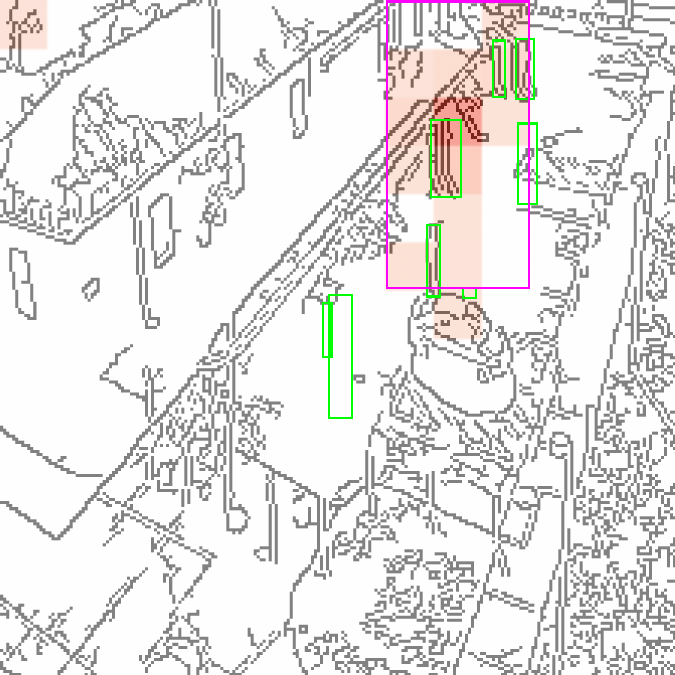}
         \caption{}
    \end{subfigure}
    \begin{subfigure}[b]{0.2\textwidth}
    \centering
         \includegraphics[width=\linewidth]{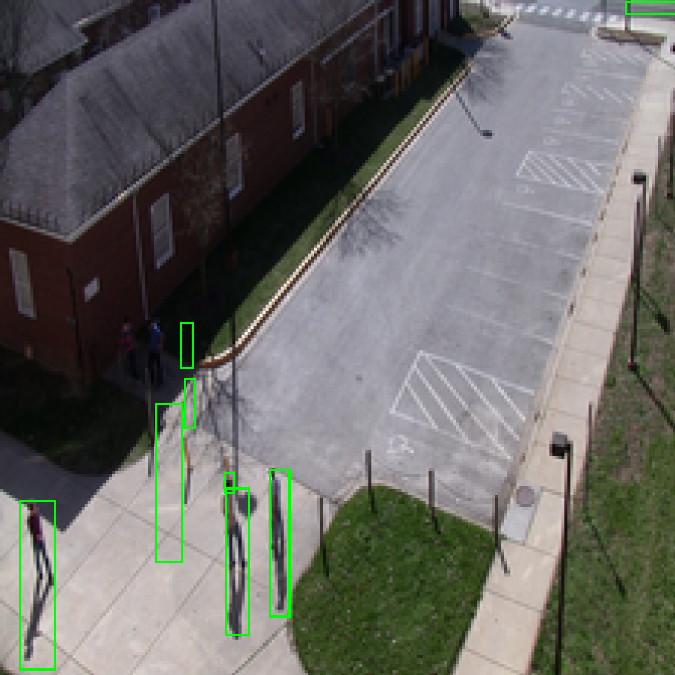}
         \\
         \vspace{-0.1em}
         \includegraphics[width=\linewidth]{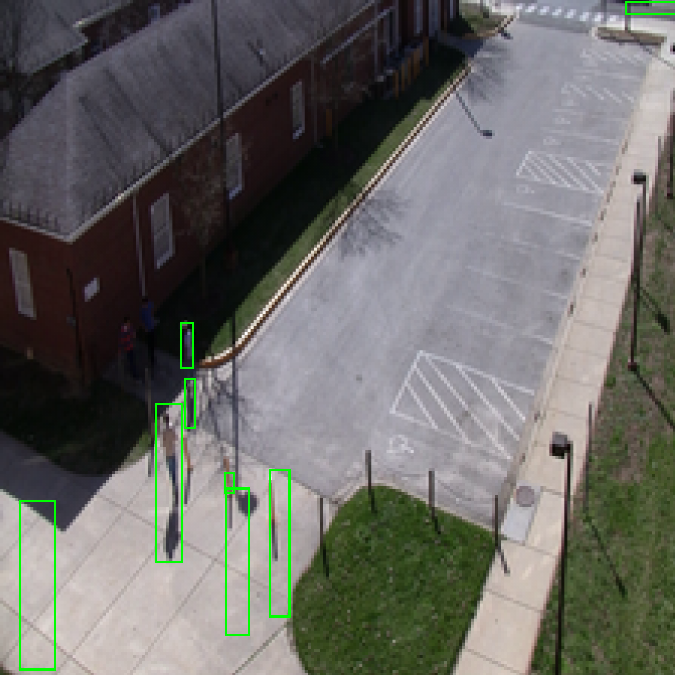}
         \caption{}
    \end{subfigure}
    \begin{subfigure}[b]{0.2\textwidth}
    \centering
         \includegraphics[width=\linewidth]{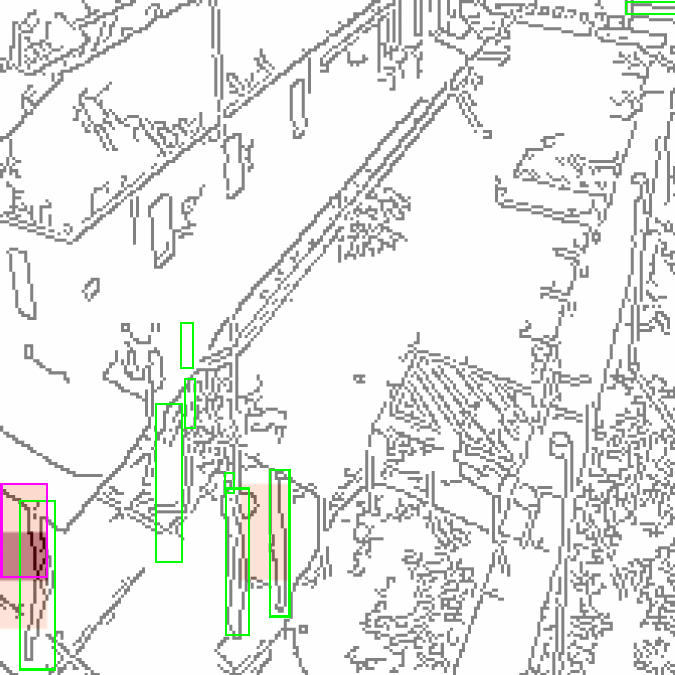}
         \\
         \vspace{-0.1em}
         \includegraphics[width=\linewidth]{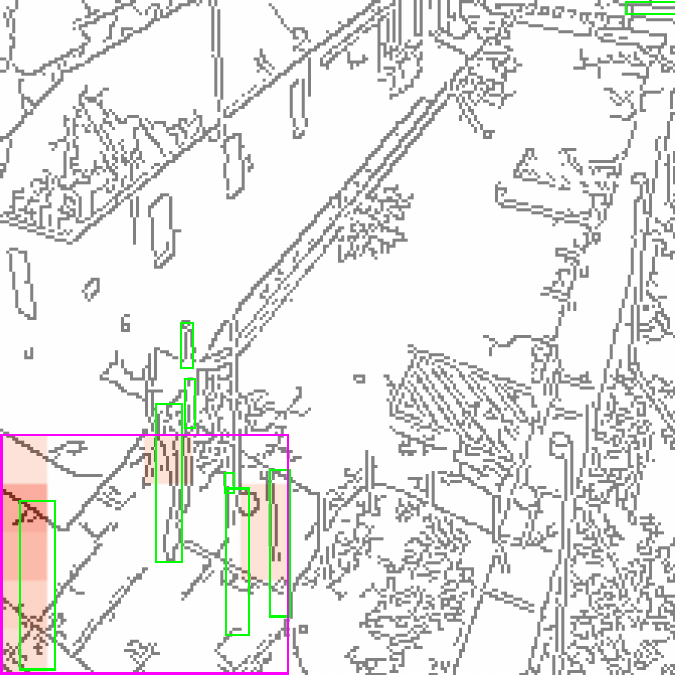}
         \caption{}
    \end{subfigure}
    \hfill
    \begin{subfigure}[b]{0.2\textwidth}
    \centering
         \includegraphics[width=\linewidth]{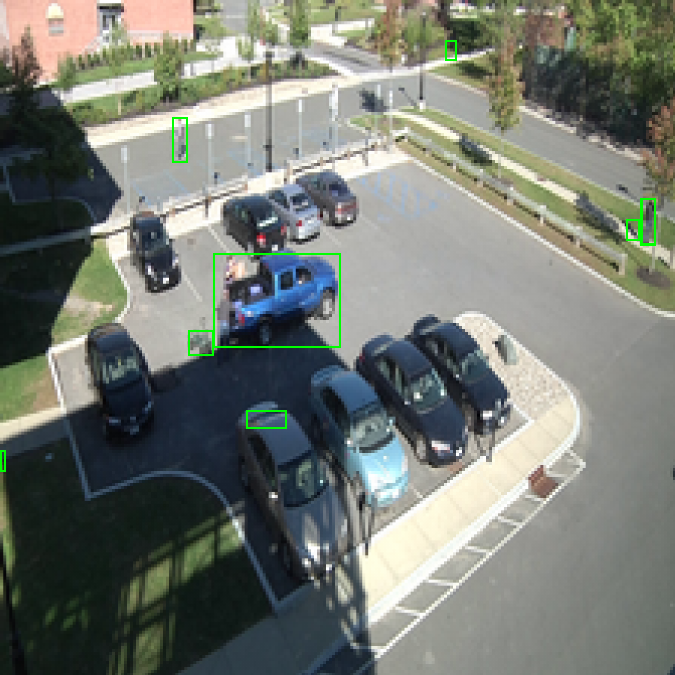}
         \\
         \vspace{-0.1em}
         \includegraphics[width=\linewidth]{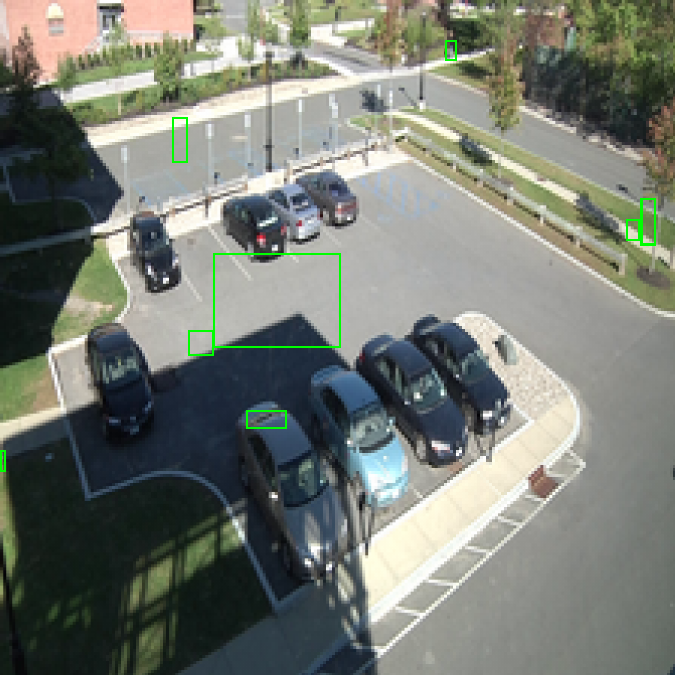}
         \caption{}
    \end{subfigure}
    \begin{subfigure}[b]{0.2\textwidth}
    \centering
         \includegraphics[width=\linewidth]{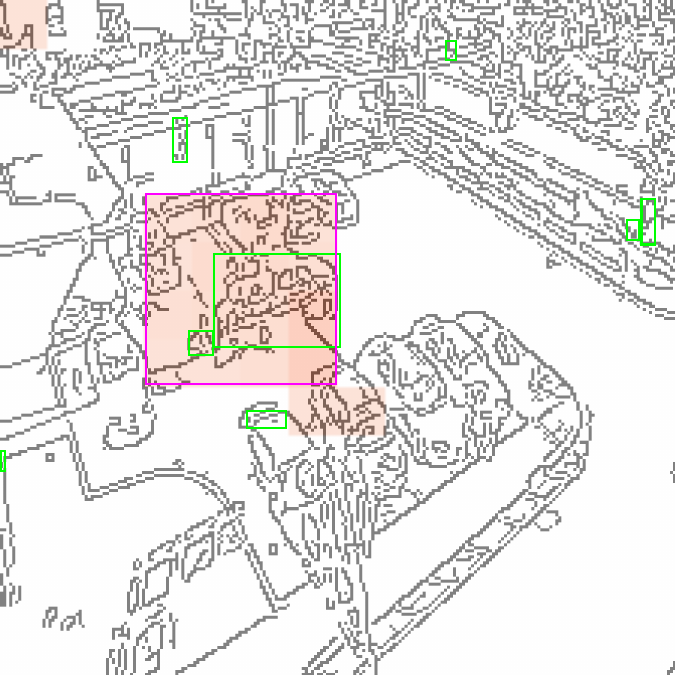}
         \\
         \vspace{-0.1em}
         \includegraphics[width=\linewidth]{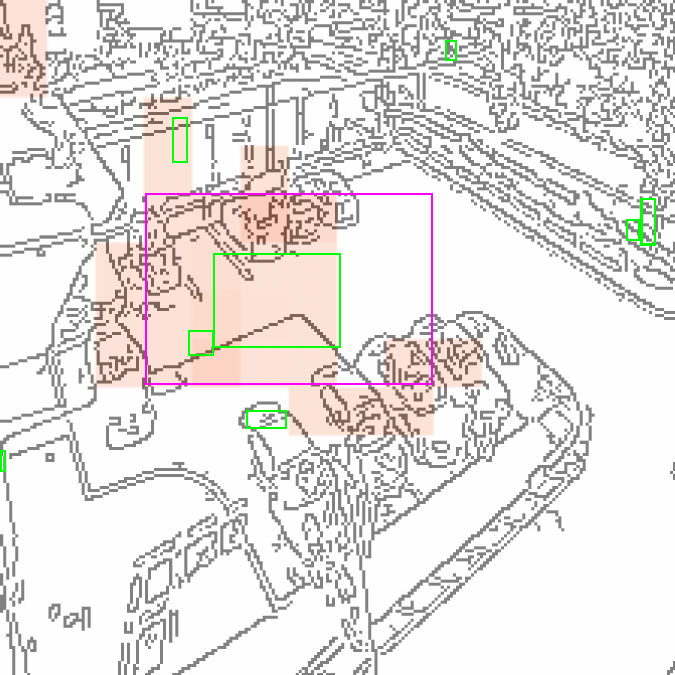}
         \caption{}
    \end{subfigure}
    \caption{\tab's change localization on \std with \VITlow. \std contains images with multiple changes, which naturally leads to lower values in the attention maps.}
    \label{appfig:std_zero_loc}
\end{figure}

\clearpage
\section{\openIn additional results}
\label{appsec:open_add}

\subsection{Correcting the attention map for change pairs}
\label{appsec:open_crtatt}

\begin{figure}[H]
    \centering
    \includegraphics[width=0.7\linewidth]{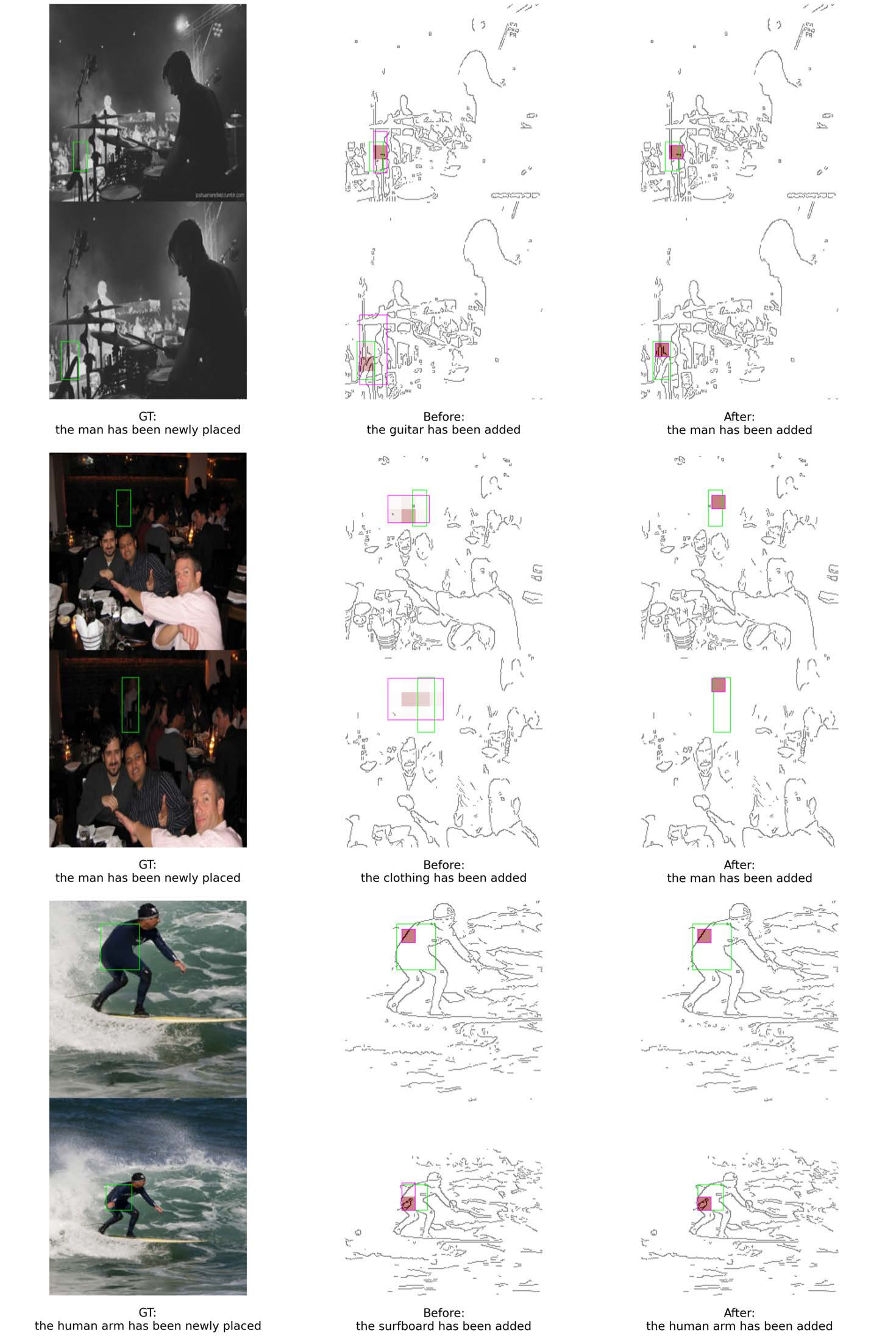}
    \caption{Editing the attention map in \tab with \VITlow helps the VLM to caption the changes more accurately.}
    \label{appfig:op_edit_0}
\end{figure}
\begin{figure}[H]
    \centering
    \includegraphics[width=0.7\linewidth]{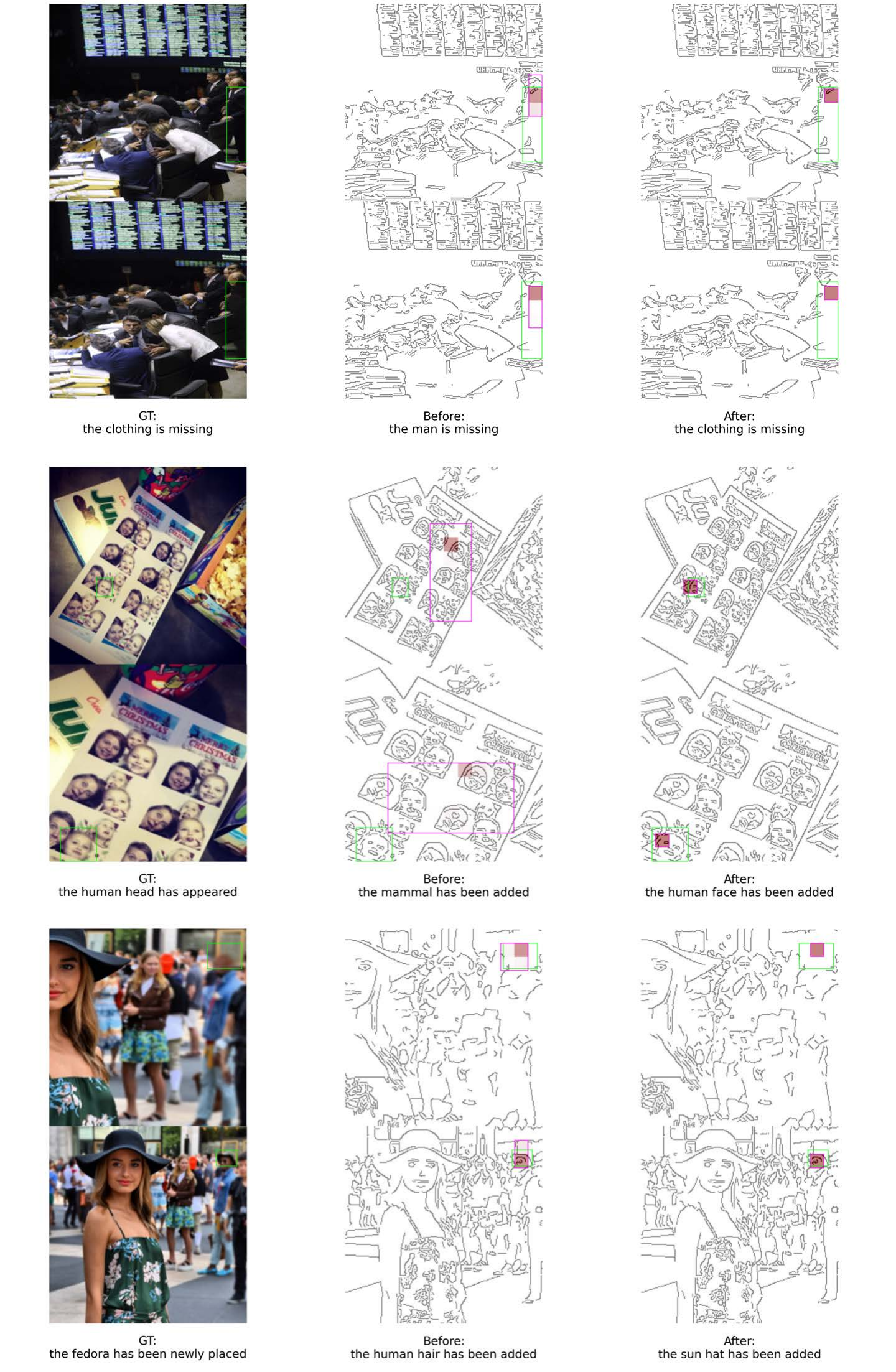}
    \caption{Editing the attention map in \tab with \VITlow}
    \label{appfig:op_edit_1}
\end{figure}
\begin{figure}[H]
    \centering
    \includegraphics[width=0.7\linewidth]{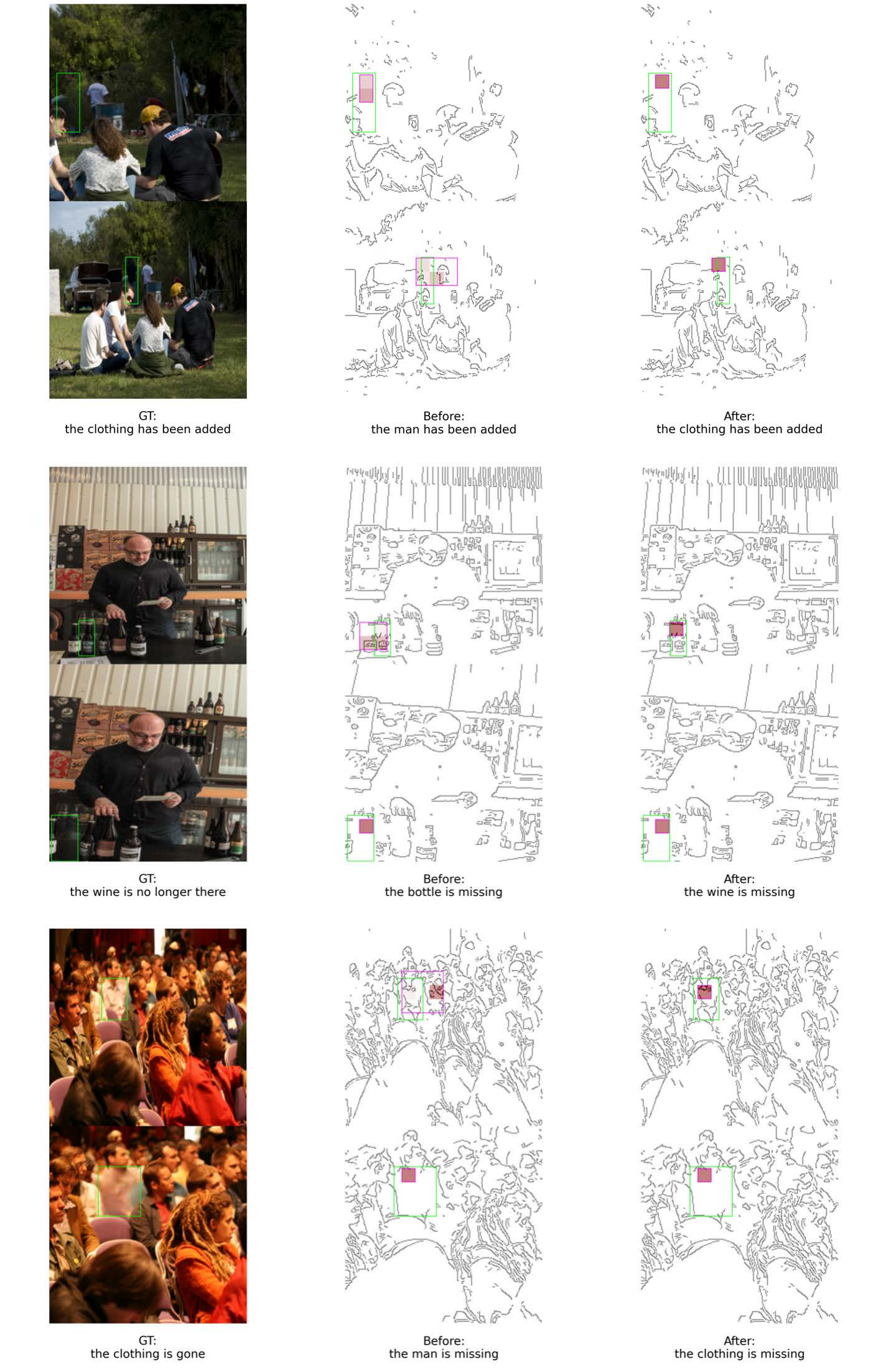}
    \caption{Editing the attention map in \tab with \VITlow}
    \label{appfig:op_edit_2}
\end{figure}

\begin{figure}[H]
    \centering
    \includegraphics[width=0.7\linewidth]{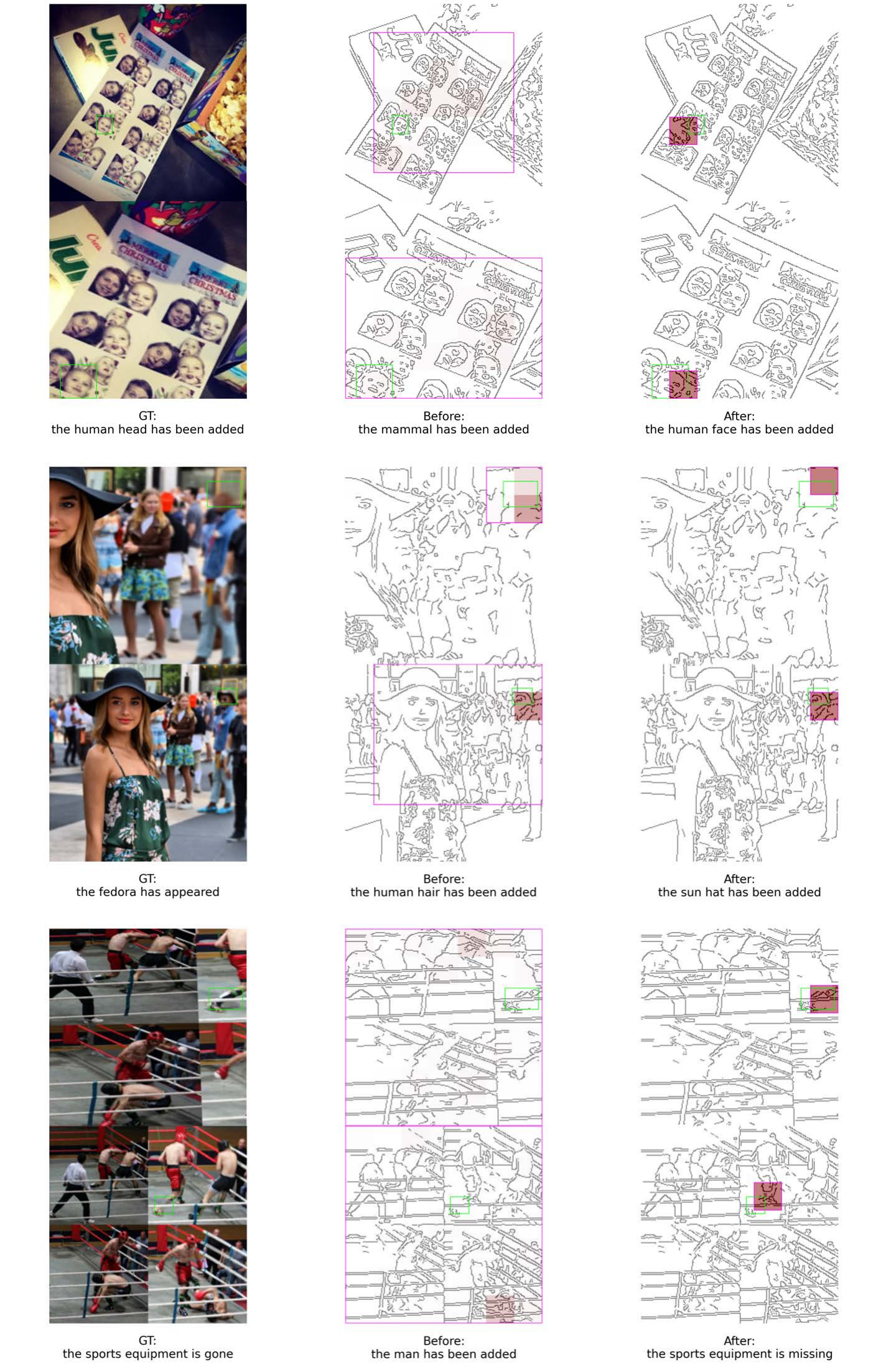}
    \caption{Editing the attention map in \tab with \VIThigh}
    \label{appfig:op_edit_3}
\end{figure}
\begin{figure}[H]
    \centering
    \includegraphics[width=0.7\linewidth]{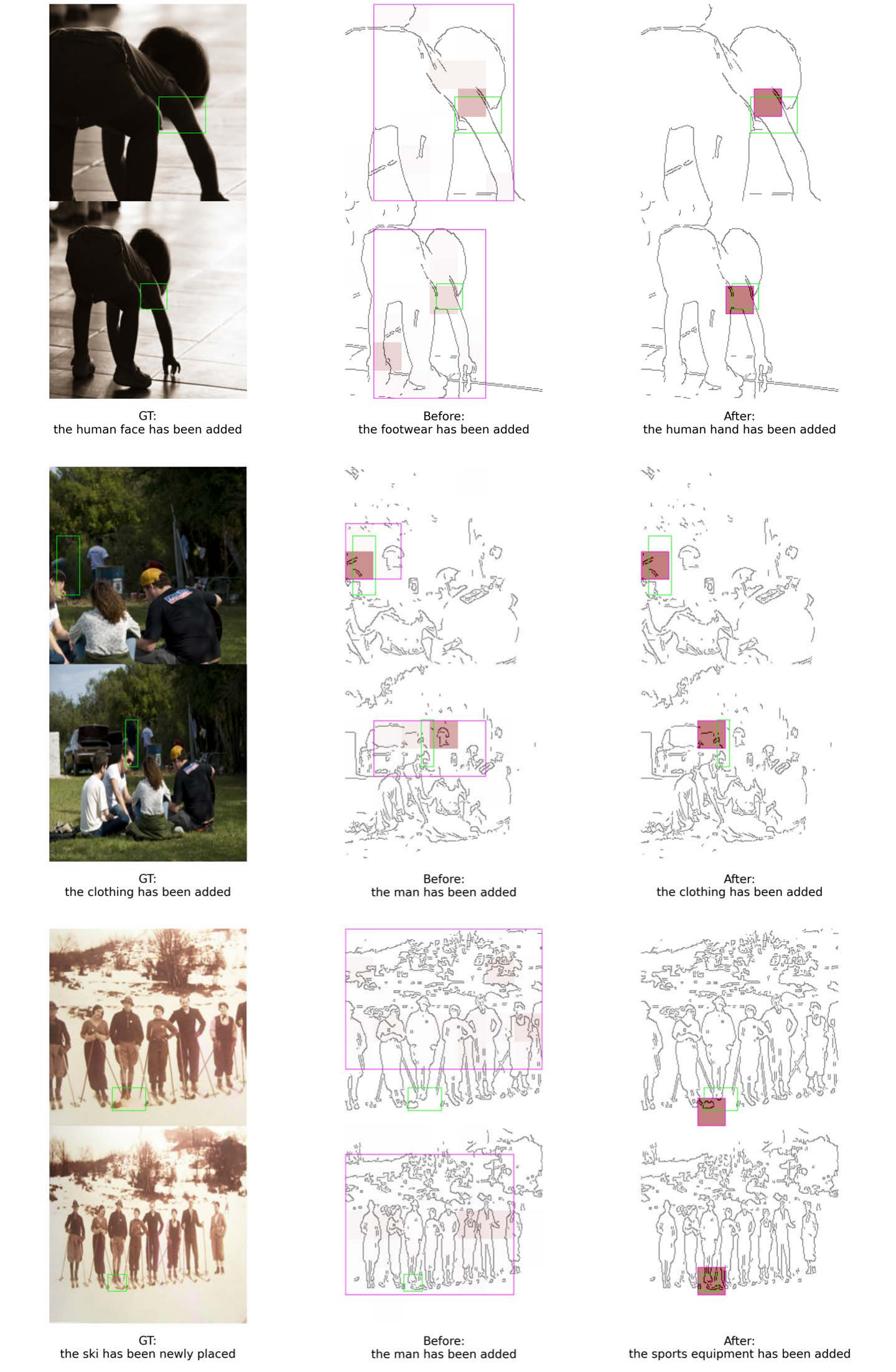}
    \caption{Editing the attention map in \tab with \VIThigh}
    \label{appfig:op_edit_4}
\end{figure}
\begin{figure}[H]
    \centering
    \includegraphics[width=0.7\linewidth]{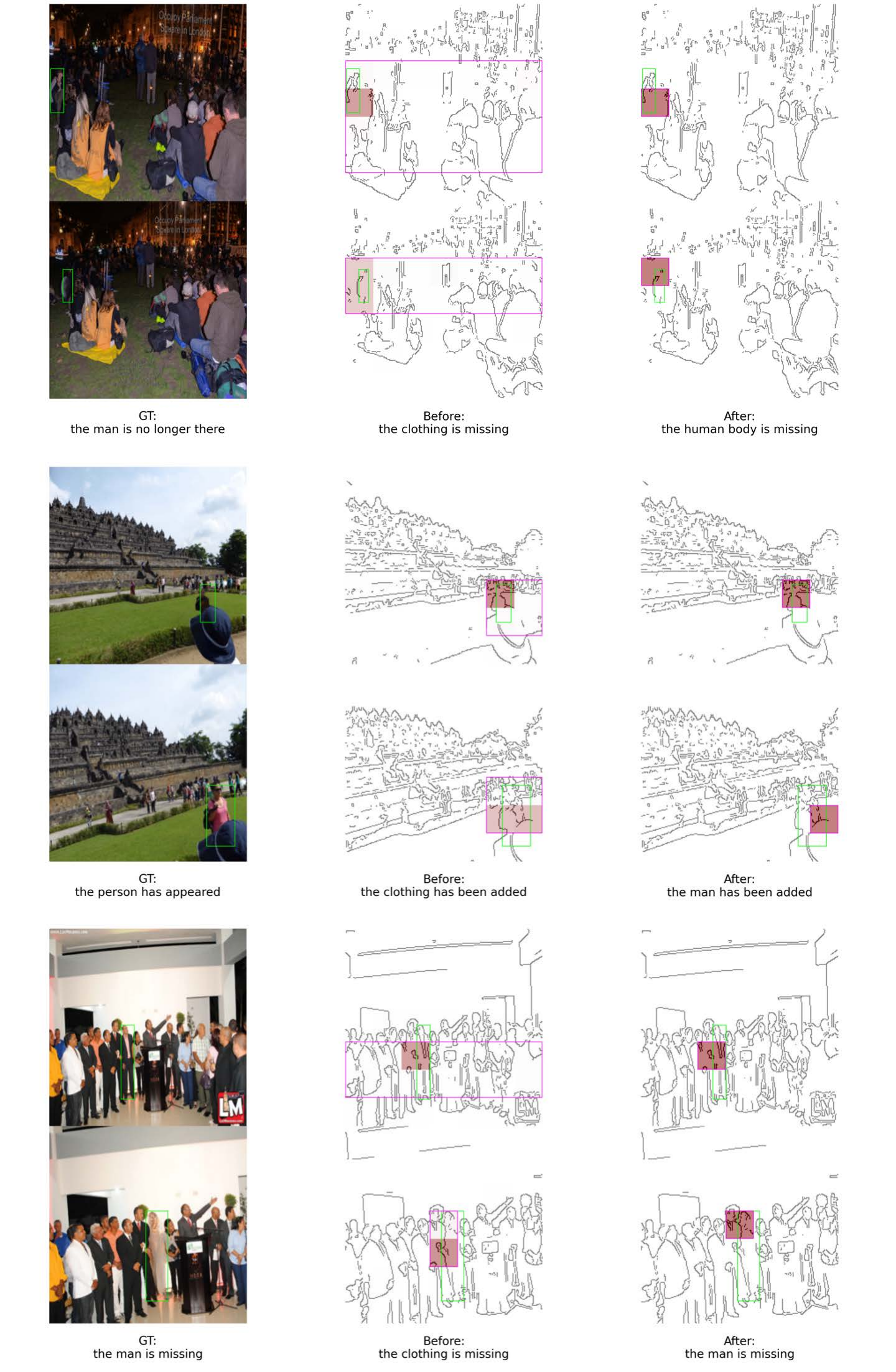}
    \caption{Editing the attention map in \tab with \VIThigh.}
    \label{appfig:op_edit_5}
\end{figure}

\clearpage
\section{\clevr additional results}
\label{appsec:clevr_add}

\subsection{Captioning and Localization}
\label{appsec:clevr_cap}

\begin{figure}[H]
    \centering
    \includegraphics[width=0.7\linewidth]{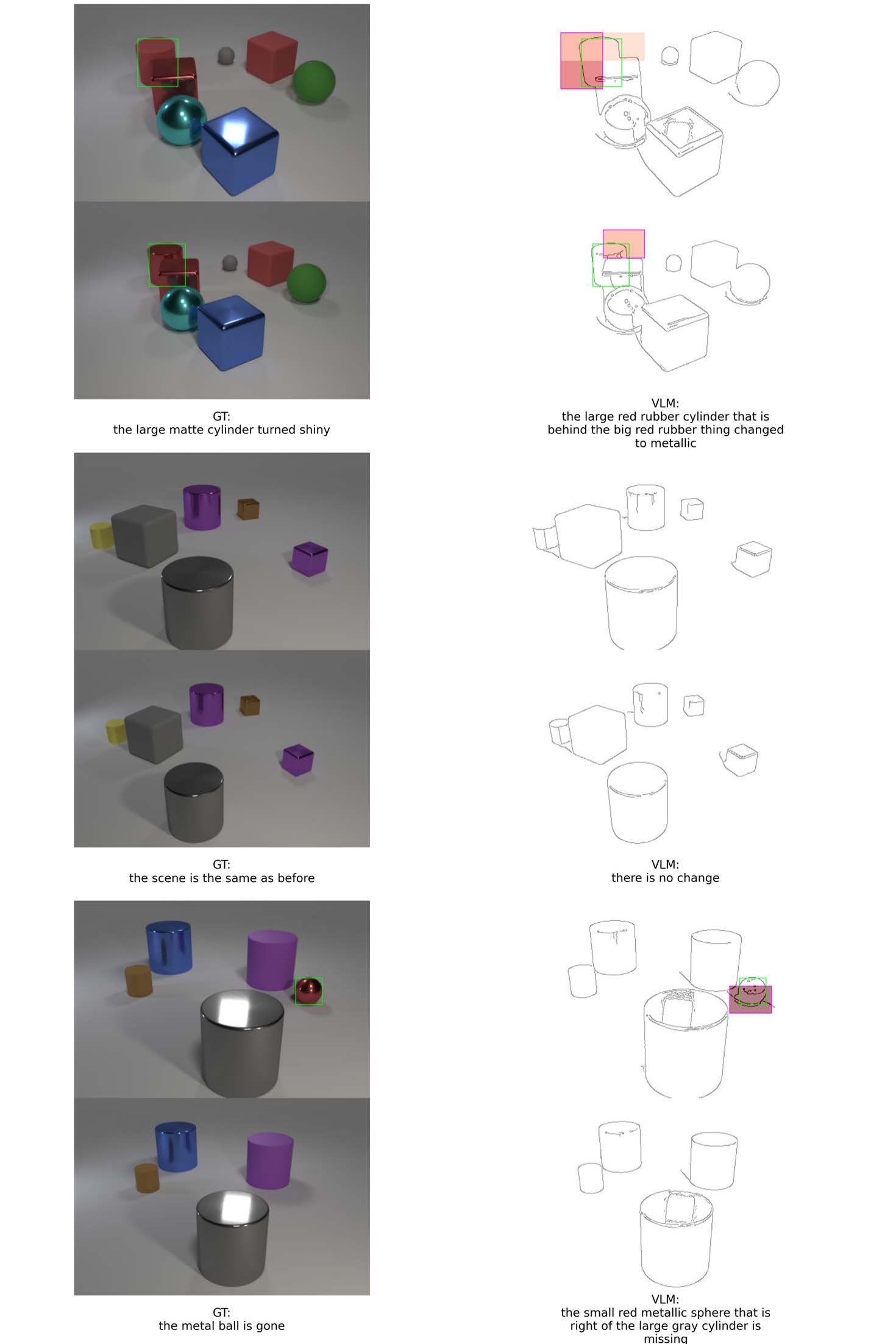}
    \caption{Captioning: \tabforidc with \VIThigh}
    \label{appfig:cl_cap_0}
\end{figure}
\begin{figure}[H]
    \centering
    \includegraphics[width=0.7\linewidth]{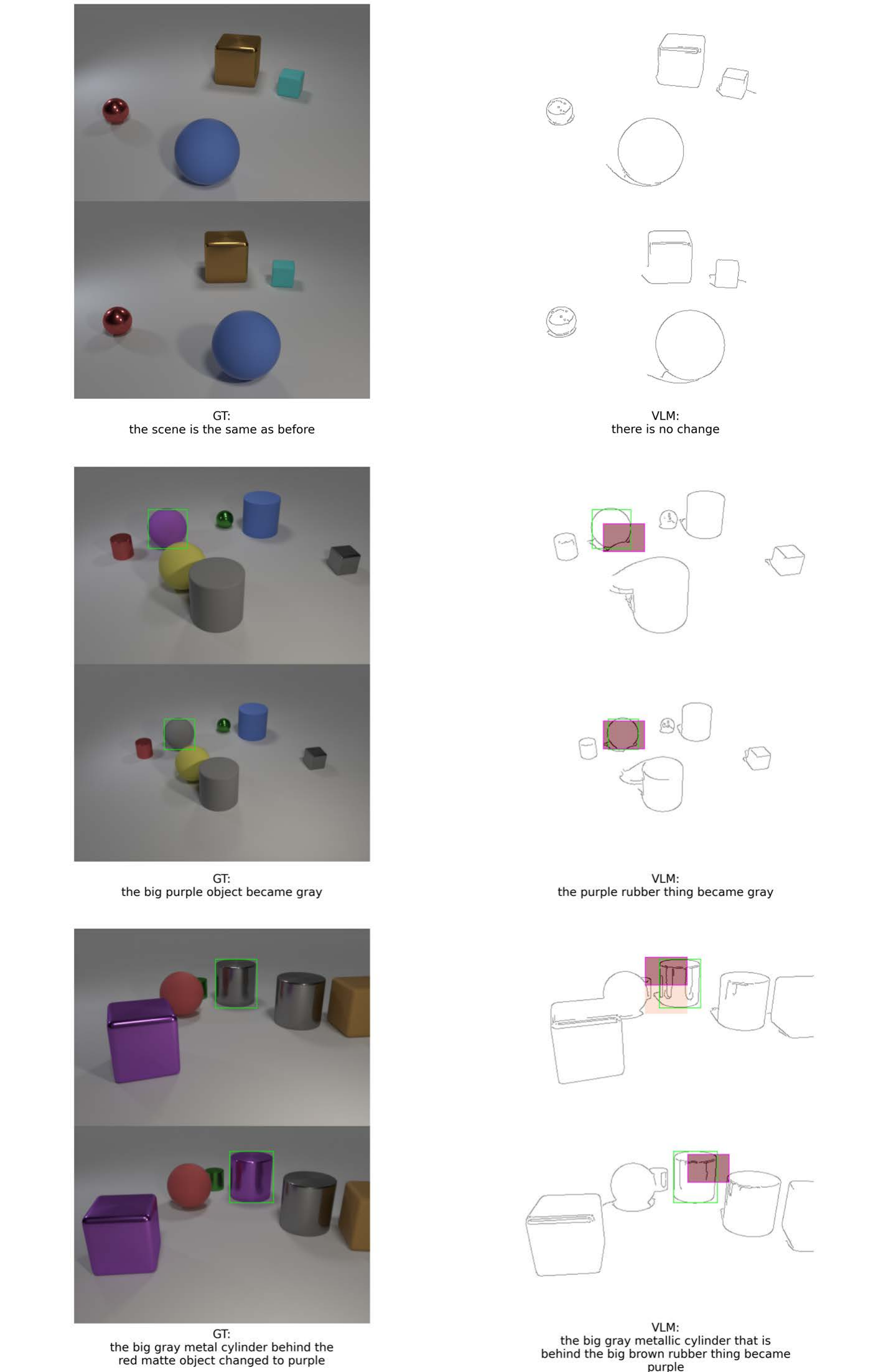}
    \caption{Captioning: \tabforidc with \VIThigh}
    \label{appfig:cl_cap_1}
\end{figure}
\begin{figure}[H]
    \centering
    \includegraphics[width=0.7\linewidth]{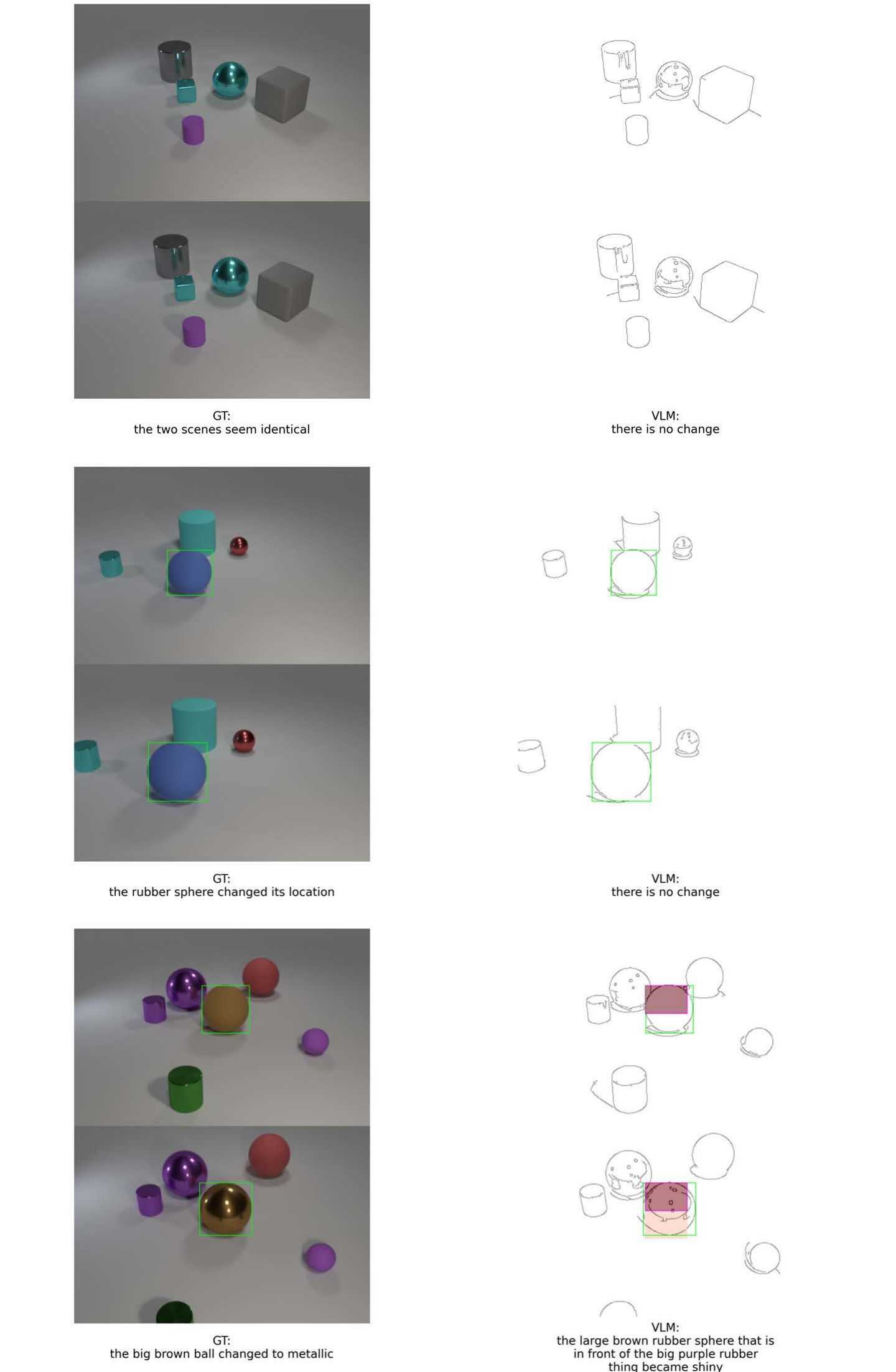}
    \caption{Captioning: \tabforidc with \VIThigh}
    \label{appfig:cl_cap_2}
\end{figure}

\begin{figure}[H]
    \centering
    \includegraphics[width=0.7\linewidth]{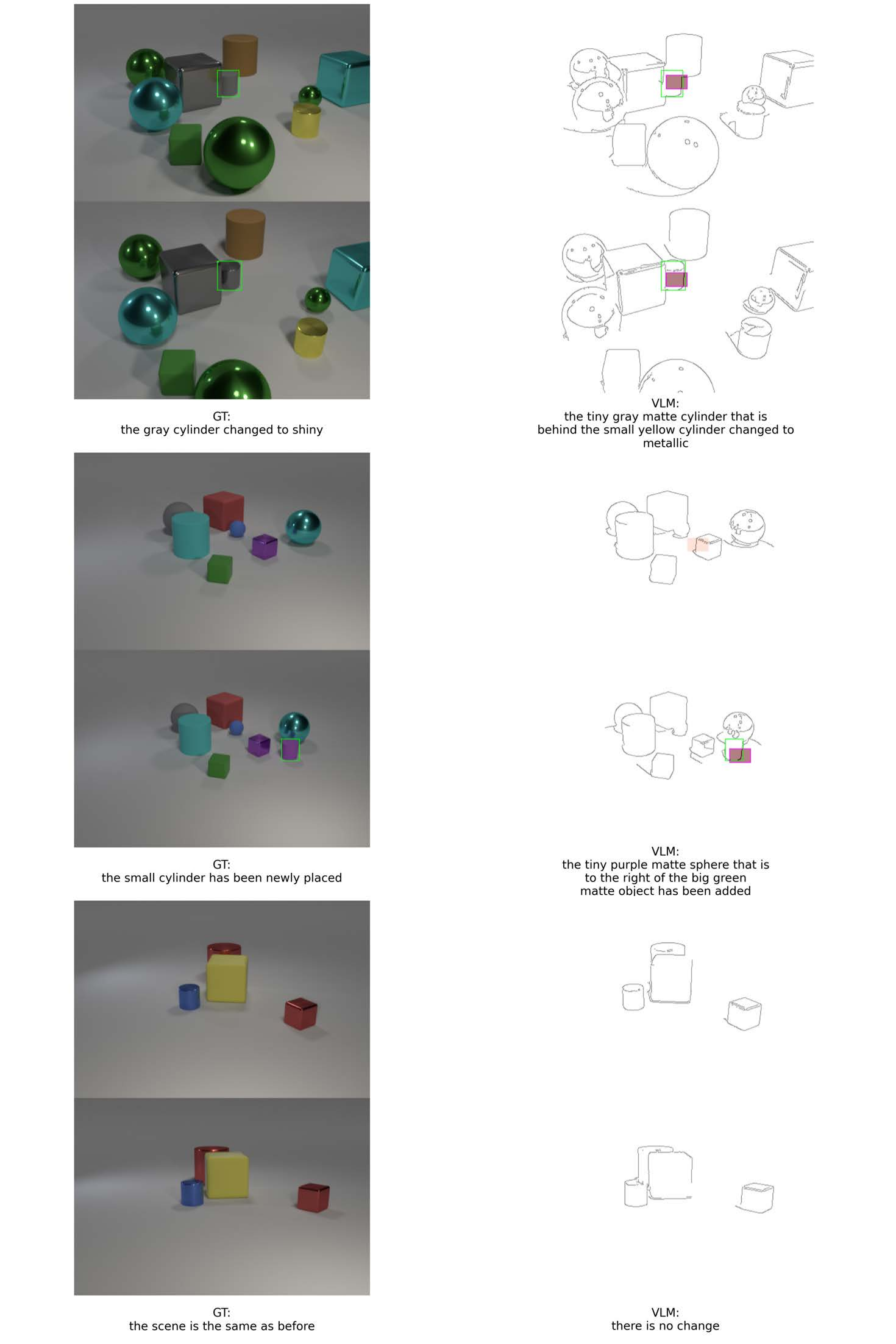}
    \caption{Captioning: \tabforidc with \VITlow}
    \label{appfig:cl_cap_3}
\end{figure}
\begin{figure}[H]
    \centering
    \includegraphics[width=0.7\linewidth]{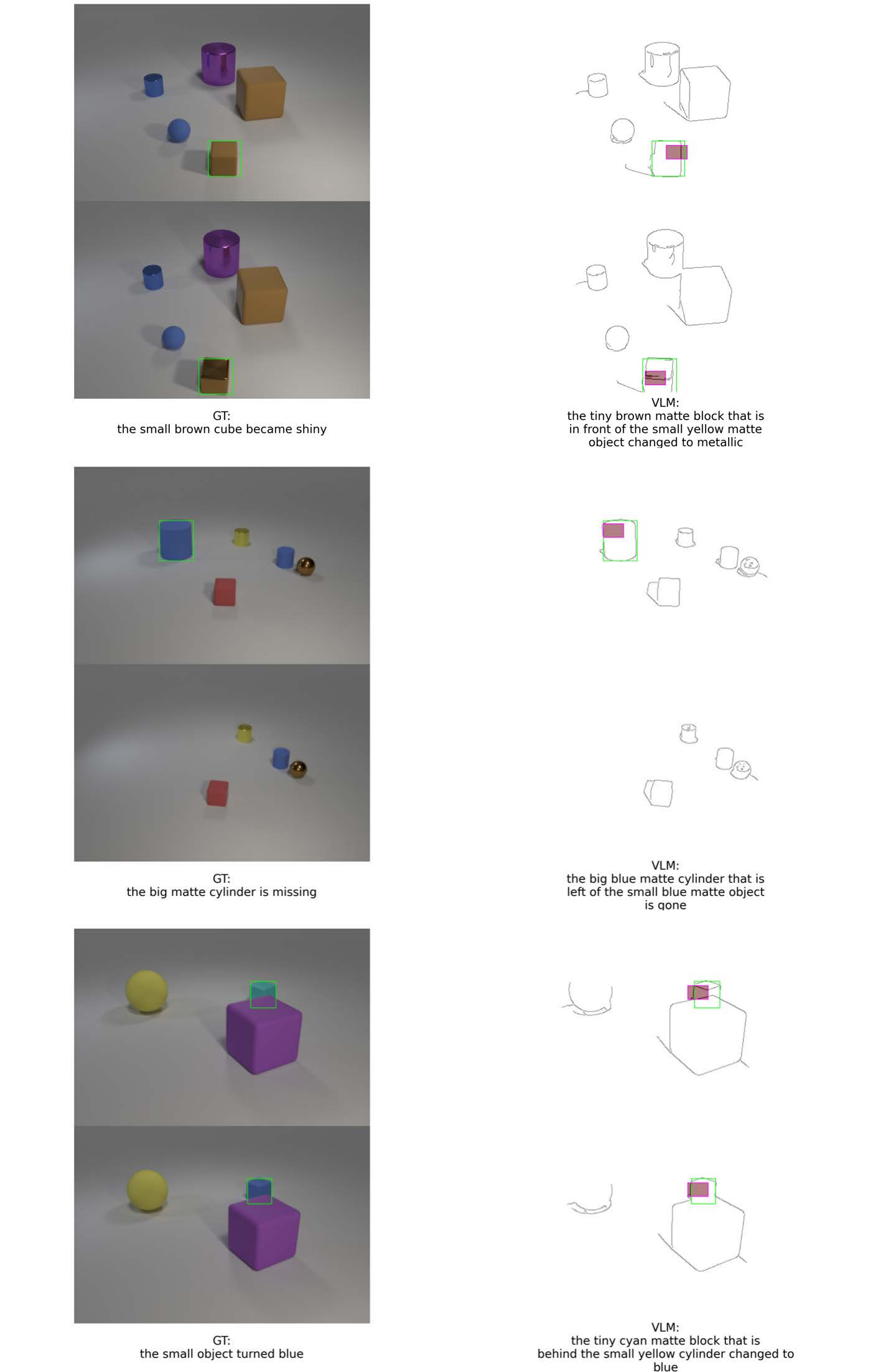}
    \caption{Captioning: \tabforidc with \VITlow}
    \label{appfig:cl_cap_4}
\end{figure}
\begin{figure}[H]
    \centering
    \includegraphics[width=0.7\linewidth]{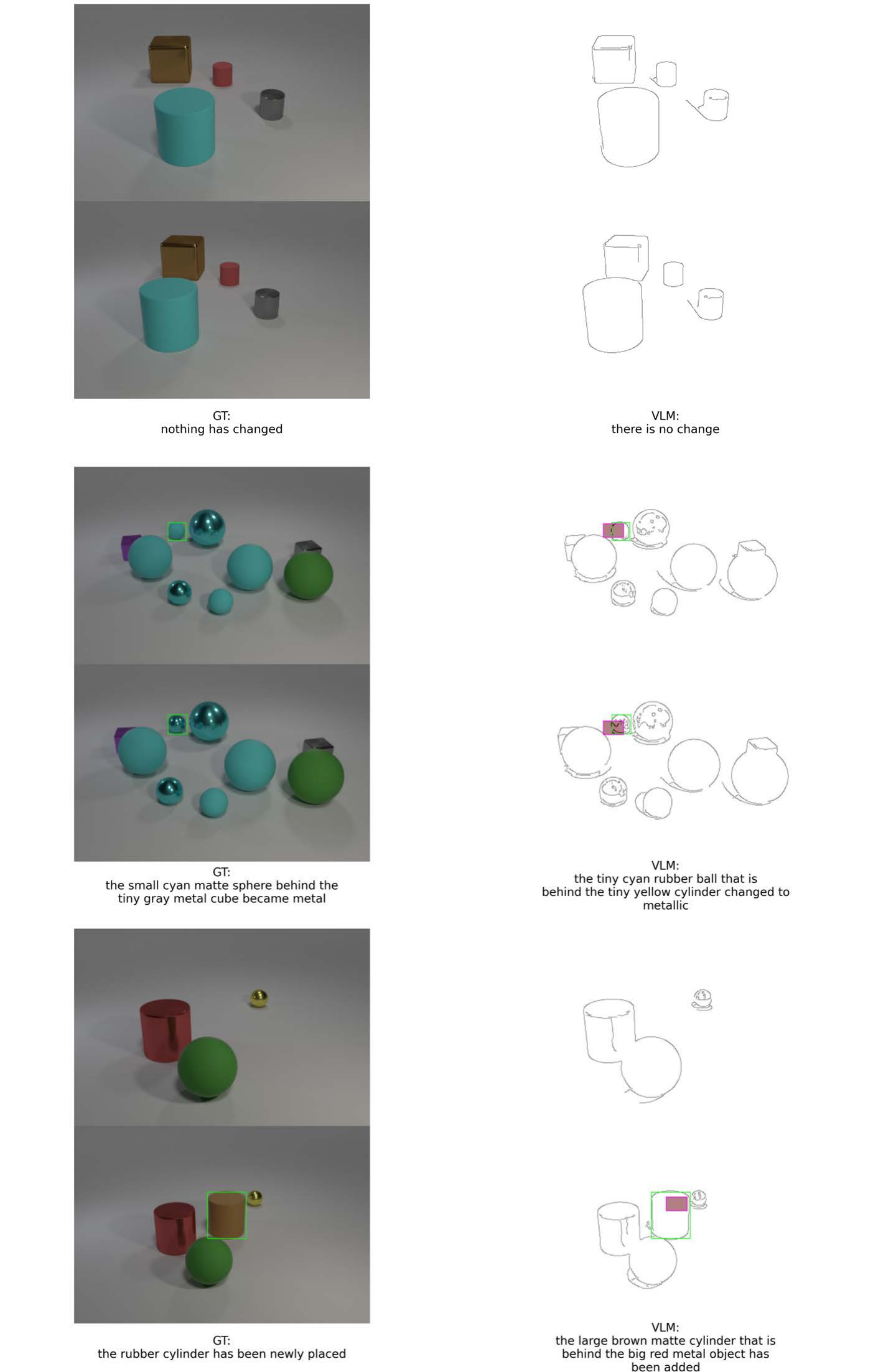}
    \caption{Captioning: \tabforidc with \VITlow}
    \label{appfig:cl_cap_5}
\end{figure}

\subsection{Correcting the attention map for change pairs}
\label{appsec:clevr_crtatt}

\begin{figure}[H]
    \centering
    \includegraphics[width=0.7\linewidth]{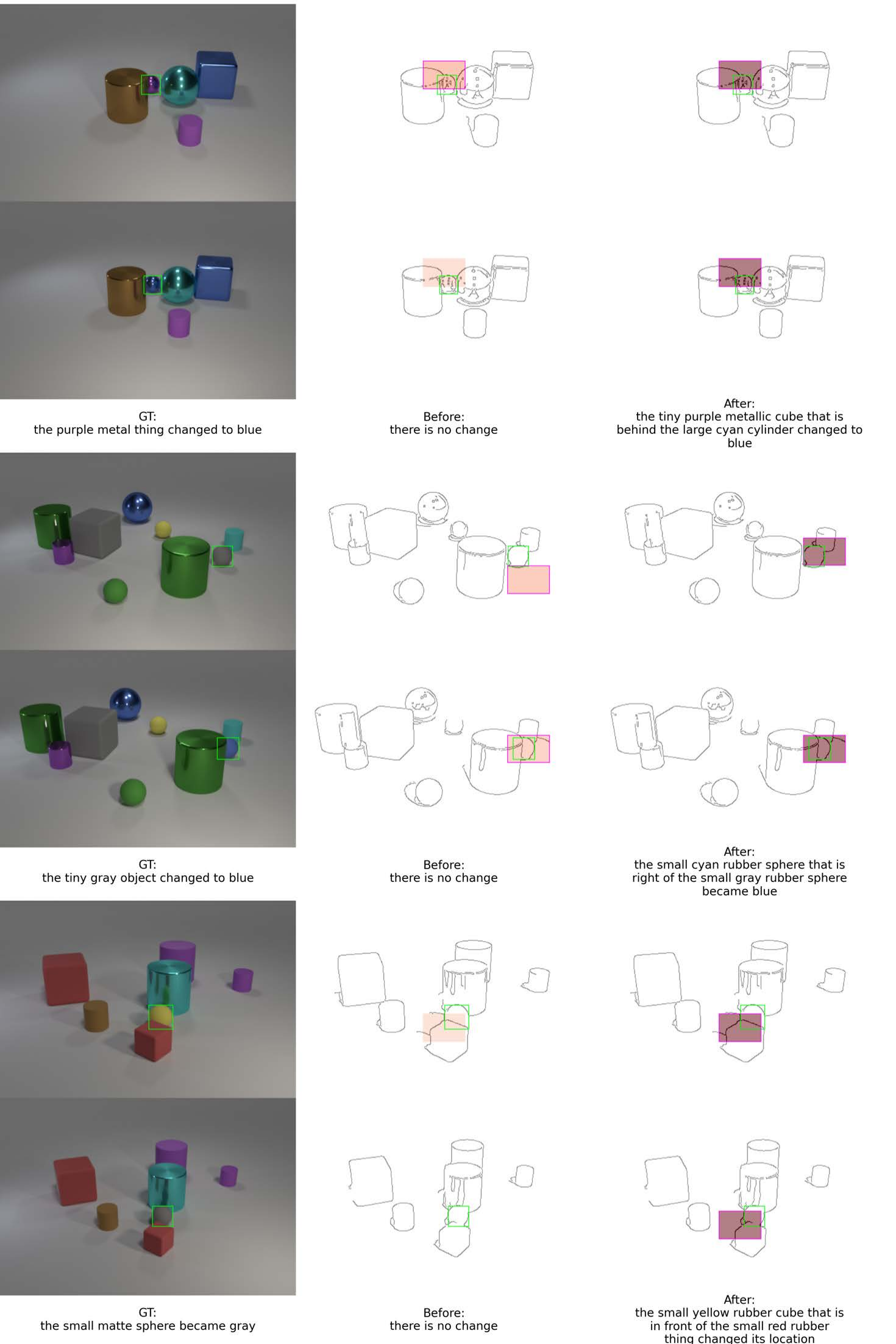}
    \caption{Editing the attention map in \tab with \VIThigh}
    \label{appfig:cl_edit_1}
\end{figure}
\begin{figure}[H]
    \centering
    \includegraphics[width=0.7\linewidth]{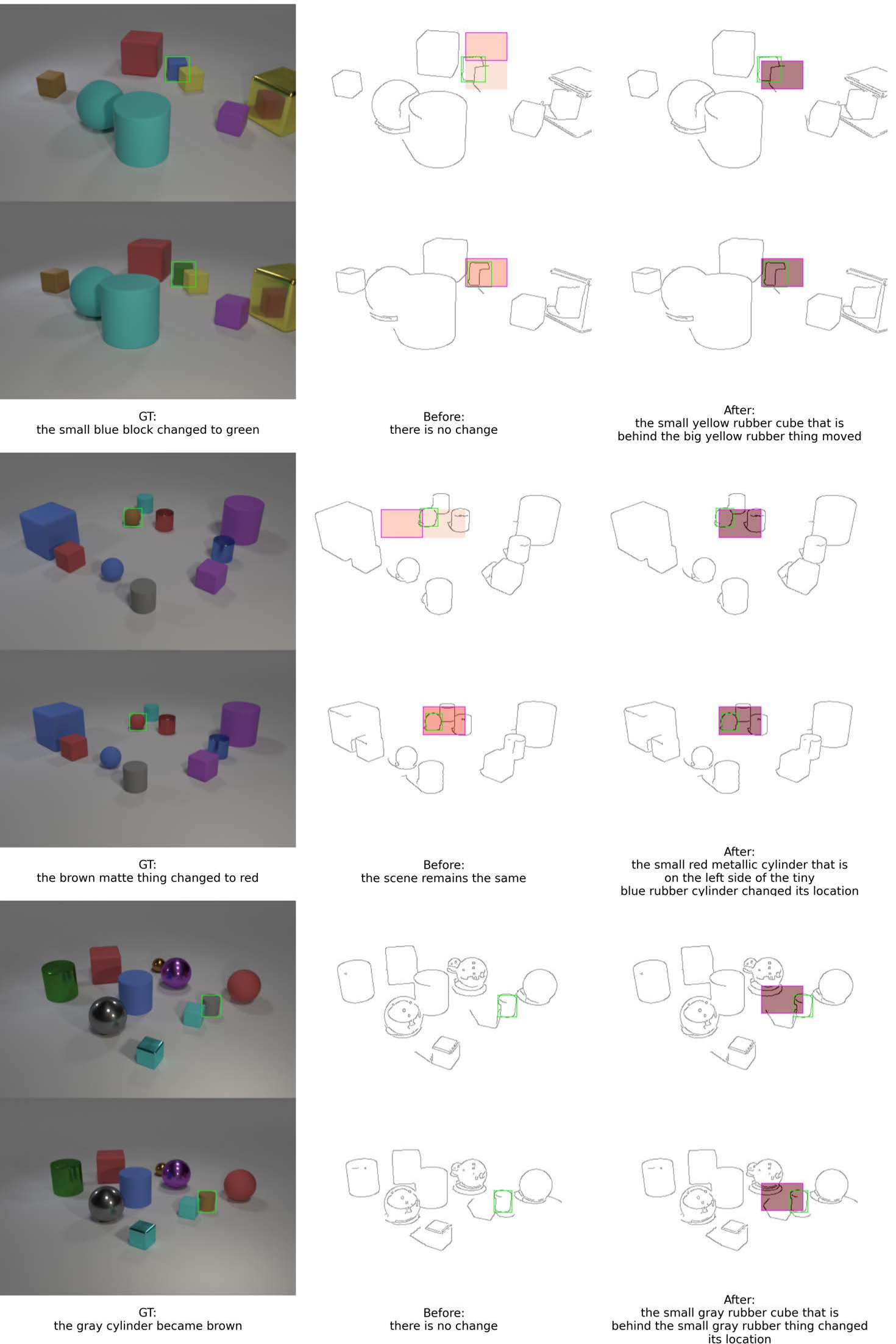}
    \caption{Editing the attention map in \tab with \VIThigh}
    \label{appfig:cl_edit_2}
\end{figure}
\begin{figure}[H]
    \centering
    \includegraphics[width=0.7\linewidth]{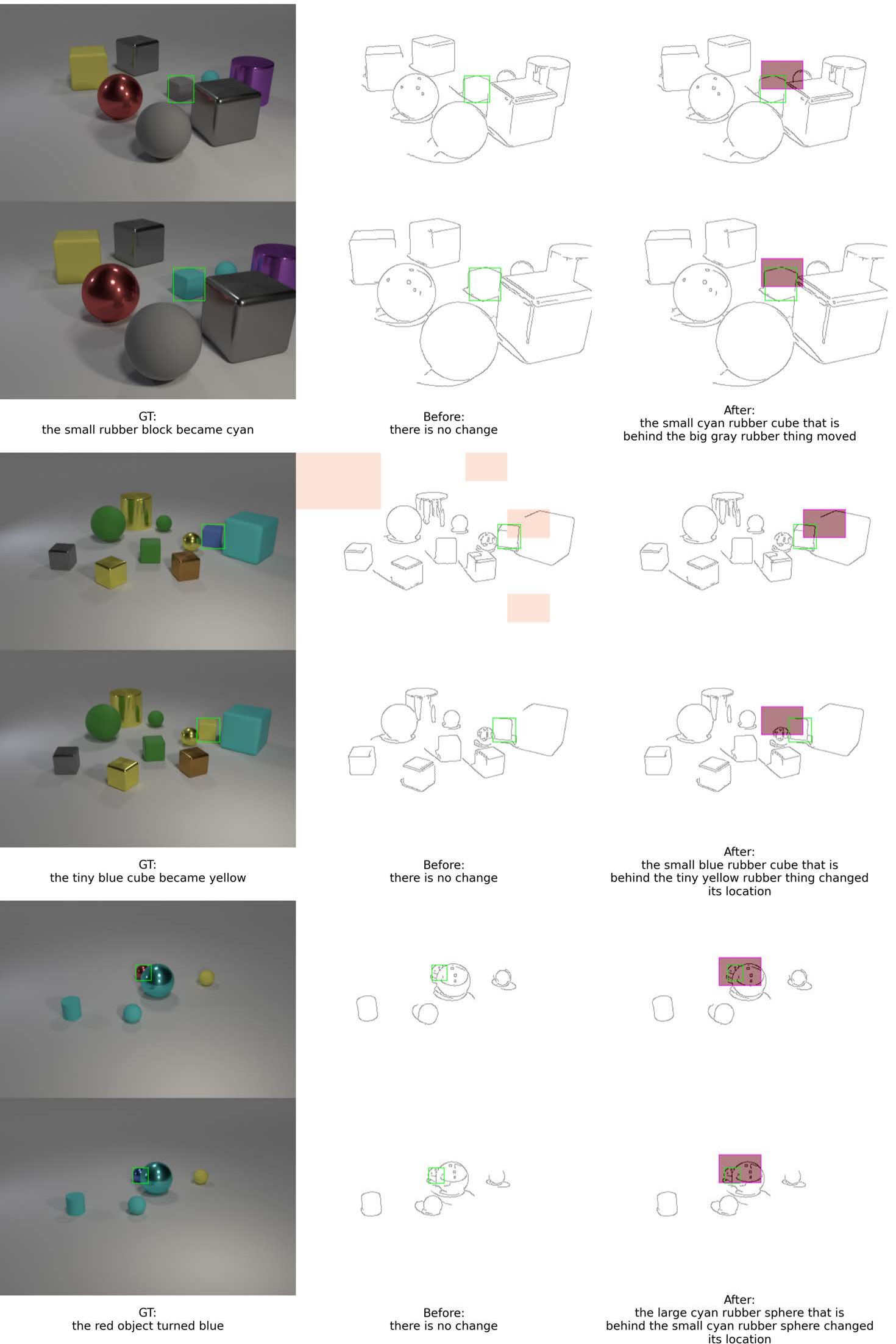}
    \caption{Editing the attention map in \tab with \VIThigh}
    \label{appfig:cl_edit_3}
\end{figure}
\begin{figure}[H]
    \centering
    \includegraphics[width=0.7\linewidth]{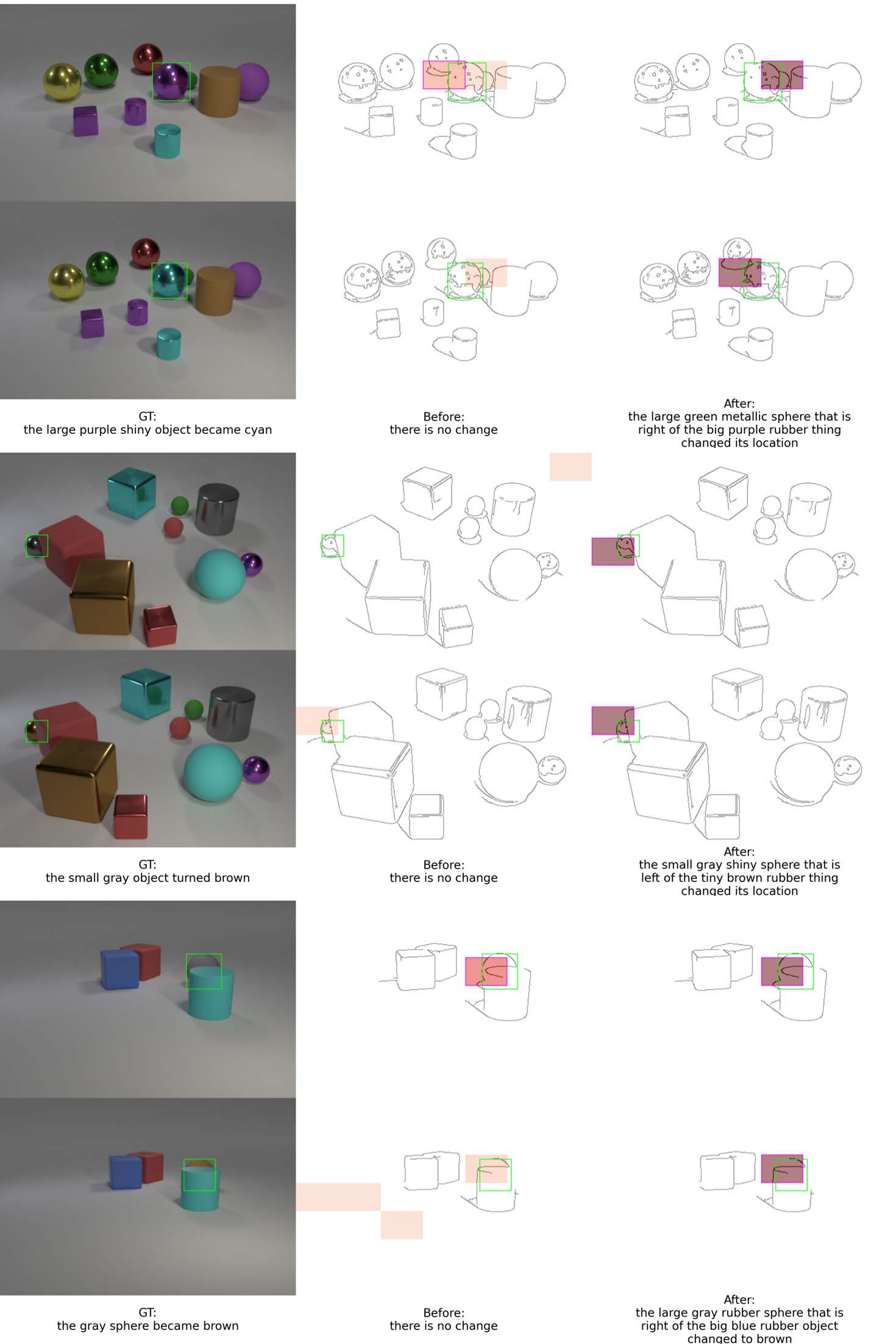}
    \caption{Editing the attention map in \tab with \VIThigh}
    \label{appfig:cl_edit_4}
\end{figure}

\begin{figure}[H]
    \centering
    \includegraphics[width=0.7\linewidth]{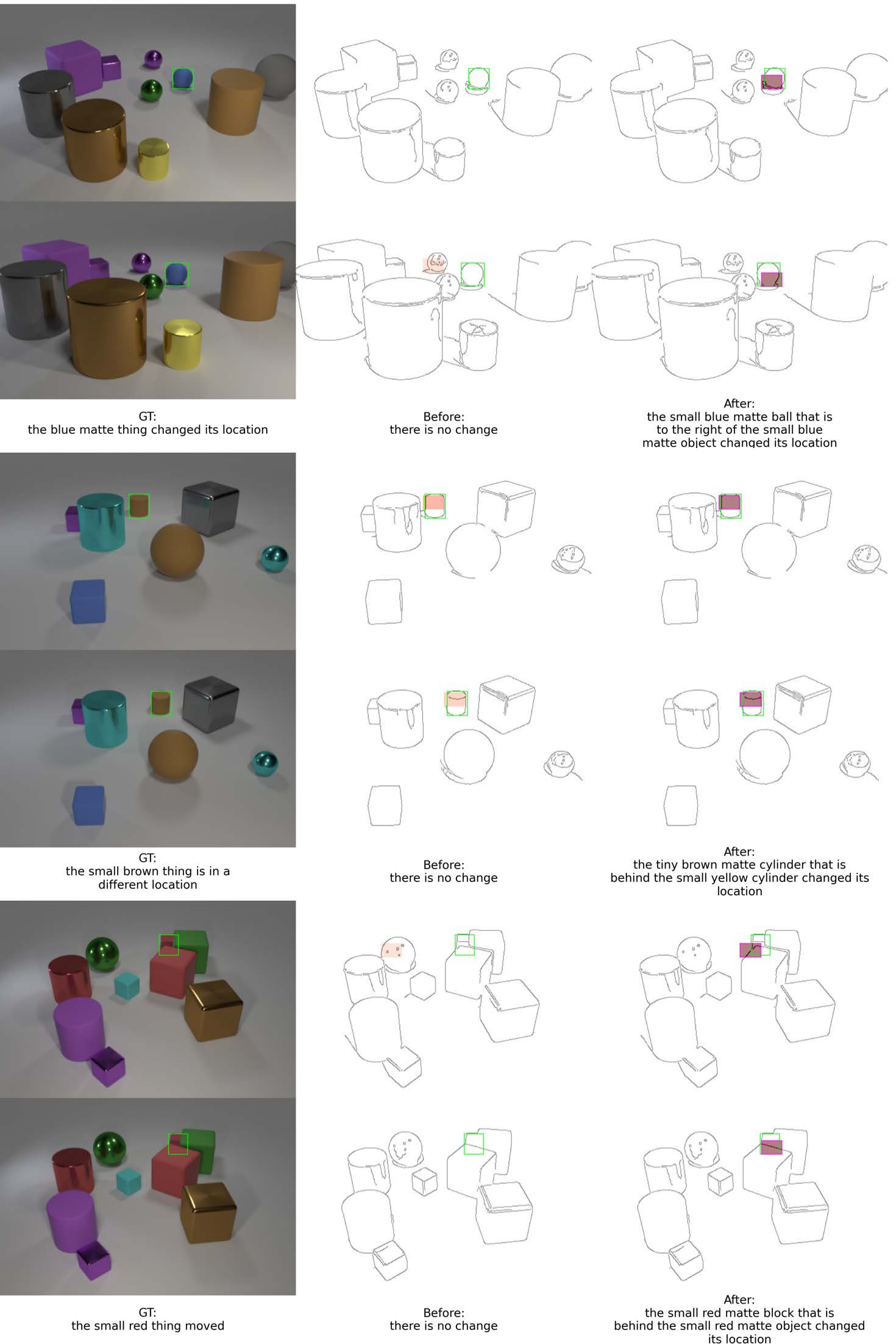}
    \caption{Editing the attention map in \tab with \VITlow}
    \label{appfig:cl_edit_5}
\end{figure}
\begin{figure}[H]
    \centering
    \includegraphics[width=0.7\linewidth]{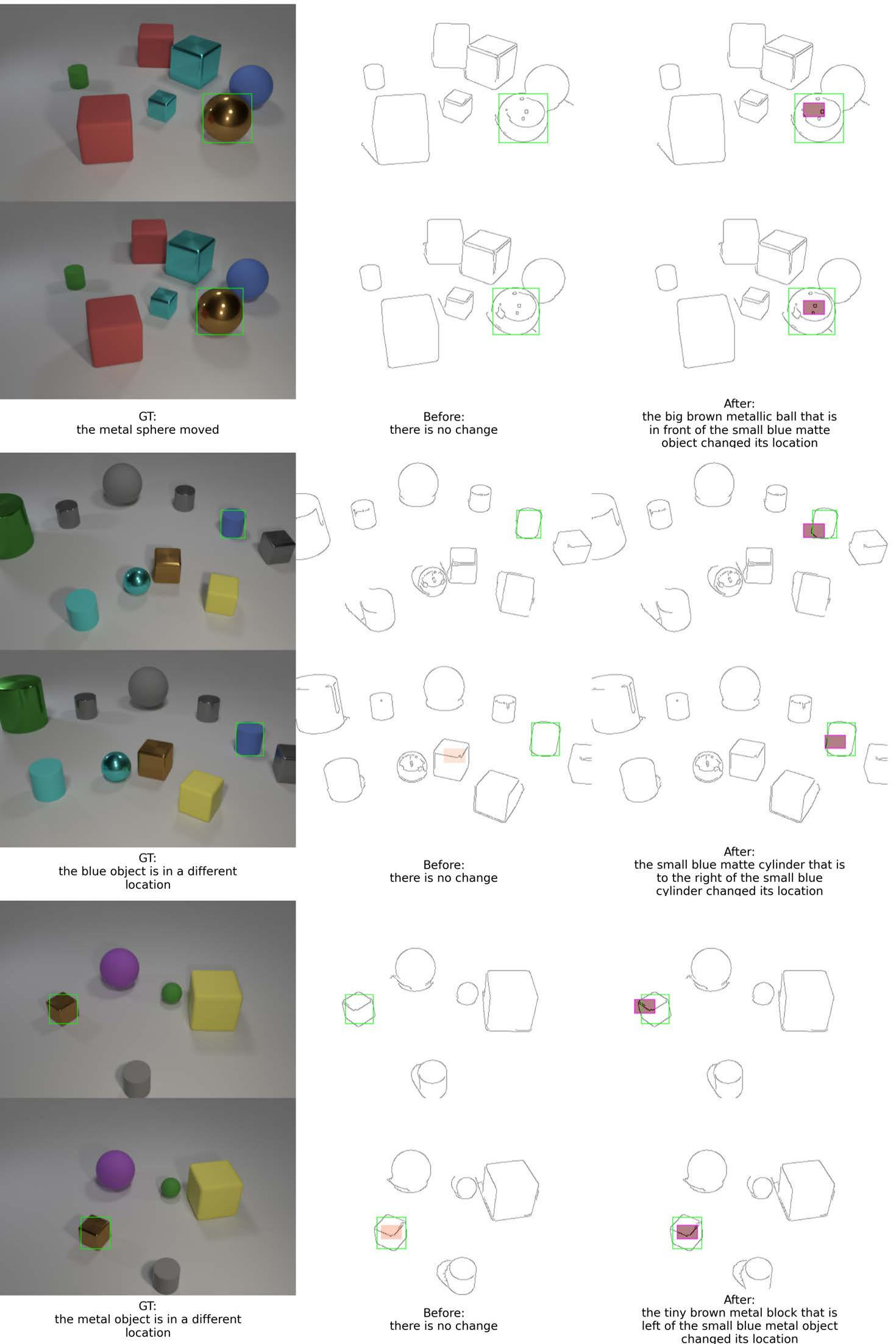}
    \caption{Editing the attention map in \tab with \VITlow}
    \label{appfig:cl_edit_6}
\end{figure}
\begin{figure}[H]
    \centering
    \includegraphics[width=0.7\linewidth]{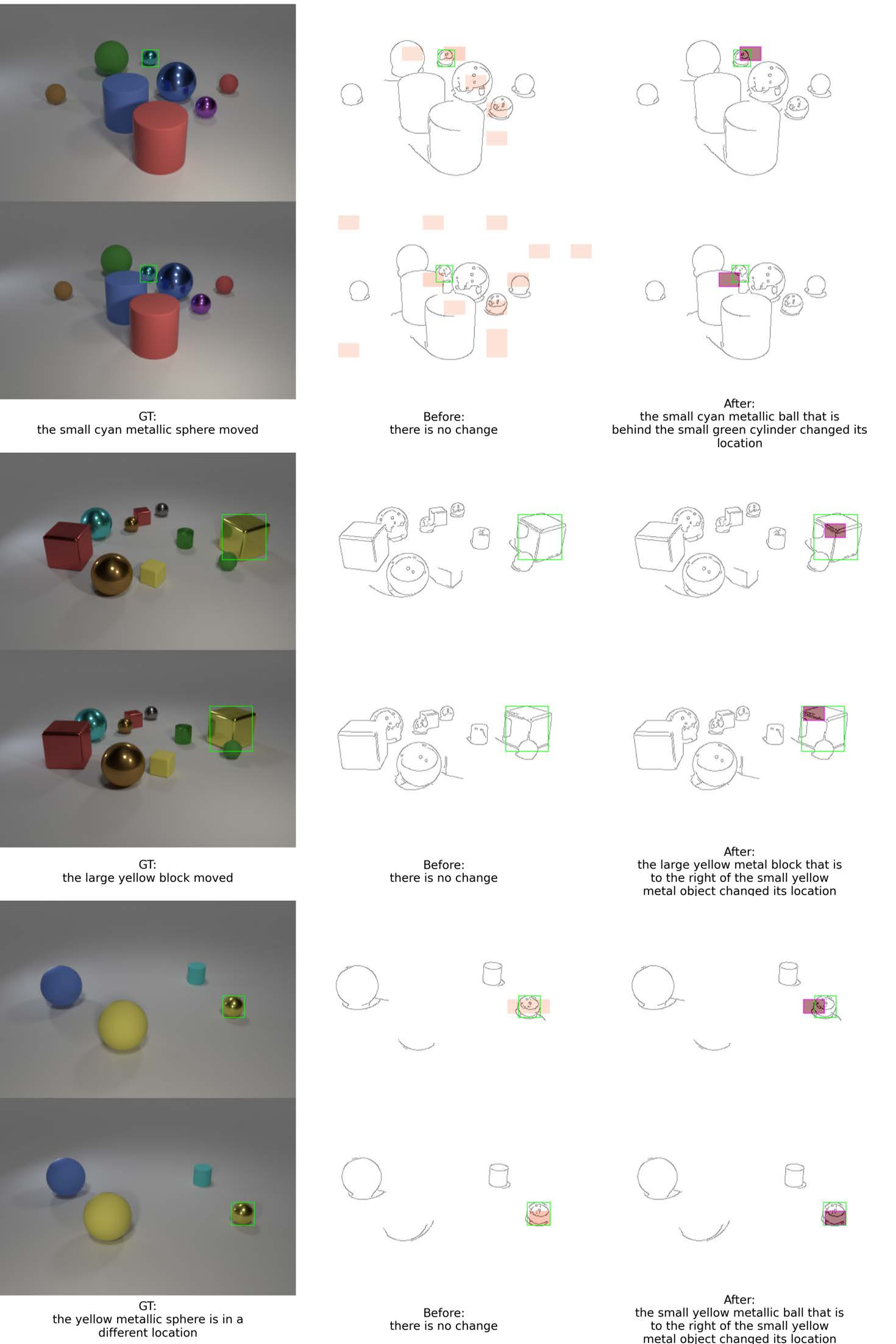}
    \caption{Editing the attention map in \tab with \VITlow}
    \label{appfig:cl_edit_7}
\end{figure}

\clearpage
\subsection{Zeroing the attention map for no-change pairs}
\label{appsec:clevr_zeroatt}

\begin{figure}[H]
    \centering
    \includegraphics[width=0.7\linewidth]{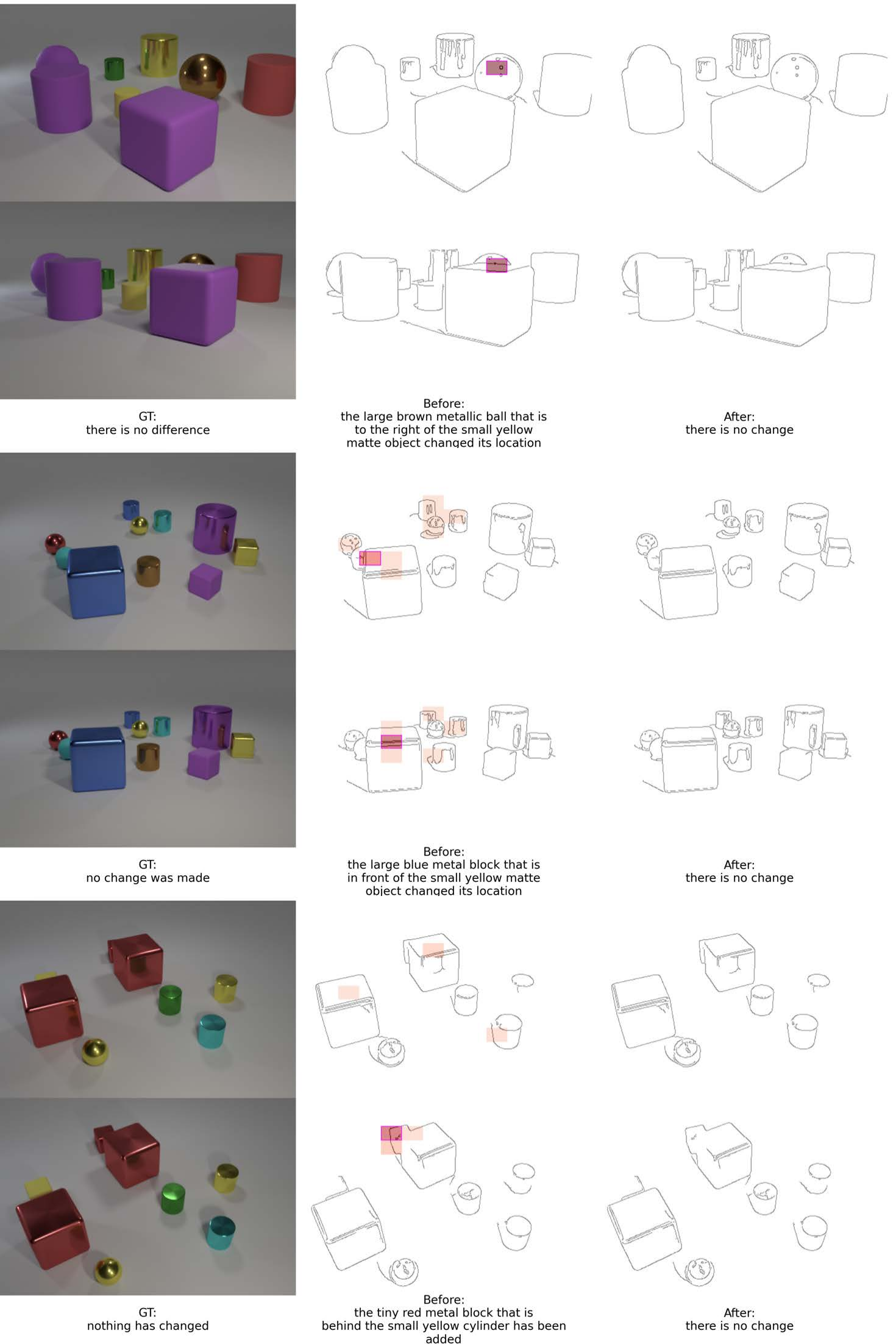}
    \caption{Editing the attention map in \tab with \VITlow}
    \label{appfig:cl_edit_8}
\end{figure}
\begin{figure}[H]
    \centering
    \includegraphics[width=0.7\linewidth]{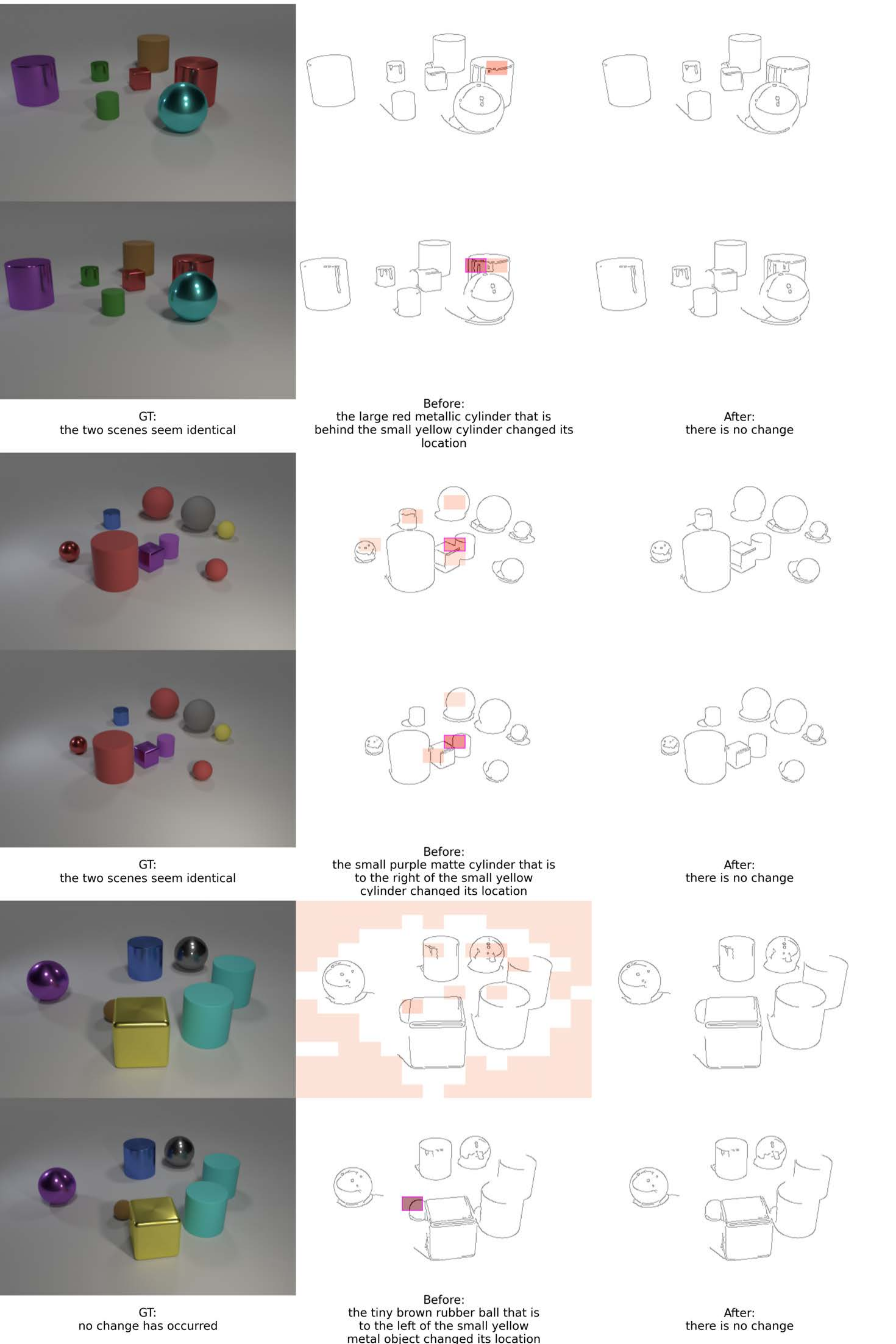}
    \caption{Editing the attention map in \tab with \VITlow}
    \label{appfig:cl_edit_9}
\end{figure}
\begin{figure}[H]
    \centering
    \includegraphics[width=0.7\linewidth]{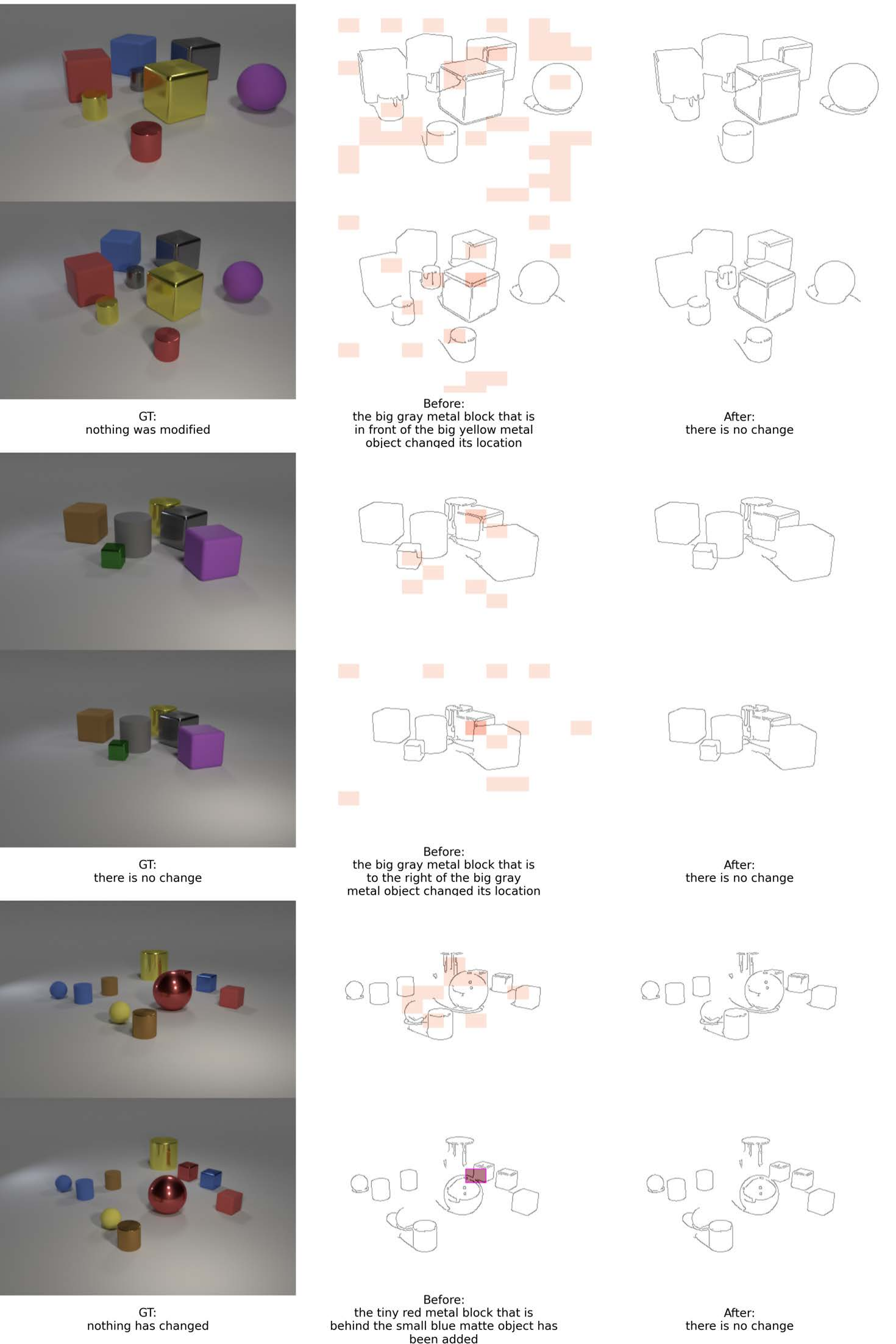}
    \caption{Editing the attention map in \tab with \VITlow}
    \label{appfig:cl_edit_10}
\end{figure}

\begin{figure}[H]
    \centering
    \includegraphics[width=0.7\linewidth]{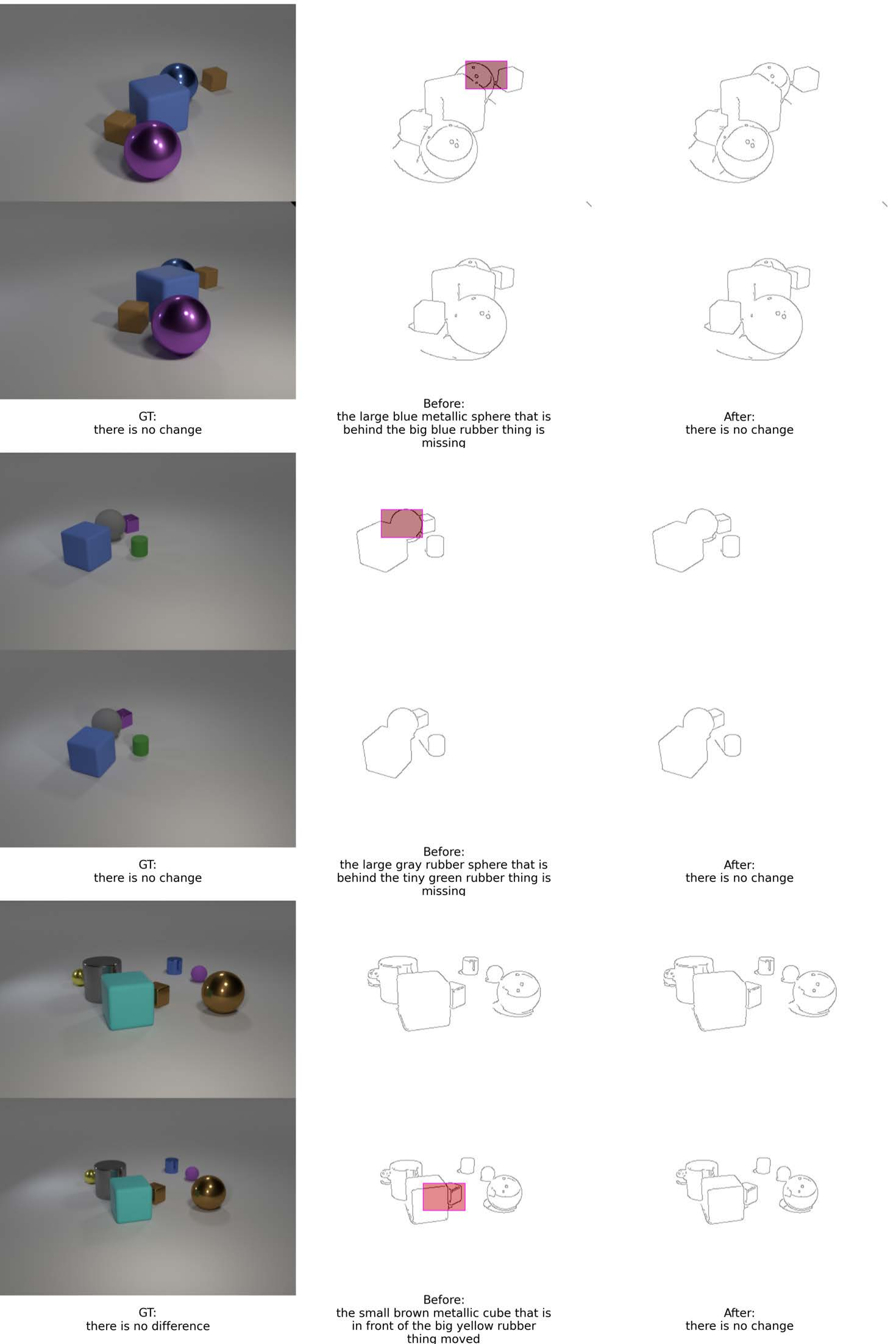}
    \caption{Editing the attention map in \tab with \VITlow}
    \label{appfig:cl_edit_11}
\end{figure}
\begin{figure}[H]
    \centering
    \includegraphics[width=0.7\linewidth]{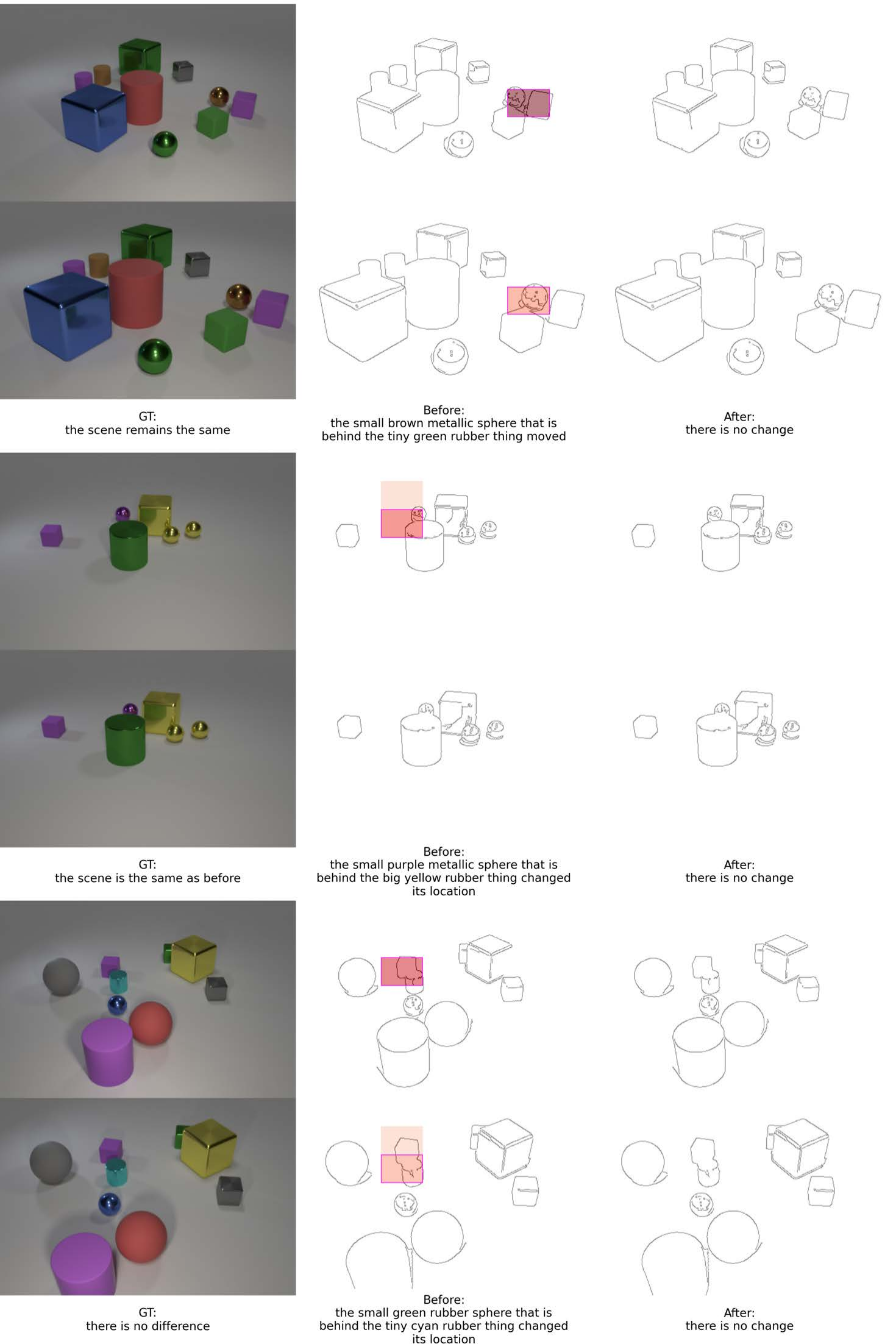}
    \caption{Editing the attention map in \tab with \VITlow}
    \label{appfig:cl_edit_12}
\end{figure}
\begin{figure}[H]
    \centering
    \includegraphics[width=0.7\linewidth]{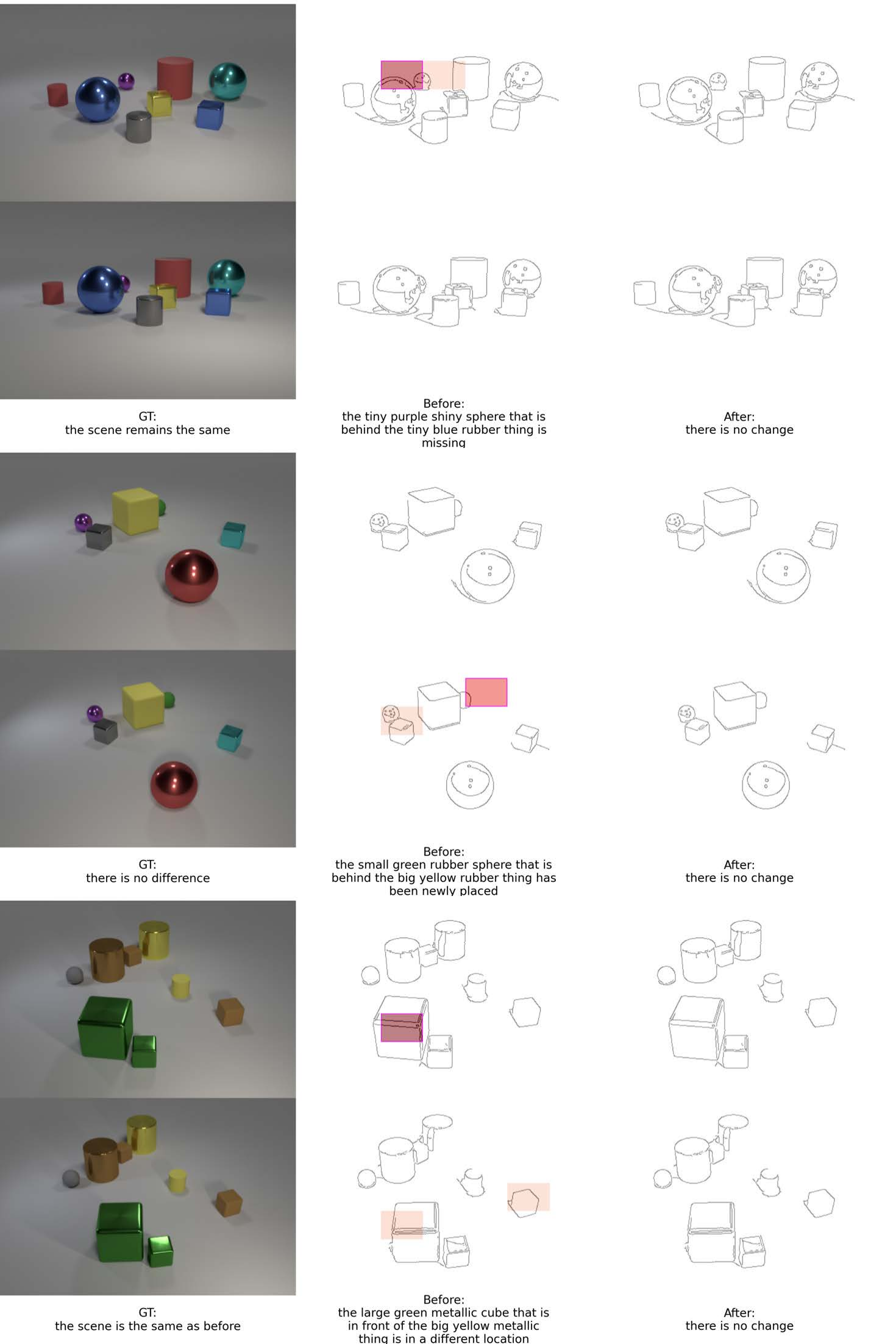}
    \caption{Editing the attention map in \tab with \VITlow}
    \label{appfig:cl_edit_13}
\end{figure}

\end{document}